%% file: paper.tex
\documentclass[review]{elsarticle}

\usepackage[a4paper, total={6in, 9in}]{geometry}

\usepackage{lineno,hyperref}
\modulolinenumbers[5]

\usepackage[utf8]{inputenc}

\usepackage{graphicx}
\usepackage{subcaption}
\usepackage[ruled]{algorithm2e}
\usepackage{amssymb}
\usepackage{amsmath}
\usepackage{multirow}
\usepackage{adjustbox}	
\usepackage{booktabs}
\usepackage{url}
\usepackage{hyperref}

\usepackage{makecell}

\usepackage{comment}
\usepackage{arydshln} 
\usepackage{float}
\usepackage[colorinlistoftodos]{todonotes}
\usepackage[acronyms]{glossaries}
\usepackage[nohyperlinks]{acronym}
\usepackage{pifont}
\input{acronims.tex} 

\usepackage{titlesec}
\titleformat{\paragraph}
{\normalfont\normalsize}{\theparagraph}{1em}{}
\titlespacing*{\paragraph}
{0pt}{3.25ex plus 1ex minus .2ex}{1.5ex plus .2ex}

\graphicspath{{figs/}}
\DeclareGraphicsExtensions{.pdf,.png,.jpg}

\newcommand{\xmark}{\ding{55}}

\newcommand{\seadronessee}{\textsc{SeaDronesSee}}%
\newcommand{\mobdrone}{\textsc{MOBDrone}}%
\newcommand{\afo}{\textsc{AFO}}%
\newcommand{\tinyperson}{\textsc{TinyPerson}}%
\newcommand{\swimmers}{\textsc{Swimmers}}%

\usepackage{gensymb} 

\journal{Elsevier}

\makeglossaries 
\input{acronims} 

\begin{document}
\begin{frontmatter}

\title{Maritime Search and Rescue Missions with Aerial Images: A Survey}

\author{Juan P. Martinez-Esteso\corref{cor1}}
\cortext[cor1]{Corresponding author}
\ead{juan.martinez11@ua.es}

\author{Francisco J. Castellanos}
\ead{fcastellanos@dlsi.ua.es}

\author{Jorge Calvo-Zaragoza}
\ead{jcalvo@dlsi.ua.es}

\author{Antonio Javier Gallego}
\ead{jgallego@dlsi.ua.es}

\address{University of Alicante, C/ San Vicente del Raspeig s/n, San Vicente del Raspeig (Alicante) - 03690, Spain}

\begin{abstract}
The speed of response by search and rescue teams at sea is of vital importance, as survival may depend on it. Recent technological advancements have led to the development of more efficient systems for locating individuals involved in a maritime incident, such as the use of Unmanned Aerial Vehicles (UAVs) equipped with cameras and other integrated sensors. Over the past decade, several researchers have contributed to the development of automatic systems capable of detecting people using aerial images, particularly by leveraging the advantages of deep learning. In this article, we provide a comprehensive review of the existing literature on this topic. We analyze the methods proposed to date, including both traditional techniques and more advanced approaches based on machine learning and neural networks. Additionally, we take into account the use of synthetic data to cover a wider range of scenarios without the need to deploy a team to collect data, which is one of the major obstacles for these systems. Overall, this paper situates the reader in the field of detecting people at sea using aerial images by quickly identifying the most suitable methodology for each scenario, as well as providing an in-depth discussion and direction for future trends.
\end{abstract}


\end{frontmatter}



\section{Introduction}  
\label{sec:introduction}

\begin{figure}[!ht]
    \centering
    \includegraphics[width=1.\textwidth]{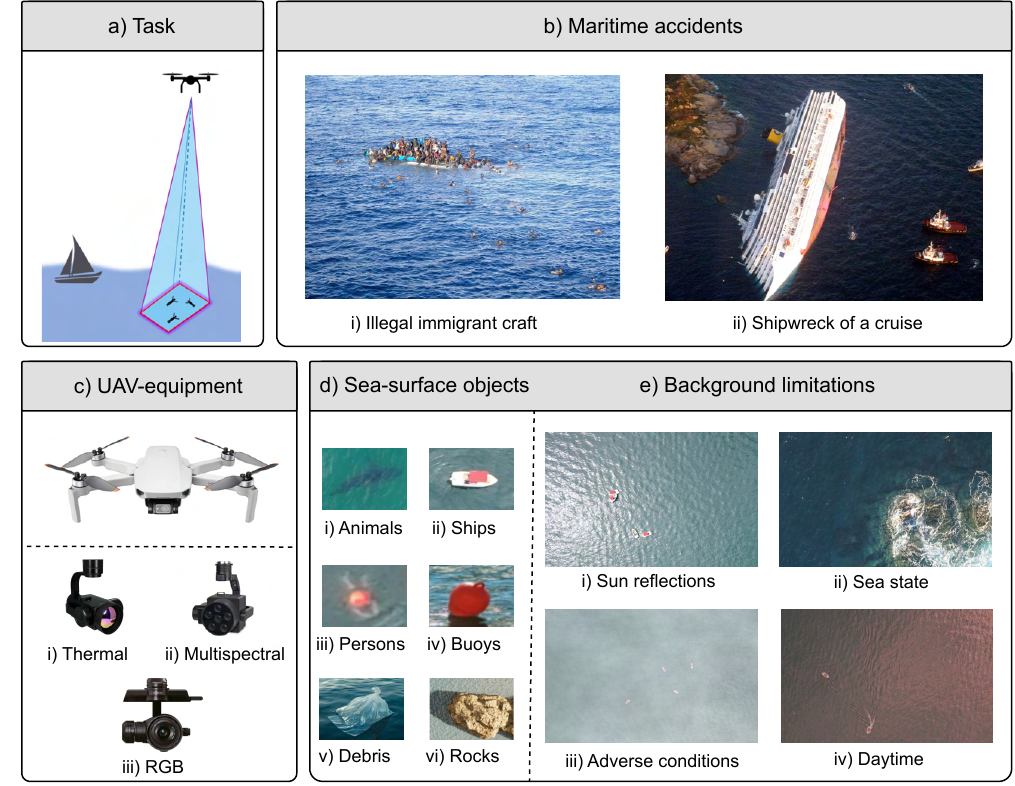}
    \caption{Graphical overview of the task, concerns, UAV equipment, and background limitations.}
    \label{fig:introd}
\end{figure}

Maritime Search and Rescue (SAR) operations are critical for saving lives during emergencies at sea~\cite{breivik2013advances}. The nature of these operations can vary significantly depending on the environment and circumstances, but generally involves responding to incidents such as shipwrecks, capsized vessels, or collisions. Additionally, the increasing number of perilous maritime crossings, often related to migration, has led to a rise in drownings, further complicating SAR efforts. These missions encompass three main tasks: searching, assisting, and rescuing affected people. Although the approach to each mission may differ depending on the situation, the speed of response is a crucial factor in all scenarios, particularly when lives are at stake~\cite{papanicolopulu2016duty}. However, accelerating rescue operations often requires greater deployment of resources, which in turn raises costs and can compromise the mission's success. Furthermore, the availability of specialized equipment, which is often scarce, adds another layer of complexity and expense. Addressing these challenges demands urgent global attention, alongside technological advancements to improve the surveillance, detection, and rescue of individuals at sea. In Figure~\ref{fig:introd}, we present a summary of the main aspects discussed in this section: a) A graphical representation of how the task is performed, b) some of the most significant and recurring maritime accidents, c) the equipment used for these operations, d) the types of objects that can be found on the sea surface, and e) varying visual appearances of the sea surface.

Given these challenges, the development of more efficient systems, such as Unmanned Aerial Vehicles (UAVs), commonly known as drones (see Figure~\ref{fig:introd}c), has the potential to greatly enhance the success of rescue missions. UAVs represent a paradigm shift in locating survivors. Equipped with sensors and high-resolution cameras, these drones can capture various types of images---thermal, multispectral, and visible spectrum---to provide aerial views of the environment. This enables the detection of victims' positions in maritime accidents, assisting in their rescue, and transmitting their location and condition to rescue teams in real-time. UAVs can swiftly cover large sea areas and access locations that are difficult or impossible for traditional search vehicles, making them versatile and efficient tools~\cite{waharte2010supporting}. Compared to helicopters, UAVs offer significant advantages, such as the ability to fly at lower altitudes, faster response times, and lower costs~\cite{restas2015drone}, thus optimizing rescue operations~\cite{nair2019designing}. UAVs can be remotely controlled by trained pilots or operate autonomously, blending technological innovation with the critical need for timely humanitarian aid. Nevertheless, the effective use of UAVs requires integration with robust and reliable systems capable of accurately extracting essential information from images, as the sea surface often contains other elements that complicate the rescue process. These include (see Figure~\ref{fig:introd}d): ships, debris, buoys, marine animals, rocks, waves, or foam, which create visual noise and make it more difficult to identify people in need of assistance.

Recent advancements in artificial intelligence have created opportunities for more efficient methodologies in various tasks, and the SAR field is no exception. The literature includes studies that introduce new methods to facilitate these operations~\cite{seger2019coagency,lomonaco2018intelligent}. Although the combination of UAVs with machine learning techniques has the potential to improve SAR efforts, it remains challenging due to the nature of the task. Using aerial images to detect people at sea is particularly difficult, as the small size of humans in contrast to the vast ocean makes detection arduous. In addition to speed, precision is also a crucial factor in determining the success rate of rescue operations~\cite{martinez2021search}. Various solutions and architectures based on neural networks address challenges such as classification, segmentation, tracking, and object detection in maritime environments~\cite{kyrkou2018dronet}. However, external factors that include uncontrollable water reflections~\cite{lygouras2019unsupervised}, rapid changes in sea state caused by waves~\cite{fefilatyev2012detection}, adverse weather conditions, and variations in illumination at different times of the day can negatively impact performance (see Figure~\ref{fig:introd}e). Consequently, recent research has focused on mitigating these issues to improve detection accuracy and build more robust models~\cite{cruz2016aerial,wang2023cooperative}. A critical challenge for these algorithms is the requirement for large amounts of labeled data~\cite{kim2022effects, al2015efficient}, which is often difficult to obtain in real-world settings. Addressing data scarcity, such as through the generation of synthetic data, can help enhance the performance and effectiveness of machine learning algorithms, not only in this field~\cite{zhang2023semi} but also in a wide range of applications~\cite{andriyanov2020using, bansal2022systematic}.

The existing literature covers several surveys on closely related areas, including object detection from images captured from boats, covering topics such as bridges, buoys, or ships~\cite{zhang2021survey}, marine object detection~\cite{wang2022review}, SAR on land~\cite{bai2023review}, and the use of electro-optical sensors for object detection at sea~\cite{lyu2022sea}. Additionally, some reviews describe the operation of unmanned vehicles, including UAVs, unmanned surface vessels, and underwater unmanned vehicles, which emphasize the importance of these technologies for automatic monitoring~\cite{li2023survey}. There is also a survey on datasets for general object detection in maritime contexts~\cite{su2023survey}. These reviews highlight the relevance of various aspects within the SAR field. 

In contrast to the aforementioned reviews, our paper focuses specifically on the detection of people at sea for rescue operations using UAVs. To the best of our knowledge, this is the first comprehensive review dedicated to this topic. We provide an in-depth analysis of materials, existing methodologies, and experimental results in this domain, while also discussing current limitations, identifying future research directions, and drawing conclusions to guide ongoing efforts. 
By addressing this gap in the literature, our objective is to contextualize the reader and offer a reference point for future work in this area, which is particularly important today due to the increasing number of maritime accidents each year~\cite{emsa2018annual}.

The main contributions of our survey, compared to the existing literature, are as follows:

\begin{itemize}
    \item An exhaustive study of current methodologies for SAR operations at sea, including classification, segmentation, object detection, and tracking approaches.
    \item A detailed review of real and synthetic datasets used in maritime rescue contexts.
    \item An overview of tools and techniques used for generating synthetic data and its integration with real data.
    \item A comparison of results obtained from methods applied to the most widely used benchmarks.
    \item A discussion of the current research landscape, including limitations, future directions, and open challenges.
\end{itemize}

The remainder of this paper is organized as follows: Section~\ref{sec:method} presents a review of state-of-the-art SAR methods for detecting individuals at sea, from traditional approaches to the latest advancements. Section~\ref{sec:datasets} introduces the most widely used public materials. Section~\ref{sec:metrics} outlines the evaluation metrics typically employed in the field. Section~\ref{sec:results} summarizes the experimental results of existing contributions. Section~\ref{sec:discussion} discusses the methods, results, open challenges, and trends identified. Finally, Section~\ref{sec:conclusions} presents concluding remarks and directions for future research.

\section{Methods}
\label{sec:method}


The research in maritime SAR can be divided into several categories based on the objectives of the methodologies. These categories include \emph{image classification} to identify the presence of people at sea, \emph{image segmentation} to detect the pixels representing a person, \emph{object detection} to retrieve the exact location of a person, \emph{tracking} to monitor individuals in video sequences, and \emph{general detection approaches} applied to maritime SAR. Additionally, there is a growing focus on the use of \emph{synthetic data generation} to create varied and representative datasets for training models. This structure will guide our discussion of existing methods throughout the current section. We also include a final mention to briefly review related research lines that complement the areas mentioned above. 

Each task addressed in SAR has its specific requirements. For example, most monitoring systems in this domain involve the integration of artificial intelligence mechanisms into UAVs. These systems must balance fast processing speeds with high precision, as they are often deployed in real-time rescue operations. Moreover, since UAVs have limited computational resources, the systems must operate efficiently to maintain sufficient autonomy during exploration missions.

Table~\ref{tab:methodology} provides an overview of the methods included in our survey, summarizing the most relevant characteristics. To facilitate a quick comparison of the strengths and weaknesses of each approach, the table also includes a broad estimation of performance (column \textit{Acc.}) and efficiency (column \textit{Eff.}), based on the results reported in each study, to indicate their relative potential compared to other works. In the following sections, we introduce and compare the key factors of these methods, highlighting the evolution of contributions in the field. 

\input{table_methods}

\subsection{Classification}

One of the tasks frequently addressed in the literature on SAR for detecting people at sea is classification. This can be approached in two ways: by classifying an input image to determine the presence of humans, or by identifying the type of an object once it has been detected. The first rows of Table~\ref{tab:methodology} include the approaches described in this section.

\begin{figure}[!ht]
    \centering
    \includegraphics[width=0.8\textwidth]{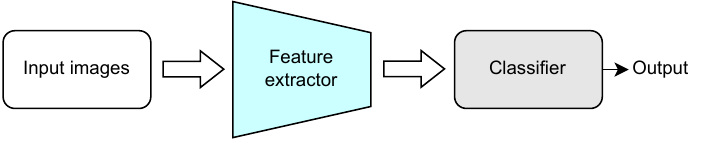}
    \caption{General scheme for the classification task. First, the images are passed through a feature extractor, and then a classifier is used to decide to which category each image belongs.}
    \label{fig:classification_method}
\end{figure}

Most methods use a feature extractor that processes the images, followed by a classifier (see Fig.~\ref{fig:classification_method}). One of the first proposals was by Leira et al.~\cite{leira2015automatic}, who addressed the classification of previously detected objects using traditional techniques applied to thermal images. Their approach used object size, temperature data from a thermal camera, and Hu's invariant moments to address scale, rotation, and translation~\cite{HuMoments1962}, which were then applied in a nearest-neighbor classifier (NNC).

Rodin et al.~\cite{rodin2018object} employed thermal images and a \ac{GMM} to discriminate between foreground objects and background. They trained a \ac{CNN} to classify boats, buoys, people, and pallets. However, the results were inconsistent due to motion blur, which distorted the appearances of objects. 

The most recent contribution in this area comes from Gallego et al.~\cite{gallego2019detection}, who proposed a method to classify multispectral images using a custom \ac{CNN} into two categories: body or not body. While the results were promising, precise localization was still required due to the large coverage area of the images.

\subsection{Object segmentation}
\label{sec:method:object_segmentation}

Segmentation is a common process in many image processing applications, with the aim of detecting pixels that represent relevant information according to the context to which it is applied. In the case of maritime SAR, the segmentation process involves classifying each pixel of the image according to whether it represents a person, a boat, or any other element to be considered. This allows isolating the detected elements with a high degree of detail for subsequent post-processing to extract the required information. For example, it could be useful to detect the contours of people in the sea, thereby determining their exact position and condition. The approaches that adopt this type of task are summarized in the second part of Table~\ref{tab:methodology}.

\begin{figure}[!ht]
    \centering
    \begin{subfigure}{0.35\textwidth}
        \includegraphics[width=1\textwidth]{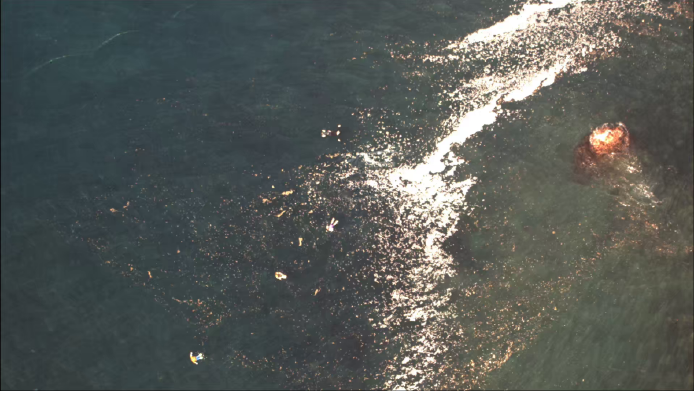}
        \caption{Original image.}
        \label{fig:points_random}
    \end{subfigure}
    \begin{subfigure}{0.35\textwidth}
        \includegraphics[width=1\textwidth]{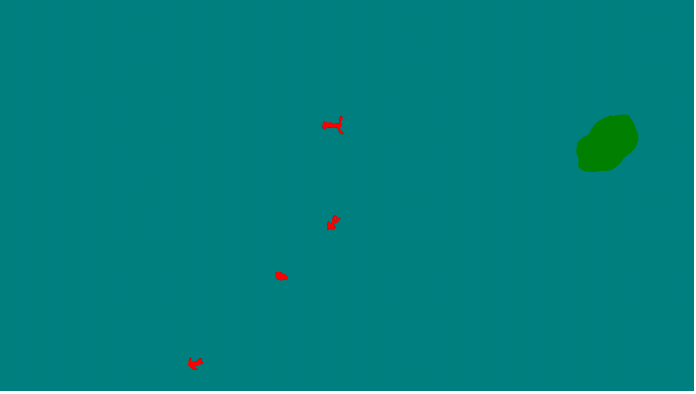}
        \caption{Semantic segmentation.}
        \label{fig:points_sequential}
    \end{subfigure}
    
    \begin{subfigure}{0.35\textwidth}
        \includegraphics[width=1\textwidth]{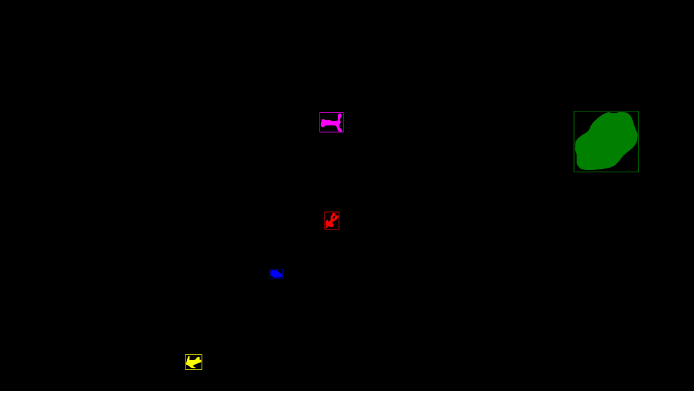}
        \caption{Instance segmentation.}
        \label{fig:points_inkrate}
    \end{subfigure}
    \begin{subfigure}{0.35\textwidth}
        \includegraphics[width=1\textwidth]{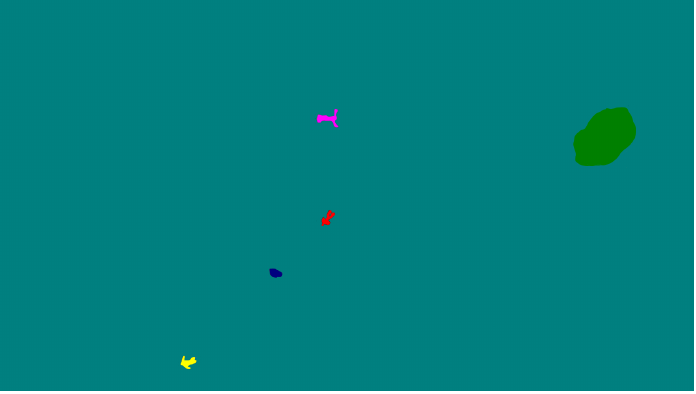}
        \caption{Panoptic segmentation.}
        \label{fig:points_entropy}
    \end{subfigure}
 \caption{Examples of the different perspectives for segmentation tasks for maritime rescue operations. In this particular case, the images present objects belonging to two categories: person and rock.}
 \label{fig:segmentation_general}
\end{figure}

Segmentation can be approached from different perspectives according to the system's requirements. Specifically, the literature generally covers three cases, represented by an example in Fig.~\ref{fig:segmentation_general}: \textit{semantic}, \textit{instance}, and \textit{panoptic} segmentation. Semantic segmentation involves classifying each pixel in an image into a predefined category, such as ``person'', ``boat'', or ``background''. Instance segmentation focuses on distinguishing between different instances of the same category, allowing for the identification and tracking of individual objects. Panoptic segmentation combines the features of both semantic and instance formulations, providing a unified approach where each pixel is assigned to a semantic label and, if applicable, an instance, offering a comprehensive understanding of the scene. 


Early works employed traditional \ac{CV} techniques to detect the elements to be considered. In this regard, the works of Sumimoto et al.~\cite{sumimoto1994machine} and Yamamoto et al.~\cite{yamamoto1999optical} analyzed the color space of the images and applied thresholding techniques to detect colors such as orange and yellow, which are commonly associated with life jackets. Although the results obtained were satisfactory, it was not a practical solution, as individuals to be rescued do not always wear life jackets. An improvement in this regard is to use multispectral and thermal images. In this context, the works by Shi et al.~\cite{shi2008architecture} and Ran and Ren~\cite{ran2010search} developed segmentation systems based on a combination of light, infrared sensors, and satellite data. These works applied processes such as background removal, thresholding, and morphological transformations, isolating targets from complex backgrounds. Following the idea of using specialized sensors, Kim and Lee~\cite{kim2014small} went a step further by proposing a clutter rejection technique specifically tailored for small targets using \acp{IR} images. This approach reduced the number of false positives with a low degradation in the detection rate, making it more practical for real-world use but at the expense of efficiency.

\begin{figure}[!ht]
    \centering
    \includegraphics[width=1.0\textwidth]{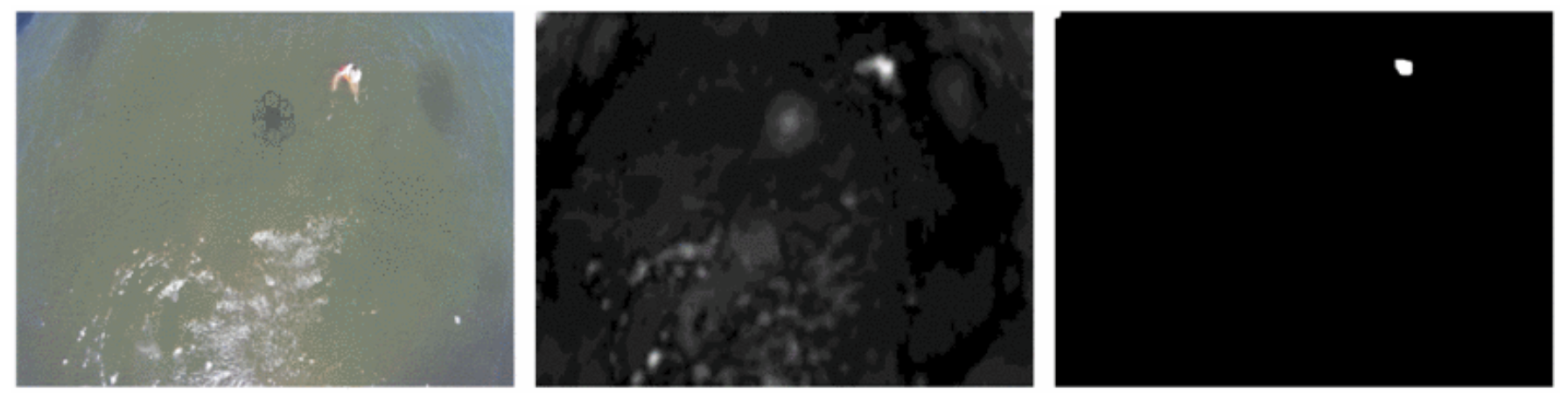}
    \caption{Example of the segmentation approach by Mendonça et al.~\cite{mendoncca2016cooperative}. The image on the left depicts the aerial image of a swimmer, the image on the center shows the saliency-based target detection, and the image on the right shows the final mask after applying the binary threshold filter.}
    \label{fig:segment_method}
\end{figure}

Another technique applied in the literature is saliency, which has been explored by several authors. Ren et al.~\cite{ren2011target} contributed with an approach based on three steps: find common patterns---typically representing the sea and the sky---from the intensity component obtained using \ac{SVD}; calculate the difference between the intensity and the common patterns; and apply saliency to detect the targets. In this work, they found that the use of the frequency domain improved detection, and in a later study~\cite{ren2012target}, the authors extended this by proposing a cumulative method that enhanced robustness through frame-by-frame saliency calculation and thresholding. Another example is the work of Mendonça et al.~\cite{mendoncca2016cooperative}, which proposed using color features, saliency, and binarization to filter and detect shipwreck survivors. The method leveraged specific characteristics, such as distinctive color and geometry, to create a mask that highlights the position of the targets, as illustrated in Figure \ref{fig:segment_method}.

The conversion to the frequency domain has also been utilized in conjunction with \ac{GMM}. Specifically, Dinnbier et al.~\cite{dinnbier2017target} proposed this combination to remove the background from images captured by a visible light camera, thereby detecting objects floating on the water's surface. 
They compared their approach with the use of \ac{FFT} and \ac{GMM} separately, finding that their method produced more stable results across all tested cases. A more recent approach was developed by Park et al.~\cite{park2023aerial}, who implemented the N-FINDR algorithm to detect small objects in maritime environments. Their algorithm used hyperspectral images and compared them with a database of similar images of ships. This comparison enabled the algorithm to filter out common patterns, thereby ignoring the ships and focusing on the detection of people and other small objects on the sea surface. The authors concluded that this approach resulted in reduced detection errors, showing the utility of such images, especially in coastal areas, where they used mannequins to simulate rescue operations.
These last two proposals were the most prominent in the field of segmentation for maritime SAR, achieving good accuracy and slightly improving efficiency with respect to previous approaches, but remaining insufficient (see Acc. and Eff. columns in Table~\ref{tab:methodology}). Precisely, due to this lack of efficiency caused by their computational cost, the deployment of these CV-based techniques in real-time contexts is limited, resulting in an almost discontinuity in the research of this type of methodology.

Approaches based on \ac{DL} for segmenting people at sea remain limited. We believe that the scarcity of modern methods in this domain can be attributed to two main factors: the difficulty of acquiring real-world maritime images and the high cost associated with pixel-wise labeling of large-scale datasets, both of which are essential for training these models. Consequently, most proposals in this area rely on object detection techniques, which require training datasets that are easier and more cost-effective to label. This is the focus of the following section.

\subsection{Object detection}
\label{sec:method:object_detection}

Object detection is a well-established process in the literature across multiple domains~\cite{xiao2020review}. This process involves identifying and locating the position of target objects within an image, typically using bounding boxes. Note that this process is closely related to segmentation. However, a fine-grained detail in the prediction is not required in this case. Instead, it is sufficient to identify the area of interest where the target object is potentially located. A summary of the methods reviewed in this section can be found in the third part of Table~\ref{tab:methodology}.

Object detection has been extensively explored in maritime SAR, employing traditional \ac{CV} methods, neural networks, and more modern approaches based on attention mechanisms and transformers. These paradigms are reviewed in the following sections.

\subsubsection{Traditional CV techniques}

The early approaches to object detection for locating people at sea primarily relied on traditional \ac{CV} algorithms. For example, Leira et al.~\cite{leira2015automatic} developed a system that combined \ac{CV} techniques. They first employed a Gaussian kernel filter to reduce noise in thermal images, followed by edge detection using the Prewitt operator. To remove the edges caused by sea waves, they applied a thresholding technique. Finally, a connected component algorithm was used to extract the contours of the targets, allowing the calculation of bounding boxes. Another example is the work by Hoai and Phuong~\cite{hoai2017anomaly}, who studied the effectiveness of anomaly detection algorithms in various SAR scenarios, including maritime contexts. They found that simulating different scenarios with multiple color spaces required distinct approaches, indicating a lack of generalization. For maritime SAR, their best results were achieved by adapting the RX detector proposed by Reed and Yu~\cite{reed1990adaptive} to this specific context. Although less common, we can also find recent works based on traditional \ac{CV}, such as the study by Zhao et al.~\cite{zhao2024heuristic}, which explored the use of heuristics. This method leveraged distinctive features inherent in maritime scenarios and utilized clustering techniques to process object sizes, proposing anchor boxes as potential target candidates. The authors concluded that while the results were competitive with recent neural networks, efficiency was lower and many challenging cases remained unresolved due to the high variability in the data.

\subsubsection{Neural Networks}

\begin{figure}[!ht]
    \centering
    \includegraphics[width=0.90\textwidth]{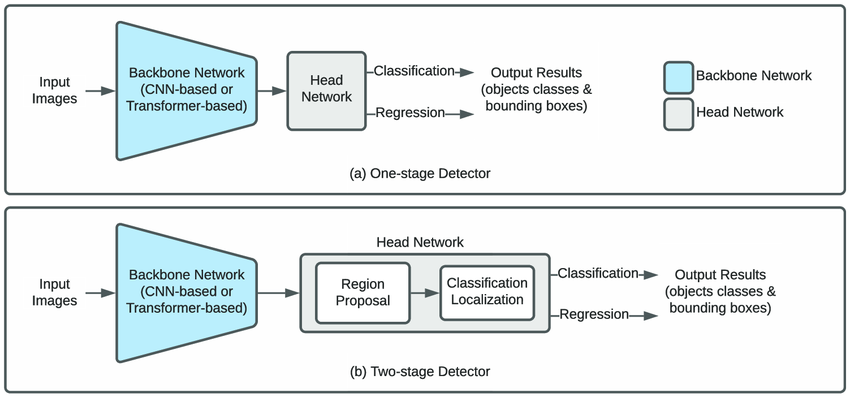}
    \caption{General pipelines for learning-based object detection methods, illustrating both one-stage and two-stage architectures. Image extracted from \cite{kang2022survey}.}
    \label{fig:od_general}
\end{figure}

Among the methods based on neural networks, there are two broad approaches---as illustrated in Figure~\ref{fig:od_general}: one-stage models that predict directly the bounding boxes and class in a single step, such as the case of RetinaNet~\cite{lin2017focal}, \ac{YOLO}~\cite{redmon2016you} and \ac{SSD}~\cite{liu2016ssd}; and two-stage models, which involve a first step to generate a set of region proposals and a second step to refine the proposals and classifying them. Examples of these models include Fast R-CNN~\cite{girshick2015fast}, \ac{FR-CNN}~\cite{ren2015faster}, and \ac{CR-CNN}~\cite{cai2018cascade}. 

\begin{figure}[!ht]
    \centering
    \includegraphics[width=1\textwidth]{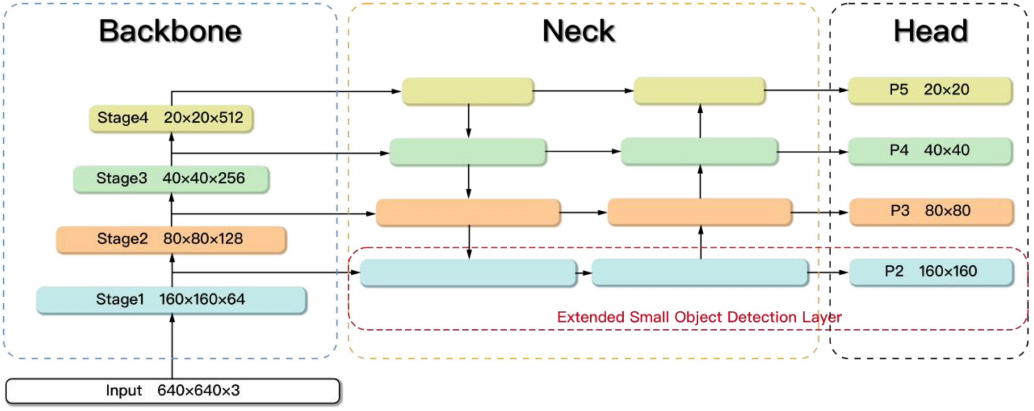}
    \caption{Improved YOLOv5 model with the small object detection layer added to extract the most superficial features~\cite{bai2022detection}.}
    \label{fig:od3_iy5}
\end{figure}

The most considered model in this context is YOLO~\cite{bochkovskiy2020yolov4}, which has been adapted with various strategies to address the challenge of detecting people at sea. Lygouras et al.~\cite{lygouras2019unsupervised} were pioneers in this area, using $9\,000$ images of swimmers captured by a drone for training a tiny YOLOv3 model to minimize resource consumption, as it had to be integrated into an embedded system with hardware limitations. This publication was an extension of the work by \cite{lygouras2018rolfer}, where GPS was used to locate the approximate area of the target. Sharafaldeen et al.~\cite{sharafaldeen2022marine} employed various YOLOv4 versions to detect people and ships in maritime environments, using visible light cameras, infrared and near-IR sensors. They concluded that the tiny version of YOLO notably increased the system's Frames per second (FPS) while only slightly reducing performance, showing its potential as a viable solution for use in UAVs. Gonçalves and Damas~\cite{gonccalves2022automatic} and Rizk et al.~\cite{rizk2022towards,rizk2022optimization} also demonstrated the effectiveness of YOLOv4 using standard RGB cameras, achieving good results and reducing the requirements needed to perform the detection task in real rescue operations. Bai et al.~\cite{bai2022detection} proposed an extension of the reduced version of YOLOv5 aimed at optimizing the detection of small objects. This was achieved by adding a specialized layer for this task consisting of the extraction of shallow features, a new feature fusion layer, and an additional prediction head, as shown in Figure~\ref{fig:od3_iy5}. Also, they proposed to reduce model complexity by replacing certain modules of YOLO with Ghost modules~\cite{han2020ghostnet}. These modifications made the architecture more precise and suitable for integration into UAVs, being the most efficient method to date while maintaining detection performance. Oda et al.~\cite{oda2023falcon} took a step further with the Falcon system, thoroughly examining the performance of YOLOv5 at varying altitudes, angles, and different levels of data compression, achieving favorable results with an image compression rate of 12.5\%. The study revealed that the central pixels of the image yielded predictions with higher confidence, which may be particularly useful in implementing a logarithmic polar compression technique. 

Despite YOLO's success in this task, the complexity of detecting small objects in such varied environments necessitates adaptations and strategies to enhance its performance. Zhang et al.~\cite{zhang2023semi} proposed a method based on semi-supervised training of YOLOv5 specifically designed for detecting people at sea and in coastal areas. The method utilizes instance segmentation to generate masks that identify small samples where objects have been detected. A semi-supervised training module then uses both real labels and pseudo-labels to train the model. Following this, \ac{MMD} and a detector are used to approximate the candidate bounding boxes and compare them with the actual bounding boxes. The experiments demonstrated highly accurate results, better than \ac{SOTA} approaches, making it a promising candidate for this task. Furthermore, the work of Fernandes et al.~\cite{fernandes2023enhancing} stands out for developing a method that produces results invariant to the drone's altitude and the angle of incidence of the images with respect to the maritime surface. To achieve this, they extended YOLOv7 by incorporating metadata and employing adversarial training to develop a robust model capable of performing well across a broader range of scenarios. The authors concluded that their method was effective and outperformed the commonly used technique of data augmentation~\cite{zhao2023yolov7}, which is typically employed to increase the variability of training data.

\begin{figure}[!ht]
    \centering
    \includegraphics[width=0.5\textwidth]{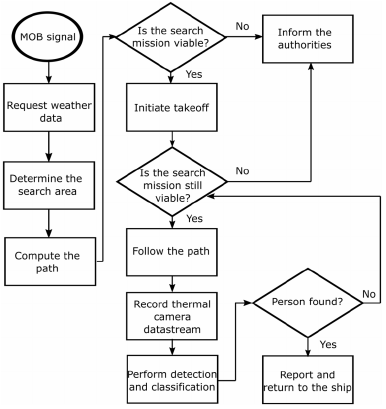}
    \caption{Flowchart of the SAR system proposed by Feraru et al.~\cite{feraru2020towards}.}
    \label{fig:od1}
\end{figure}

Other architectures have also been explored in the literature, albeit to a lesser extent. For example, the work by Gallego et al.~\cite{gallego2019detection} proposed a MobileNet-based model adapted to process multispectral images for locating people at sea. Feraru et al.~\cite{feraru2020towards} introduced a method that combines \ac{FR-CNN} with \ac{GNSS} to estimate the position of the target. The article outlines a comprehensive workflow for addressing the rescue task from autonomous UAV, as shown in Figure~\ref{fig:od1}, taking into account various factors such as weather conditions and route calculations. While the method performs effectively, the experiments revealed a significant limitation under adverse weather conditions, which remains a major challenge in this field. To improve small object detection performance, Zhou et al.~\cite{zhou2021texture} proposed a \ac{TEFF} network to enhance the feature extraction at different scales and adapt the fusion of features between layers. Its strong point was its adaptability, being tested in detectors such as \ac{FR-CNN} and RetinaNet, improving them. Bhuiya et al.~\cite{bhuiya2022surveillance} conducted a study comparing various object detectors (VGG16~\cite{simonyan2014very}, ResNet50V2~\cite{he2016deep}, InceptionV3~\cite{szegedy2016rethinking}, Xception~\cite{chollet2017xception}, and MobileNetV2~\cite{sandler2018mobilenetv2}) for locating targets at sea. Among these models, VGG16 reported the best results, making it the most suitable model among those considered. Unlike other studies, this research explored the use of a drone swarm moving according to enhance the speed of the rescue operation. They proposed to use the Particle Swarm Optimization (PSO) algorithm~\cite{al2016modified}, achieving an average duration that was three times shorter compared to other algorithms such as Grey Wolf Optimization~\cite{mirjalili2014grey} and Bat Optimization~\cite{yang2010new}. Another notable work is that by Moosbauer et al.~\cite{moosbauer2019benchmark}, who proposed an algorithm that automatically generates segmentation labels from previously annotated bounding boxes to train a \ac{MR-CNN}~\cite{he2017mask}. Utilizing the Mahalanobis distance between pixels within each bounding box and the mean intensity of the image (which represents the sea or sky), the authors achieved improved object detection results compared to \ac{FR-CNN}, demonstrating the effectiveness of their approach.

Cafarelli et al.~\cite{cafarelli2022mobdrone} explored the potential of using the task-aligned one-stage object detection model~\cite{feng2021tood}, VarifocalNet~\cite{zhang2021varifocalnet}, and \ac{MR-CNN}~\cite{he2017mask}, among other models previously studied. Their experiments investigated the impact of drone altitude, providing stable results at higher altitudes with VarifocalNet.
Another option considered in the literature is to combine object detectors that work together to produce a final result. In this context, Gasienica-Jozkowy et al.~\cite{gasienica2021ensemble} proposed an ensemble method based on a fusion of different object detection architectures, which were optimized by applying weights to control the contribution of each model to the final result. This method was able to improve upon standard object detection models, obtaining the best results with a combination of \ac{FR-CNN}~\cite{girshick2015fast} and RetinaNet~\cite{lin2017focal}.

\subsubsection{Attention mechanisms and Transformers}

Current trends highlight the use of visual attention mechanisms and transformer-based approaches, which have shown effectiveness in various detection tasks. As seen in the contest organized at the Workshop on Maritime Computer Vision~\cite{kiefer20231st}, several relevant approaches utilizing this technology stand out. The best-performing method, called Maritime-VSA, utilizes the \ac{VSA} module~\cite{zhang2022vsa}, which enhances the detection of targets in high-resolution images by employing a multi-scale process. The second-best result was achieved by a model based on the DetectoRS architecture~\cite{qiao2021detectors}. Trained on multiple maritime datasets, this model achieved performance that was very close to that of Maritime-VSA. Another well-performing method was DyHead~\cite{dai2021dynamic}, which employs attention mechanisms to unify localization and classification heads while leveraging test-time augmentations to enhance the model's robustness. In addition, Zhao et al.~\cite{zhao2023yolov7} presented the YOLOv7-sea method by using a new prediction head and an attention module (called \ac{SimAM}) to detect tiny-scale people and find regions of interest. Zhu et al.~\cite{zhu2023yolov7} also included this module in a new method called YOLOv7-CSAW, adding the K-means++ algorithm for optimal prior anchor box size, the C2f module for a lightweight model, and an \ac{ASFF} to enhance high-level semantic features in small targets. They also implemented the wise \ac{IoU} (WIoU) loss function to deal with large positioning errors and missed detections, reducing false negatives.

Beyond this competition, several methods have also emerged that stand out for incorporating attention mechanisms. Gao et al.~\cite{gao2022multiscale} proposed a model based on multiscale attention, featuring two main modules: \ac{MAEM} and \ac{MFFM}. These modules work together to focus attention on relevant features across different scales, thereby handling the complexity of detecting small objects in diverse maritime environments. Yang et al.~\cite{yang2023high} introduced an extension of YOLOv5 called YOLO-BEV that incorporates a bidirectional feature fusion module inspired by the work by Liu et al.~\cite{liu2018path}, along with a head designed for the prediction of extremely small objects. They proposed the C2fSESA (Squeeze-and-Excitation Spatial Attention) module to enhance feature extraction, thereby improving the detection capabilities over the original YOLOv5. Furthermore, this implementation proved to be the most accurate and efficient of all, being the most suitable for deployment in real scenarios. Wang et al.~\cite{wang2023sea} also utilized YOLOv5 as the base model, integrating the \ac{CBAM} to identify areas of attention within high-resolution images. The use of \ac{CBAM} allowed for better focus on critical regions in complex maritime scenes, thereby improving detection performance in scenarios where high-resolution imagery is essential. Shi et al.~\cite{shi2024mtp} took the approach further by adopting a more recent version of YOLO (the ninth) and proposed the MTP-YOLO detector, which includes an \ac{MCE-CBAM} to focus the model's attention on areas where objects are likely to be located. Additionally, they introduced a cross-stage partial layer with \ac{C2fELAN}, designed to preserve key features that represent people in the images while maintaining a low number of parameters to reduce complexity. Zhang et al.\cite{zhang2023sg} proposed a detector named SG-Det (Shuffle-GhostNet Detector) based on GhostNet~\cite{han2020ghostnet} which meets the requirements of high precision and high-speed detection using a lightweight feature fusion architecture named \ac{BiFPN}-tiny and incorporated the \ac{ASPP} module into the network to obtain multi-scale information. 
Zhang et al.~\cite{zhang2023enhanced} (different authors than the previous work) incorporated the BiFormer attention module to enhance the small-scale target detection in a new algorithm called ABT-YOLOv7 which integrated an \ac{AFPN} to conserve the target feature information. The authors highlight the sensitivity of the model in detecting objects under complex conditions, such as low-light scenarios and cases with reflections, increasing its utility in realistic environments.

\begin{figure}[!ht]
    \centering
    \includegraphics[width=0.75\textwidth]{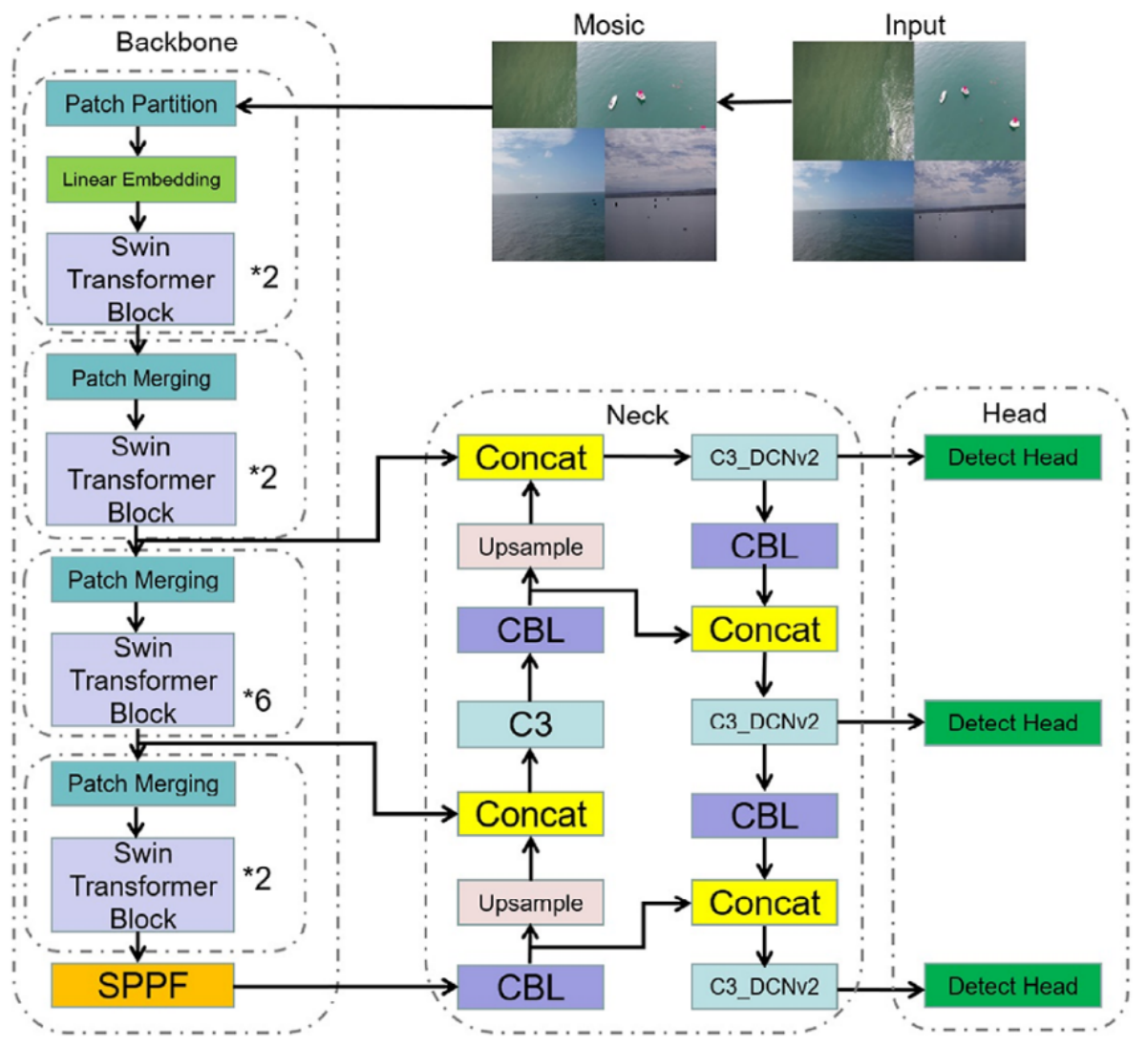}
    \caption{Flowchart of the SAR system proposed by Liu et al.~\cite{liu2024yolov5s}.}
    \label{fig:yolov5_swint}
\end{figure}

Due to their popularity in recent research across various domains, transformer-based models have also been adapted for maritime SAR tasks. Specifically, the approaches found in the literature are usually based on \ac{DETR}~\cite{carion2020end}, \ac{Swin-T}~\cite{liu2021swin}, and \ac{ViT}~\cite{dosovitskiy2020image}. Cafarelli et al.~\cite{cafarelli2022mobdrone} were pioneers in attempting to employ a transformer-based model for detecting people at sea, considering a \ac{DETR}~\cite{zhu2020deformable}. Their study concluded that the model was suitable when dealing with images captured at low altitudes, where the targets were sufficiently large. However, at higher altitudes, \ac{DETR} was outperformed by other \ac{SOTA} models such as VarifocalNet~\cite{zhang2021varifocalnet}. Another architecture is \ac{Swin-T}, which has been widely used in several works. Liu et al.~\cite{liu2024yolov5s} implemented YOLOv5s-SwinDS, incorporating a \ac{Swin-T}-based module in the backbone instead of the YOLOv5s one, introducing multi-level feature fusion characteristic. This adjustment can be observed in Fig.~\ref{fig:yolov5_swint}, where we can also notice that utilized Deformable Convolutional Networks v2 (DCNv2)~\cite{zhu2019deformable} instead of traditional convolutions in the feature output, which improved recognition capabilities for irregular targets. Wang et al.~\cite{wang2023sea} also combined this architecture with YOLOv5s in their method called Sea-YOLOv5s model, but in this case introducing it into the small object detection head, enhancing the ability of the YOLOv5 model to recover abundant contextual information. Other approaches that used the \ac{Swin-T} as backbone in combination with attention modules for their proposals were Gao et al.~\cite{gao2022multiscale}, Maritime-VSA~\cite{kiefer20231st}, and DyHead~\cite{dai2021dynamic}, both discussed above. Finally, we identified only the approach by Lu et al.~\cite{lu2023high} based on \ac{ViT}, which combines this architecture with DINO~\cite{zhang2022dino} to develop a hybrid backbone built upon \ac{SIMMIM}. By using denoising methods and an efficient optimizer during training, they obtained better performance than \acs{SOTA} methods with fewer parameters.

This emerging trend toward the use of transformers has shown highly accurate results. However, these models have significant computational costs (see Acc. and Eff. columns in Table~\ref{tab:methodology}) and are exceedingly large, requiring substantial storage space. These factors make them impractical for deployment on embedding systems in UAVs.

\subsection{Tracking}
\label{sec:method:tracking}

\begin{figure}[!ht]
    \centering
    \includegraphics[width=0.99\textwidth]{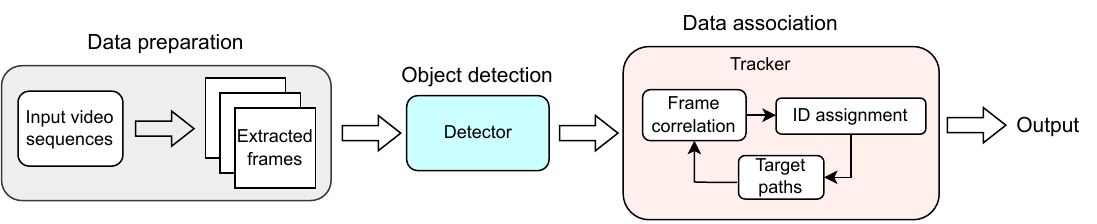}
    \caption{General pipeline of the tracking process. The input video sequence is divided into frames, which are processed by a detector to generate predictions. The tracker then establishes correspondences between frames, assigns IDs to each object, and predicts trajectories to maintain consistency over time.}
    \label{fig:tracking_scheme}
\end{figure}

Another area of significant interest in the field of maritime SAR is tracking, which focuses on continuously monitoring and following targets over time. This is crucial in dynamic environments, such as the ocean, where conditions can rapidly change, and the precise location of people or objects can shift due to currents, waves, or other factors. Tracking ensures that once a target is identified, it can be monitored effectively, enabling timely rescue operations and reducing the risk of losing sight of it. To address this task, approaches must work with video sequences from which input frames are extracted. Then, a detector is used to retrieve the position of bodies in each frame, followed by a tracker that extracts correspondences between frames, assigns IDs to each object, and predicts its trajectory over time. The general scheme of this process is shown in Figure~\ref{fig:tracking_scheme}, and an overview of the approaches developed for person tracking at sea is presented in the fourth section of Table~\ref{tab:methodology}.

Early tracking models relied on basic mathematical and statistical methods, as well as traditional computer vision techniques, to compare information between consecutive frames and track targets. For example, Westall et al.~\cite{Westall2008} elaborated a study in which they evaluated four different color spaces (CS), four point-target detection techniques, and two temporal tracking methods: Dynamic Programming (DP) and \ac{HMM}. 
The same authors improved this work by incorporating a data fusion algorithm to enhance the detection process~\cite{Westall2009}, sourcing multiple data sets from different video color spaces and combining them. In contrast, Kaiyu and Chaojian~\cite{kaiyu2009vision} were based on the characteristics of infrared frames and the sea environment, combining the approaches of frame difference and target region growth to track targets quickly and accurately. Leira et al.~\cite{leira2015automatic} proposed a tracking algorithm combining the Kalman filter with a simple linear motion model and a global nearest-neighbor algorithm in thermal images, reporting that the system successfully tracked targets but failed to achieve real-time performance.

More recent studies rely on more modern \ac{DL} techniques for tracking. This advancement has been crucial to developing more generalizable models, a feature particularly needed in the maritime environment, where scene conditions are constantly changing. Kiefer and Zell~\cite{kiefer2023fast} implemented a real-time end-to-end future frame prediction autoencoder to identify regions of interest in a video stream, speeding up the process and demonstrating that these systems are a promising avenue even in an environment of limited resources. Kiefer et al.~\cite{kiefer2023memory} built temporal memory maps, adding metadata without causing a large computational overhead. They applied this proposal to the output of a YOLOv7-Tiny for video object detection, and they then employed the method with DeepSORT and \ac{ReID} module for tracking people in video sequences. Yang et al.~\cite{yang2024sea} presented the Metadata Guided MOT (MG-MOT) algorithm, an adaptable motion-based multi-object tracking method with \ac{ReID} that combined short-term tracking data into coherent long-term tracks, leveraging relevant metadata from UAVs, such as GPS position, drone altitude, and camera orientation. As detectors, they used different backbones of YOLOv8, achieving significantly improved performance compared to \ac{SOTA} methods.

Zhang et al.~\cite{zhang2023lightweight} demonstrated that attention mechanisms are also effective for tracking, introducing the \ac{SimAM} module into their YOLOv7-FSB method to focus attention on areas with swimmers. They also included ByteTrack for real-time tracking, FSNet as the backbone, SP-ELAN module, and a \ac{BiFPN} to accelerate the process for integration into embedded systems on UAVs. The high accuracy of this detector, combined with its low storage requirements due to a reduced number of parameters, resulted in excellent performance when coupled with the tracker, yielding promising results.

The use of transformers has also been explored in this field, although to a lesser extent. These models are capable of handling long sequences and capturing long-term dependencies, making them seemingly well-suited for tracking tasks~\cite{li2022long}. However, despite their potential, the implementation of transformers in this domain requires a substantial amount of training data and computational power, which can be challenging in real-time operations. Thus, we only found the proposal implemented by Huang et al.~\cite{huang2023reidtracker}, called ReIDTracker\_Sea, which employs a \ac{Swin-T} backbone integrated with the Composite Backbone Network V2 (CBNetV2) architecture for providing high-quality instance boxes. They carried out multi-object tracking in maritime UAVs scenarios in a completely unsupervised way, achieving satisfactory results but low performance in terms of speed.

\subsection{Synthetic data generation}
\label{sec:method:synthetic}

The works discussed above represent effective strategies for the search and rescue of people at sea from UAVs. However, they have limitations, such as the scarcity of real-world data to train learning-based models and the low variability of these data, as we will see in Section~\ref{sec:datasets}. The challenge arises from the high cost of deploying resources to collect such data, along with the need for licenses and specialized equipment, which makes gathering large volumes of training data very difficult. To mitigate this problem, one possibility is to generate synthetic data, which is cheaper, much simpler, and more accessible. This section covers synthetic data generation methods, summarized in the fifth section of Table~\ref{tab:methodology}. These methods play a key role in current methodologies that aim to produce more robust and generalizable models. 

\begin{figure}[!ht]
    \centering
    \begin{subfigure}{0.2\textwidth}
        \includegraphics[width=1\textwidth, height=2cm]{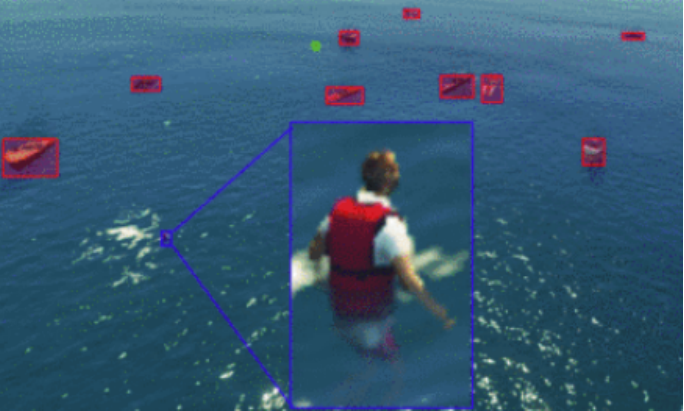}
        \caption{DeepGTAV~\cite{kiefer2022leveraging}.}
        \label{fig:syn1}
    \end{subfigure}
    \begin{subfigure}{0.2\textwidth}
        \includegraphics[width=1\textwidth, height=2cm]{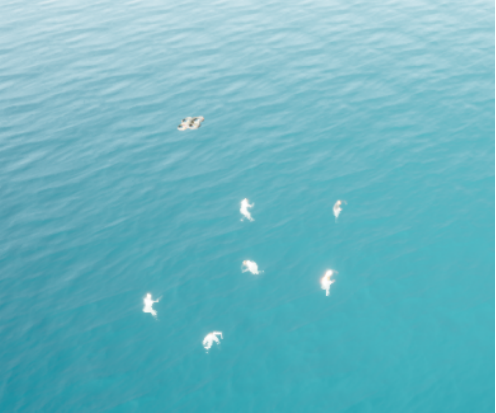}
        \caption{AirSim~\cite{poudel2023novel}.}
        \label{fig:syn2}
    \end{subfigure}
    \begin{subfigure}{0.2\textwidth}
        \includegraphics[width=1\textwidth, height=2cm]{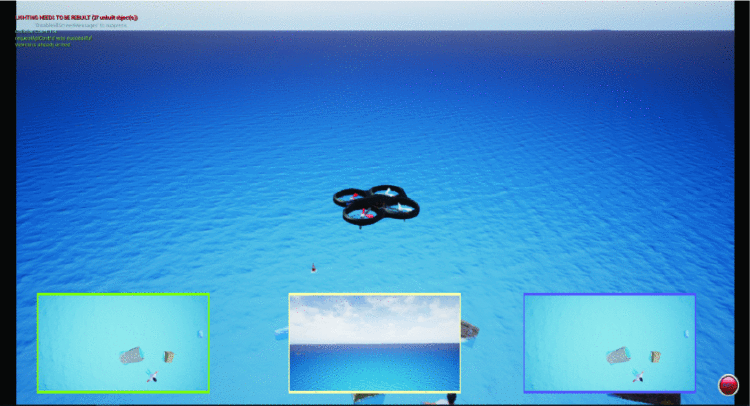}
        \caption{Environment by~\cite{do2021novelty}.}
        \label{fig:syn3}
    \end{subfigure}
    
    \begin{subfigure}{0.3\textwidth}
        \includegraphics[width=1\textwidth, height=2.5cm]{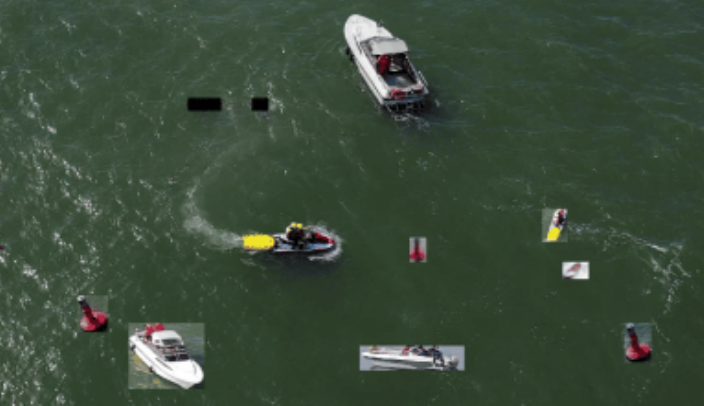}
        \caption{Poseidon tool~\cite{ruiz2023poseidon}.}
        \label{fig:syn4}
    \end{subfigure}
    \begin{subfigure}{0.3\textwidth}
        \includegraphics[width=1\textwidth, height=2.5cm]{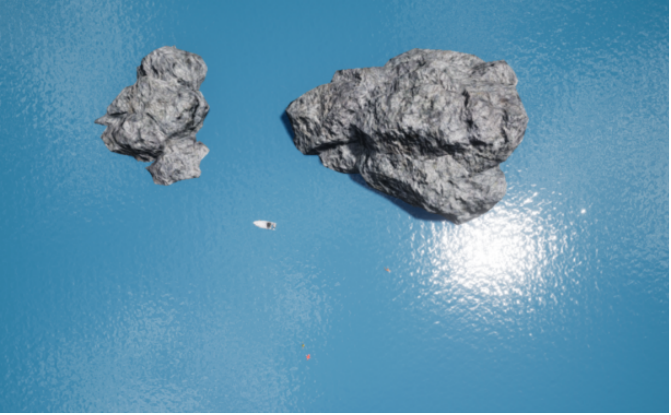}
        \caption{Synthetic image from ~\cite{martinez2025use}.}
        \label{fig:syn5}
    \end{subfigure}
 \caption{Visual examples of virtual maritime environments and synthetic images generated for SAR tasks at sea from an aerial perspective.}
 \label{fig:ejemplos_syn}
\end{figure}

To generate synthetic data, one option is to use graphical modeling tools. For instance, Usilin et al.~\cite{usilin2020training} proposed creating a completely synthetic dataset to detect inflatable life rafts at sea---instead of people---generated using Blender.\footnote{\url{https://www.blender.org/}} These data were tested with the Viola-Jones detector~\cite{viola2004robust}, yielding favorable results in real-world data despite the gap between the two distributions. Also with Blender, Usilin et al.~\cite{lin2023seadronesim} presented the SeaDroneSim environment, which provided ground truth for segmentation tasks. They trained a \ac{FR-CNN} model with only synthetic data and obtained successful results also evaluating in real scenarios. 
In turn, the 3D computer graphics game engine \ac{UE}~\footnote{\url{https://www.unrealengine.com}} has also been used to create realistic scenes. For example, Do et al.~\cite{do2021novelty} built a simulated marine environment, but in this case including GPS localization from UAV (see Figure~\ref{fig:syn3}). They tested their synthetic data with \ac{FR-CNN} achieving good accuracy in the same distribution; however, the method performed poorly if the victim moved away from the UAV, and the gap with the real world was still a long way off. Martinez et al.~\cite{martinez2025use} built a more realistic environment based on \ac{UE}, generating a new completely synthetic dataset called SynBASe (see Fig.~\ref{fig:syn5}) with elements like rocks and adverse weather conditions (sunset, fog, and rain). They tested their dataset with YOLOv5, combining synthetic and real data during training, and demonstrated the effectiveness of this strategy.

Another option for this task is the use of simulators, such as AirSim\footnote{\url{https://microsoft.github.io/AirSim/}}, an open-source, cross-platform simulator for drones and ground vehicles, among others, which has been used for image generation in a few research projects. The work of Do et al.~\cite{do2021novelty} made a first attempt to simulate the movement of the UAV in the environment mentioned above. This was later improved by Poudel et al.~\cite{poudel2023novel}, who developed a dynamic simulation environment with variable conditions by integrating AirSim with \ac{UE} and \ac{ROS}. This enabled the control of a UAV using manual commands, while detecting objects with high accuracy using a pre-trained YOLOv7 model (see Figure~\ref{fig:syn2}). However, they did not conduct tests with real images and the flight altitude in the experiments was only 8 meters.

Another method we identified for generating synthetic data is to capture images from realistic video games through the development of frameworks. Yun et al.~\cite{yun2019small} captured images by scene recording using the 3D game editor ARMA 3~\footnote{\url{https://github.com/cloftus96/Synthetic-Data-Generation}}. By employing \ac{SSD}, they slightly improved the results when testing with real images, but they were still affected by light, obstructions, and waves. The DeepGTAV framework presented by Kiefer et al.~\cite{kiefer2022leveraging} was used to build a new synthetic dataset from GTA V videogame, called DGTA-SDS (see Figure~\ref{fig:syn1}). They trained the EfficientDet-D0 and YOLOv5 detectors (pre-trained on synthetic data and fine-tuned for real data) and concluded that synthetic data can improve the results compared to training with real data alone, but the performance is still not competitive. 

Finally, several methods have been developed to create semi-synthetic aerial images of maritime environments by combining real data with synthetic elements (such as bodies, ships, rocks, or even appearance alterations). Ruiz et al.~\cite{ruiz2023poseidon} created a semi-synthetic data augmentation tool to address the class imbalance problem. They took cutouts of the annotations from the minority classes and pasted them onto different images. As shown in Figure~\ref{fig:syn4}), the coherence between the annotations and the background was poor. Nevertheless, they tested their approach using YOLO models, which improved results for the underrepresented classes. To solve the problem of coherence, Mohammadi et al.~\cite{khan2024investigating} used U2-Net~\cite{zivkovic2004improved} to remove background in annotations, contributing with a new data augmentation framework. They demonstrated that adding synthetic data to a limited number of real images was suitable for the task when using YOLO. On the other hand, Tran et al.~\cite{tran2024safesea} focused on the background images, developing the SafeSea framework with \ac{BLD}~\footnote{\url{https://github.com/omriav/blended-latent-diffusion}}, a tool that allowed the transformation of actual sea images in new sea state backgrounds by specifying a mask and text description while keeping maritime objects. By training a YOLOv5 model, they observed that the model performed worse in synthetically generated adverse sea scenarios. However, it lacked control during the editing process, which limited background diversity and imposed constraints on generating realistic waves.

\subsection{General detection approaches applied to maritime SAR}

The literature includes methods that are not specifically designed for maritime SAR but have been evaluated for this task to assess their generalizability. This section discusses these cases, which are summarized at the bottom of Table~\ref{tab:methodology}.

The fifth version of YOLO was employed in some works that evaluated versatility and applicability across land and sea, achieving high accuracy. For example, Zhang et al.~\cite{zhang2022finding} employed a YOLOv5-based DCLANet model along with an image cropping strategy that prioritized accuracy over speed, or Wang et al.~\cite{wang2022remote} who contributed with a multi-class Super-Resolution Generative Adversarial Network that supports YOLOv5. Also using this version, but proposing a feature fusion method, we find LAI-YOLOv5s~\cite{deng2023lightweight} with a new large-size detection head and an upgraded backbone network, or SF-YOLOv5~\cite{liu2022sf} based on PANet~\cite{liu2018path} and BiFPN~\cite{tan2020efficientdet}, both being lightweight. Similarly, but with the YOLOX version, Chen at al.~\cite{chen2023yolox} developed the TO–YOLOX method, including the Group-CBAM module to enhance the perception of tiny objects in remote sensing images. However, its weakness resided in the fact that it was not always fully suitable for scenarios containing numerous large and small objects. Zeng et al.~\cite{zeng2023yolov7} proposed an improvement of YOLOv7, named YOLOv7-UAV, introducing the DpSPPF module and a binary K-means anchor generation to extract feature information across different scales. However, it had limitations such as optimization and performance under adverse weather conditions. The TOD-YOLOv7 approach~\cite{tang2023long} also introduced an improvement for YOLOv7 by adding a tiny object detection module and a coordinate attention mechanism to detect people on the beach, using a deep convolution module on the GPU for a simplified implementation that achieves high efficiency.

Fast R-CNN has also been utilized in several studies. Although it is not as efficient or accurate as YOLO, it still performs the task adequately. Hu et al.~\cite{hu2023fspn} proposed Faster R-CNN-FSPN, which optimizes the process through feature reconciliation and spatial context feature selection. However, it struggled to detect individuals affected by external factors such as occlusions. The same issue occurred to Hong et al.~\cite{hong2021sspnet}, who built the SSPNet approach upon this architecture, applying a context attention module, a scale enhancement module, and a scale selection module. Shao et al.~\cite{shao2024small} reduced the missing rate with the FG-RCNN method, based on feature interaction and guided learning. In the same line and to reduce false positives, Chen et al.~\cite{chen2023graphormer} proposed a Graphomer-based Contextual Reasoning Network (GCRN) with a context relation module and a context reasoning module that enhanced the feature expression of small objects. 

Other studies opted to use less common architectures but still achieved favorable results. Guo et al.~\cite{guo2023save} proposed the HANet model based on CenterNet, which alleviated misalignment between feature subspaces and detection subspaces. Another example is the Swin-T-EFPN~\cite{zhang2023efpnet} method, which used a Multi-Dimensional Attention Module (MDAM) to generate attention maps and a feature fusion module that maintained consistency between deep and shallow features. Qi et al.~\cite{qi2022small} improved speed by proposing SODNet, a model that integrates specialized feature extraction and information fusion techniques. 


\subsection{Related research lines}
\label{sec:method:related_works}

In addition to the typical tasks associated with maritime SAR, the literature includes several noteworthy related works. This section discusses some of these research lines, including ship detection and tracking, marine life detection, and SAR applied to land-based operations. These works are included in this review because their proposals may be directly applicable to the detection of individuals at sea, offering valuable ideas and insights to consider.

\subsubsection{Ship detection and tracking}
\label{sec:method:related_works:ship_detection_tracking}

Just as there is literature on the detection and tracking of individuals at sea, there are also works focused on these same tasks for ships. The main motivations behind this area include traffic monitoring, security, and accident prevention~\cite{huang2020ship}. Among the works that address this topic, one notable example is by Ribeiro et al.~\cite{ribeiro2017data}, which presents a dataset of images captured from a UAV in a top-down view over the open sea, featuring ships in various positions. As a reference, they employed various general-purpose trackers---not specific for maritime scenarios---to compare their results with proposals based on correlation filters, the Fourier domain, and Otsu's algorithm to separate bright and dark areas. Shao et al.~\cite{shao2018seaships} included images taken from \ac{USV} for obstacle avoidance. They employed well-known architectures such as VGG-16, ResNet, InceptionV3, \ac{FR-CNN}, YOLO, and SSD, which have been also explored for detecting people at sea (see Section~\ref{sec:method:object_detection}). Gallego et al.~\cite{gallego2018automatic} explored the use of U-Net, an architecture not addressed in the detection of people at sea, but which would be interesting to explore to extract the segmentation mask of swimmers at the pixel level. An enhanced \ac{MR-CNN} model~\cite{he2017mask} was implemented by Nie et al.~\cite{nie2020attention}, extracting both the bounding boxes and the segmentation mask of ships.

Detecting and tracking ships are also useful in combating illegal activities, such as smuggling, unregulated dumping of pollutants in fisheries, and the unauthorized crossing of international boundaries by unidentified vessels. This interest has promoted the development of comprehensive automated systems suitable for the task. For example, we can find approaches based on AlexNet~\cite{ma2019maritime}, SSD~\cite{ghahremani2018cascaded}, YOLO~\cite{ch2023classification,pobar2023yolov7}, and U-Net~\cite{ch2023classification}, and trackers such as Kalman filters in~\cite{fefilatyev2012detection}, KCF~\cite{henriques2014high} in \cite{su2024integrated}, or DeepSORT in ~\cite{han2024detection}, similar to those used for tracking of people at sea in Section~\ref{sec:method:tracking}.


\subsubsection{Marine animal detection}

Another field of interest with common characteristics to the topic addressed in this work is monitoring marine animals. Nowadays, technological advances are facilitating new non-invasive approaches to detecting, monitoring, and assessing wildlife in various ecological settings. Those missions involve flight sessions at certain altitudes over the sea, in which qualified environmental experts are able to remotely observe the marine animals that emerge on the water surface. Among the applications, we can find marine animal detection, species identification, tracking behavior analysis, anomaly detection, or real-time monitoring~\cite{dujon2021machine,berg2022weakly,maire2014convolutional}. There are also studies for detecting specific species, such as sharks~\cite{sharma2022sharkspotter}, sea turtles~\cite{gray2019convolutional}, white dolphins~\cite{zhang2021chinese,chien2022study}, or whales~\cite{gaur2023whale}. The images used in these cases match in features with the task of detecting people at sea, with the main difference of the target. Therefore, we can find common solutions in both research lines, tending in the last decade towards deep learning models such as YOLO, \ac{FR-CNN}, or U-Net. As with the task at hand, collecting data to train these models is costly and requires careful planning, licensing, and collaboration among experts, UAV operators, and data scientists to guarantee ethical and effective monitoring. Therefore, these data sources could be used to extend existing training sets with samples that introduce greater variability and enhance the generalization capacity of the models. For a more detailed analysis, the reader is referenced to the following sources~\cite{nandi2024advances,verfuss2019review}  

\subsubsection{SAR on land}

SAR operations on land present challenges similar to those in maritime scenarios, requiring efficient and effective strategies for locating and rescuing individuals in danger. Traditional methods relied on ground-based teams, canine units, and aerial manned aircraft which processed images using simple techniques based on color segmentation~\cite{turic2010two}. However, advancements in technology have revolutionized these missions. An example in this field is the detection of people in urban environments, where high population density and obstructive elements complicate the task. However, the methods used are very similar to those applied in maritime SAR. Examples include the use of standard object detection models such as SSD~\cite{mishra2020drone}, \ac{FR-CNN}~\cite{sambolek2020person}, YOLOv3 with thermal images~\cite{dong2021uav}, the integration of GPS devices~\cite{murphy2019autonomous}, or even motion recognition techniques~\cite{geraldes2019uav}.

In contrast to urban scenarios, the SAR missions in wild environments involve flying over expansive regions of forests~\cite{tusnio2021efficiency}, mountains~\cite{niedzielski2021first}, snow-covered regions~\cite{bejiga2017convolutional, sambolek2021automatic}, flooded areas~\cite{hasan2021search}, tsunamis~\cite{koshimura2020tsunami}, or deserts~\cite{horyna2023decentralized}, which require extended flight times and thus have more similarity to maritime SAR. Also in these scenarios, similar solutions are proposed to the ones reviewed for maritime SAR such as the use of GPS, Custom CNNs, YOLO, RetinaNet, \ac{CR-CNN}, and \ac{FR-CNN}.
For a more in-depth analysis of this topic, readers are referred to the following reviews~\cite{bai2023review, cazzato2020survey, lyu2023unmanned}.

\section{Datasets}   
\label{sec:datasets}

In this section, we outline publicly available datasets containing images captured by UAVs that may serve as benchmarks for different SAR tasks. 

\begin{figure}[!ht]
    \centering
    \includegraphics[width=1.0\textwidth]{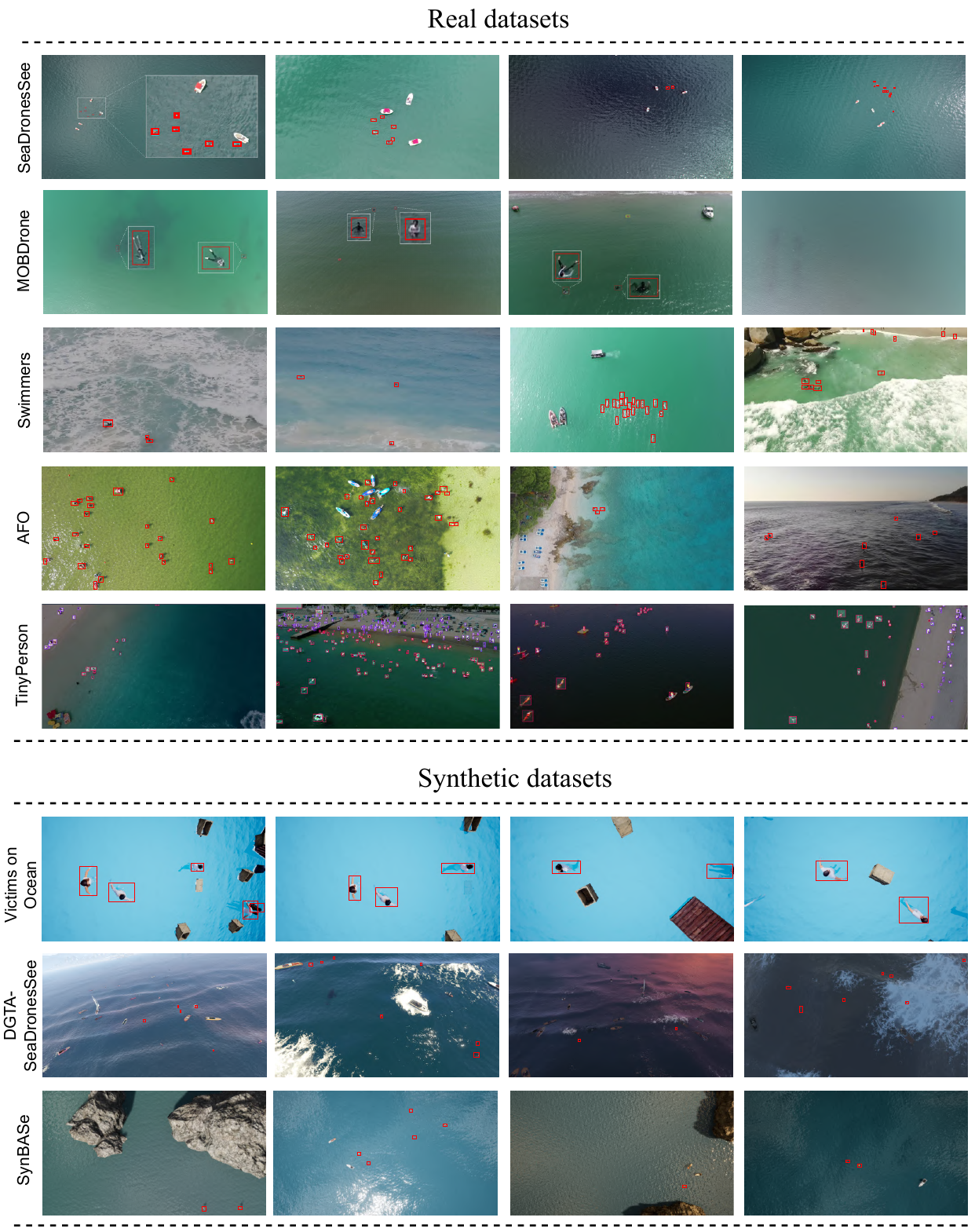}
    \caption{Representative image examples from the datasets designed for detecting people at sea, whose specifications can be found in Table~\ref{tab:datatable_sea}. Annotations of humans are marked in red.}
    \label{fig:datasets_ejemplos}
\end{figure}

Table~\ref{tab:datatable_sea} summarizes the reviewed datasets and their most relevant characteristics. Four representative examples from each dataset are shown in Fig.~\ref{fig:datasets_ejemplos}, illustrating the wide variability in these images, which reflects the inherent challenges of the task. Notable differences include altitude, perspective, lighting conditions, reflections, waves, and shadows on the sea surface, along with a common characteristic: the relatively small size of people in proportion to the entire image. Below, we provide a brief description of each dataset. 

\input{table_corpora}

\textbf{SeaDronesSee} is a dataset proposed by Varga et al.~\cite{varga2022seadronessee} that consists of over 54k high-resolution images captured by UAVs over open sea. This dataset is designed for object detection and tracking tasks, containing a set of object classes: swimmers, floaters, boats, buoys, and life jackets. The images were captured with different cameras at various altitudes, positions, and inclinations. Additionally, images were taken at different times of the day to introduce variations in lighting conditions. 

The Man OverBoard Drone benchmark (\textbf{MOBDrone})~\cite{cafarelli2022mobdrone} is a large-scale drone-view dataset for detecting people overboard. It contains $126\,170$ frames extracted from 66 video recordings collected from a UAV at altitudes ranging between $10$ and $60$ meters, approximately perpendicular to the sea surface, with some exceptions captured at $45\degree$ of inclination. The images depict variations in light conditions and were manually annotated to locate people, boats, surfboards, buoys, and wooden objects. It also contains background images of clear water without annotations.

\textbf{Swimmers} is a dataset created by Lygouras et al.~\cite{lygouras2019unsupervised} composed of images collected from different internet sources. The authors also included images collected by manually flying a drone equipped with a GoPro camera. These images were obtained with different angles and lighting conditions and contain humans swimming in open water and a few boats. The images differ in terms of the number of people in the scene, human characteristics, image resolution, context, illumination, and composition.

The Aerial Floating Objects dataset (\textbf{AFO})~\cite{gasienica2021ensemble} comprises $3,647$ images extracted from aerial drone videos captured by different drone-mounted cameras, aimed at detecting people and floating objects in the water. Some videos were recorded in controlled scenes, including people acting in various pre-established ways (swimming techniques, drifting the head or the back in the water, etc.), while other recordings were contributed by volunteer photographers. The majority class is the category \texttt{person}, accounting for over 80\% of all annotations, but there are also annotations for boats, boards, buoys, sailboats, and kayaks. To address the issue of class imbalance, two additional versions were released: one with classes grouped into two super-categories (small and large objects), and another with a single category (objects related to humans).

\textbf{TinyPerson}~\cite{yu2020scale} consists of a collection of real-world videos, subsampled every 50 frames and filtered to remove the images with high homogeneity. Some of the most relevant aspects are the tiny size of people, the large variance of their aspect ratio, the focus on people around the seaside, and the presence of images with a high density of individuals (over 200 people). The dataset includes annotations of bounding boxes and pixel sizes, differentiating among people at sea or land, hard-to-recognize persons, and regions to be ignored, such as crowds, ambiguous areas, or reflections on the water. A second version of this dataset is \textbf{SeaPerson (TinyPersonV2)}~\cite{yu2022object}, which extends the number of images to about seven times, keeping the main characteristics of TinyPerson. On average, the size of people is extremely small, approximately 22 pixels, making their detection a challenging task.

As observed, there are not many publicly available datasets for maritime SAR, primarily due to high deployment costs and the need for specialized equipment that may not be accessible to everyone. This scarcity highlights the need for more data collections, as models may face significant variability during inference. In this context, the use of synthetic data appears to be a promising solution. A few datasets have been generated through synthetic simulations using computational tools to complement the limited amount of real data available. Below, we describe the published synthetic datasets.

\textbf{Victims on Ocean}~\cite{do2021novelty} is a synthetic collection of images developed using the AirSim simulator in conjunction with \texttt{Unreal Engine 4}. It simulates a UAV equipped with a camera, flying at different altitudes above the sea and capturing images with the lens pointed perpendicular to the water's surface. The scenes contain simulations of victims and floating objects, but the dataset only includes annotations for the locations of people.

\textbf{DGTA-SeaDronesSee}~\cite{kiefer2022leveraging} is another synthetic dataset created with the DeepGTAV framework to emulate UAV views of maritime scenes. It contains large-scale, high-resolution images in several domains. The camera altitude and angles were varied and the position of the objects was randomly assigned. For each frame, the corresponding ground truth, bounding boxes, and meta-labels were generated including information such as altitude, camera rotation angles, time of the day, and weather conditions.

The Synthetic Bodies At Sea (\textbf{SynBASe}) dataset~\cite{martinez2025use} is another collection of synthetic images focused on maritime scenarios for SAR tasks from UAVs. It was developed using an automatic data generation method based on \texttt{Unreal Engine 4}, including simulated adverse conditions such as sunset, fog, rain and solar reflections. The dataset includes the annotation of a single category labeled ``swimmer''.

\section{Evaluation metrics} %
\label{sec:metrics}

This section discusses the evaluation metrics commonly used in articles that address maritime SAR tasks. We will organize these metrics based on their main usage: general purpose, object detection, and tracking. Note that some metrics may be used for different tasks, these will be categorized as general purpose.

\subsection{Efficacy metrics}
\label{sec:metrics:efficacy}

\subsubsection{General-purpose metrics}
\label{sec:metrics:efficacy:general}


One way to evaluate model performance is by counting the number of correct and incorrect predictions compared to the expected results. In general, precision (P) and recall (R) (also known as sensitivity) are two widely used metrics for assessing model performance in various classification and detection contexts. These metrics can be formulated as: 

$$\text{P}=\frac{\text{TP}}{\text{TP+FP}}, 
\quad 
\text{R}=\frac{\text{TP}}{\text{TP+FN}},$$

\noindent where TP, FP, and FN represent the number of correct predictions for the positive class (True Positives), the number of incorrect predictions for the negative class (False Positives), and the number of misclassified positive samples (False Negatives), respectively. In addition to these metrics, the F-score (F$_1$) is commonly reported. It represents the harmonic mean of precision and recall and can be formulated as follows:

$$\text{F}_1=2 \cdot \frac{\text{P} \cdot \text{R}}{\text{P} + \text{R}}.$$

These calculations focus on a single class (the positive class), allowing metrics to be computed for multiple classes in a multi-class scenario. In the task at hand, the primary objective is to classify and detect people, although many studies also report results for other relevant classes. These results may be presented as macro-F$_1$, the average F$_1$ score across all classes, or micro-F$_1$, which aggregates all TP, FP, and FN before calculating the F$_1$ score. While these metrics are typically associated with classification problems, they can also be used to assess object detection performance. In this context, a TP is defined as a correct prediction of an object that closely matches the expected one. This matching is often evaluated using the Jaccard Index---described below---and a threshold to ensure that only predictions of sufficient detection quality are considered.

Other metrics commonly used in classification tasks include specificity and accuracy. Specificity measures a model's ability to correctly identify negative instances (TN), while accuracy indicates the proportion of correctly classified instances across both positive and negative classes. However, accuracy is not suitable for class-imbalanced scenarios, as it tends to favor the majority class. These metrics can be calculated as follows:

$$\text{Specificity}=\frac{\text{TN}}{\text{TN}+\text{FP}}, 
\quad 
\text{Accuracy}=\frac{\text{TP}+\text{TN}}{\text{TP}+\text{TN}+\text{FP}+\text{FN}}.$$ 

Another general-purpose metric is the Mean Absolute Error (MAE), which represents the average magnitude of the errors between predicted and actual values in absolute terms. Lower MAE figures indicate better model performance. It can be computed as: 

$$\text{MAE}=\frac{1}{N}\sum_{i=1}^{N}|y_i - \hat{y}_i|,$$

\noindent where $N$ is the number of instances, $i$ is the evaluated instance, $y$ is the expected output, and $\hat{y}$ is the prediction. Unlike other error metrics, it treats all errors equally, which makes it more robust to outliers.

In some works, the Miss Rate (MR) is used as a metric to measure the proportion of FN, assessing the frequency with which the model fails to detect a relevant object or class. MR is calculated as the ratio of missed detections (FN) to the total number of actual objects (N). A lower MR indicates better model performance in terms of minimizing undetected objects. It can be formulated as:

$$\text{MR} = \frac{\text{FN}}{N}.$$

\subsubsection{Object detection metrics}
\label{sec:metrics:efficacy:object}

One of the most common metrics in object detection is the Intersection over Union (IoU), also known as the Jaccard index. It measures the overlap between the predicted bounding box and the ground truth by dividing the area of their intersection by the area of their union. It can be formulated as: 

$$\text{IoU} = \frac{{\text{area}(\text{BBox}_{\text{pred}}} \cap \text{BBox}_{\text{gt}})}{\text{area}(\text{BBox}_{\text{pred}} \cup \text{BBox}_{\text{gt}})},$$

\noindent where $\text{BBox}_{\text{gt}}$ and $\text{BBox}_{\text{pred}}$ represent the ground-truth and predicted bounding boxes, respectively. The IoU score ranges from 0 to 1, where 1 indicates a perfect match between the predicted and actual bounding boxes.

Another commonly used metric is Average Precision (AP), calculated as the area under the precision-recall curve at a specified IoU threshold. This metric summarizes the trade-off between precision and recall across different confidence thresholds. A higher AP score indicates better model performance, reflecting a more favorable balance between precision and recall at various IoU values. Additionally, Mean Average Precision (mAP) aggregates the results across multiple object classes by averaging the AP for each class.

$$\text{AP} = \int_{0}^{1} p(r) \cdot dr, 
\quad \quad  
\text{mAP} = \frac{1}{n}\sum_{i=1}^{n}\text{AP}_{i},$$

\noindent where $\text{AP}_i$ is the AP of the $i$-th class and \textit{n} the number of classes. 

Standard AP evaluation metrics such as PASCAL-VOC consider the area under the precision-recall curve with an IoU threshold of 0.5, while MS-COCO evaluates AP across a range of IoU thresholds, typically from 0.5 to 0.95 in intervals of 0.05. In some cases, the IoU threshold is reduced to 0.2 to address the challenge of detecting small objects, such as humans, in large-scale images. 

As a standard, mAP is usually denoted with the IoU range considered, such as mAP@0.5-0.95, which represents thresholds from 0.5 to 0.95. For a single threshold, it is denoted as mAP@0.5, indicating an IoU of 0.5.

Average Recall (AR) and Mean Average Recall (mAR) are also commonly used metrics to evaluate recall at different IoU thresholds. These metrics are defined as the area under the recall-IoU curve, with values closer to 1 indicating better performance.

\subsubsection{Tracking metrics}
\label{sec:metrics:efficacy:tracking}

To assess the performance of methods designed for tracking tasks, several specific metrics are employed. One such metric is False Positive Per Image (FPPI), which quantifies the average number of FP per image within a sequence of frames. This measure is crucial for assessing the performance of object tracking algorithms, particularly in applications where minimizing false alarms is essential, such as in security or safety-related systems. It can be calculated as follows, where $m$ is the number of false positives and $F$ is the total number of frames:

$$\text{FPPI}=\frac{m}{F}.$$

Another relevant metric is Multiple Object Tracking Accuracy (MOTA), which combines various types of errors into a single score, reflecting the overall accuracy of both the tracker and the detector. It is computed using the following equation:

$$\text{MOTA}= 1 - \frac{\sum_t (\text{FN}_t + \text{FP}_t + \text{IDSW}_t)}{\sum_{t} \text{GT}_t},$$

\noindent where $\text{FN}_t$, $\text{FP}_t$, $\text{IDSW}_t$, and $\text{GT}_t$ represent the number of FN, FP, identification switches (when a tracked object incorrectly changes its ID), and ground truth objects, respectively, for each frame $t$. This metric provides a value between $-\infty$ and 1, where 1 indicates perfect tracking.

Multiple Object Tracking Precision (MOTP) evaluates the precision of a tracking system by measuring the alignment between predicted and ground-truth bounding boxes over time, independent of object configuration recognition and trajectory consistency. Higher values indicate better performance, i.e., less positional error in object predictions. It is formulated as follows:

$$\text{MOTP}=\frac{\sum_{i,t} d_{t}^{i}}{\sum_{t} \text{TP}_t},$$

\noindent where $d_{t}^{i}$ is the distance between the ground-truth and predicted bounding boxes for the $i$-th object in frame $t$, and $\text{TP}_t$ is the number of matches (correctly identified objects) in frame $t$.

The Higher Order Tracking Accuracy (HOTA)~\cite{luiten2021hota} metric evaluates the alignment of detected object trajectories with the expected ground truth while also penalizing unmatched detections. A simplified computation is given by $\sqrt{\text{DetA} \times \text{AssA}}$, where, unlike traditional metrics, it balances Detection Accuracy (DetA), which measures how well detected objects align with their true locations, and Association Accuracy (AssA), which assesses the consistency of object identities across frames by penalizing identity switches. This metric provides a more comprehensive evaluation of performance.

Identification F$_1$ (IDF$_1$) is a widely used metric in multi-object tracking that evaluates how accurately and consistently an algorithm identifies and tracks objects over time. IDF$_1$ calculates the F$_1$ score based on the correct associations between predicted identities (or tracks) and ground-truth identities.

The ID Switch (IDSW/IDs) metric quantifies the number of times a target's identity changes over a sequence of frames. This metric assesses the stability and reliability of the tracking algorithm; lower values indicate better tracking performance.

Another metric employed in tracking is Fragmentations (Frags), which measures how often a ground-truth trajectory is interrupted and subsequently resumed due to factors such as non-linear motion or occlusion. In simple terms, it counts the instances where a tracked object's status changes to untracked while the ground-truth trajectory still exists. A lower fragmentation value indicates better tracking performance.

\subsection{Efficiency metrics}
\label{sec:metrics:efficiency}

In addition to performance, many works also consider efficiency metrics to assess whether one solution is more efficient than another. This section discusses the most commonly used efficiency metrics in the field of maritime SAR. Note that optimizing efficiency can affect efficacy, meaning that a balance between both is often necessary.

Frames per second (FPS) measures the processing speed of a model, indicating how many frames it can process in one second. In maritime SAR and other real-time detection tasks, a higher FPS is essential for timely data processing. While a model may produce accurate predictions, low FPS can render it impractical for applications such as real-time tracking or rescue missions. Thus, FPS is an important factor in evaluating models, particularly for deployment in embedded systems or UAVs with limited computational resources.

The number of Floating Point Operations (FLOPs) measures the computational complexity of a model, indicating the total number of floating-point operations (multiplications and additions) required for a single prediction. Solutions with fewer FLOPs are generally considered more efficient as they require less computational power, which is particularly important for real-time applications. 

Another comparative metric is the number of parameters (Params) required by a model. The higher the number of parameters, the more resources are needed for its execution, making it less efficient. Parameters refer to the weights and biases learned during training, and they directly affect the model's size and memory requirements.

Latency refers to the delay or time taken for data to travel from one point to another in a system. In computing and network systems, this metric indicates how quickly a system can respond to a request or perform a task. Despite being an important factor to consider, there are only a few studies that analyze it for object detection~\cite{qi2024minimizing}.

Energy consumption is another relevant efficiency metric in maritime SAR, yet it has not been studied in sufficient detail. As UAVs play an increasingly important role in emergency situations, their airborne duration is essential to cover wide search areas and provide extended assistance. However, measuring consumption poses challenges due to multiple factors. It is necessary to account for not only the consumption of the drone, but also that of the monitoring algorithm and the communication system. Additionally, uncontrollable factors such as weather conditions---like increased power requirements on windy or rainy days---complicate accurate assessments.

\section{Results} 
\label{sec:results}

This section summarizes the performance of existing methods on common benchmarks. Given the diversity in techniques, datasets, and tasks, an exhaustive comparison is not feasible. Therefore, we focus on the most relevant and representative results to establish a state-of-the-art reference for future comparisons. Note that variations in data partitions, distributions, and other factors may limit strict comparability, so these results should be considered a general reference.

We only consider results obtained on public datasets (cf. Section~\ref{sec:datasets}), as private data prevents general comparison. Consequently, classification and segmentation tasks are excluded, focusing instead on object detection and tracking, analyzed separately in the following sections.


\subsection{Object detection results}
\label{sec:results:object}

This section presents the results of object detection methods without temporal dependency. Table~\ref{tab:results_image} summarizes the methods, grouped by dataset, along with their best configurations and performance on common metrics. It is important to note that the precision and efficiency ratios calculated for the methods (see columns \textit{ratio Acc./Eff.}), where applicable, depend on the difficulty level of each dataset. A detailed analysis of each public benchmark is presented below.

\input{table_results_image}

\subsubsection{\seadronessee{} benchmark}
\label{sec:results:object:seadronessee}

Most methods proposed for the \seadronessee{} dataset are based on the YOLO architecture, which demonstrates both high efficiency and performance. For mAP@0.5, ABT-YOLOv7~\cite{zhang2023enhanced} achieves the best results, closely followed by Maritime-VSA~\cite{kiefer20231st} and YOLOv7-sea~\cite{zhao2023yolov7}. For the stricter mAP@0.5-0.95 metric, Maritime-VSA~\cite{kiefer20231st} leads with 62\%, followed by DetectorRS~\cite{kiefer20231st} (60\%), and YOLOv7-sea~\cite{zhao2023yolov7} (59\%). However, these methods have limited real-time performance (1 FPS), making them unsuitable for deployment on UAVs.

For efficiency, YOLOv7-FSB~\cite{zhang2023lightweight} outperforms in FPS (96.5) and parameter count (5.82M), significantly ahead of the next best method, Kiefer et al.~\cite{kiefer2023memory}, which achieves 25.1 FPS. Although hardware-dependent, these metrics provide a useful reference.

In summary, YOLOv7-FSB~\cite{zhang2023lightweight} offers the best balance of accuracy and efficiency, with high ratios for these indicators, being the most appropriate method. However, ABT-YOLOv7~\cite{zhang2023enhanced} and Maritime-VSA~\cite{kiefer20231st} are also strong choices if real-time performance is less critical.

\subsubsection{\mobdrone{} benchmark}

Only three studies focus on this dataset, all employing the YOLO architecture. Two of these combine images from \seadronessee{} to enhance generalization (marked with \textsuperscript{$\ddag$} in Table~\ref{tab:results_image}). YOLO-BEV~\cite{yang2023high} achieved the highest mAP@0.5 (97.1\%) and is the only method reporting mAP@0.5-0.95 (56.4\%), a competitive score for this task's complexity. For drone deployment, YOLOv7-FSB~\cite{zhang2023lightweight} is optimal with under 6 million parameters and 96.5 FPS, making it lightweight and fast. YOLO-BEV~\cite{yang2023high} also has strong detection performance at 48 FPS with under 7 million parameters. In contrast, ABT-YOLOv7~\cite{zhang2023enhanced} only reached 7.5 FPS with 52.4 million parameters, making it less suitable for the task, as reflected in its accuracy and efficiency ratios.

Therefore, YOLO-BEV~\cite{yang2023high} and YOLOv7-FSB~\cite{zhang2023lightweight} stand out as the best options for real-time scenarios, both with high ratios of accuracy and efficiency, with YOLOv7-FSB showing remarkable performance.

\subsubsection{\afo{} benchmark}

For the \afo{} benchmark, a wide range of solutions have been proposed, all achieving good accuracy with over 82\% mAP@0.5, showing no clear advantage for any specific technique. The top-performing method, by Zhang et al.~\cite{zhang2023semi}, achieved 96.3\% mAP@0.5 using YOLO and mixed images from the TinyPerson dataset. When using only the \afo{} dataset, Lu et al.~\cite{lu2023high} and the SG-DET method~\cite{zhang2023sg} performed best, with mAP@0.5 scores of 89.7\% and 87.5\%, respectively. For mAP@0.5-0.95, Gasienica et al.~\cite{gasienica2021ensemble} achieved 37.7\%, while Lu et al.~\cite{lu2023high} reached 57.6\%.

In terms of efficiency, only SG-Det~\cite{zhang2023sg} reported a frame rate, achieving 31.9 FPS. Additionally, YOLOv7-CSAW~\cite{zhu2023yolov7} and Lu et al.~\cite{lu2023high} reported model sizes of 58.7M and 57.7M parameters, potentially requiring significant resources.




\subsubsection{\tinyperson{} benchmark}

As described in Section \ref{sec:datasets}, the \tinyperson{} dataset contains images where people appear very small, making detection challenging and leading to lower performance compared to other benchmarks (see Table~\ref{tab:results_image}).

Overall, YOLO and \ac{FR-CNN} are the dominant architectures for this dataset. Top methods by mAP@0.5 include Zhang et al.~\cite{zhang2023semi} with 67.5\% (using additional datasets), DCLANet~\cite{zhang2022finding} at 66.6\%, and Swin-T-EFPNet~\cite{zhang2023efpnet} at 60.7\%. However, none of them report efficiency metrics. In this regard, several methods stand out but sacrificing precision. YOLOv7-UAV~\cite{zeng2023yolov7} reaches 86 FPS with 3.1M parameters, resulting in a high efficiency ratio of 27.7, but lower accuracy (20.6\% mAP@0.5). TO-YOLOX~\cite{chen2023yolox} achieves 103 FPS and 6.7M parameters with 29.6\% mAP@0.5. The fastest, TOD-YOLOv7~\cite{tang2023long}, reaches 208 FPS with 38M parameters, delivering 30\% mAP@0.5 and 9.5\% mAP@0.5-0.95. 

In summary, as with other benchmarks, YOLO-based methods offer the best balance between speed and accuracy, with Zhang et al.~\cite{zhang2023semi} providing the highest precision and TOD-YOLOv7~\cite{tang2023long} delivering the fastest performance.

\subsubsection{\swimmers{} benchmark}

The last dataset analyzed for object detection is \swimmers{}, which, as we can see, has also been tested with different variants of YOLO. While all methods yield similar results, the top performers by mAP@0.5 are Pruned YOLOv4~\cite{rizk2022optimization} at 72.1\%, YOLOv4 Large~\cite{rizk2022towards} at 69.4\%, and Tiny YOLOv3~\cite{lygouras2019unsupervised} at 67\%. For efficiency, Pruned YOLOv4~\cite{rizk2022optimization} stands out as both the fastest and lightest model, achieving 69 FPS with only 1M parameters, closely followed by YOLOv4 Tiny~\cite{rizk2022towards} at 46.6 FPS. The remaining methods appear less suitable for drone deployment due to lower speeds. Overall, Pruned YOLOv4~\cite{rizk2022optimization} appears to be the most suitable choice for this type of images, providing the highest ratios thanks to its reduced number of parameters.

\subsection{Tracking results}

This section analyzes the tracking performance of \ac{SOTA} methods in maritime SAR scenarios. Table~\ref{tab:results_video} compares these approaches on video sequences for detecting and tracking individuals at sea, using the two available public datasets designed for this purpose: \seadronessee and \mobdrone.

\input{table_results_video}

\subsubsection{\seadronessee{} benchmark}

For the \seadronessee{} benchmark, we found five studies with comparable results. Three are based on YOLO, one uses an autoencoder, and another employs Swin-T. Performance varies by metric: with MOTA, used by four out of five, YOLOv7-FSB~\cite{zhang2023lightweight} ranks highest (87.6\%), followed by Memory Map~\cite{kiefer2023memory} (80.8\%). However, according to HOTA---where YOLOv7-FSB lacks data---MG-MOT achieves the best results, while Memory Map performs closely on MOTA. In terms of MOTP, only two studies report similar values, making it hard to identify a superior method in this regard.

In terms of IDs and Fragments, YOLOv7-FSB~\cite{zhang2023lightweight} and MG-MOT~\cite{yang2024sea} show the best tracking consistency across sequences. Recall values are comparable across three methods, with ReIDTracker\_Sea~\cite{huang2023reidtracker} slightly higher. However, this last method is inefficient for real-time use at just 3 FPS, unlike the others.

Overall, MG-MOT~\cite{yang2024sea} and YOLOv7-FSB~\cite{zhang2023lightweight} emerge as the top approaches.

\subsubsection{\mobdrone{} benchmark}

The \mobdrone{} benchmark has only been evaluated with YOLOv7-FSB~\cite{zhang2023lightweight}, one of the top-performing models in the \seadronessee{} benchmark. It reports three metrics: 87.6\% of MOTA, 18 IDs, and 96.5 FPS, indicating that this method is well-suited for real-time tracking of individuals in maritime SAR. However, further studies are needed to compare its performance with other models applied to this benchmark.

\section{Discussion}
\label{sec:discussion}

In this section, we critically analyze the findings of this work, situating them within current trends and research in the field. We discuss the limitations of the proposed approaches, identify areas for improvement, and suggest future directions to broaden the scope of this study.

\subsection{Current trend and limitations}

\begin{figure}[!ht]
    \centering
    \frame{\includegraphics[width=0.9\textwidth]{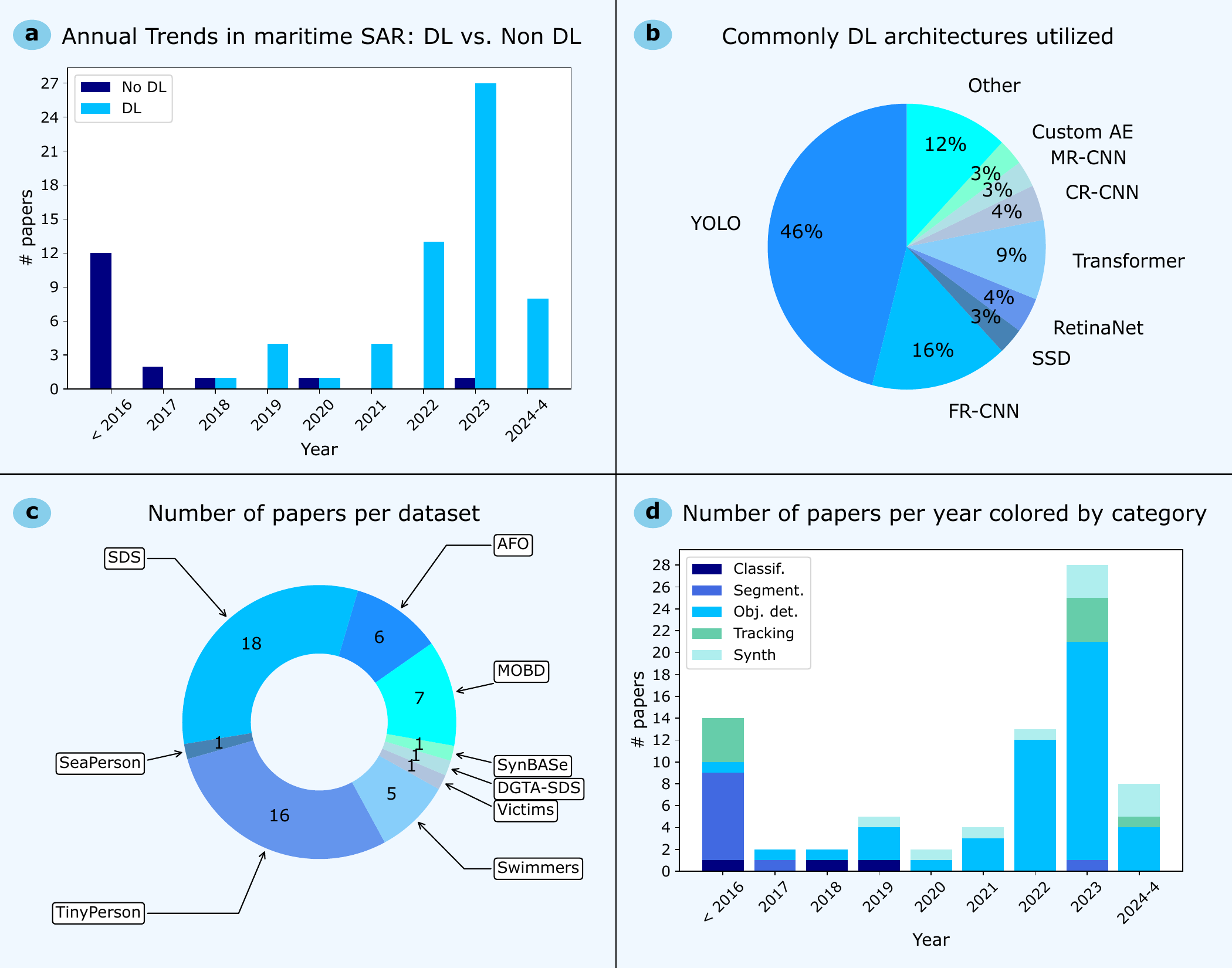}}
    \caption{Statistical analysis of publications on maritime SAR: 
    a) number of approaches per year, categorized by methodology (DL/Non-DL); 
    b) distribution of DL architectures used, with 'other' indicating custom-developed CNNs or less common architectures; 
    c) dataset usage distribution; 
    and d) year-wise distribution of publications according to the categories analyzed in this article.}
    \label{fig:analysis_plots}
\end{figure}

During the preparation of this survey, we observed a growing interest in this field. Figure~\ref{fig:analysis_plots}a illustrates this trend: while the number of papers published until 2021 was relatively low, it increased from 2022 onwards, peaking in 2023, which saw double the number of articles published up to 2016. This growth is largely attributed to the rise of \ac{DL}, which has seen a marked increase in usage since 2022. As shown in Figure~\ref{fig:analysis_plots}b, YOLO is the dominant architecture, used in 46\% of studies, followed by \ac{FR-CNN} at 16\%. Despite their high computational cost, transformers have also made a significant impact. The remaining architectures show marginal use, suggesting they are less suitable for this task or, at least, insufficiently explored.

Regarding the datasets (see Fig.~\ref{fig:analysis_plots}c), SeaDronesSee is the most widely used, followed by TinyPerson. The AFO, MOBDrone, and Swimmers datasets have also been used, but to a lesser extent. The remaining datasets, including the three synthetic ones, have not been utilized beyond their initial publications, probably due to the recent emergence of these techniques or their limited visibility.

Looking at the number of methods published grouped by category (Fig.~\ref{fig:analysis_plots}d), object detection is the most widely covered task. Traditionally, classification approaches were combined with segmentation, which was the most common task. However, with the rise of \ac{DL} and object detection, these tasks have largely been abandoned. Classification was limited in accuracy for locating people, while segmentation became impractical due to the effort and cost of manually labeling images at the pixel level. In contrast, labeling bounding boxes around objects (i.e., people) is much more efficient, which explains the growing number of datasets and studies focused on detection. Tracking was also initially explored, but it saw a decline in interest until 2023, when advancements in \ac{DL} led to its resurgence, yielding promising results. While tracking has many applications, it is more costly to develop, demands more resources and data for training, and requires high processing speeds to track targets effectively. Regarding synthetic data, its use began around 2019 and was limited until 2023, when more sophisticated methods for generating realistic images were developed. These advances have shown synthetic data to be a valuable tool, addressing data scarcity and enabling the training of robust models with large datasets.

Unlike classical \ac{CV} approaches, which relied on hand-crafted solutions that yielded good results but generalized poorly, the emergence of \ac{DL} has demonstrated superior performance in real-world applications. These algorithms self-learn features through machine-learning techniques using labeled training sets. Given the demanding criteria of effectiveness, real-time performance, and robustness required for systems designed to search and rescue people at sea, such models must excel in these areas to be both useful and practical for real-world implementation. The exhaustive literature review reveals that only a small fraction of studies manage to balance precision and speed. Approaches attempting to achieve this balance often adopt the YOLO series architecture. As shown in Tables~\ref{tab:results_image} and \ref{tab:results_video}, these models have been the most widely considered and serve as the foundation for many proposals. In most cases, these approaches achieve good FPS by maintaining a low number of parameters.

Transformer-based models are gaining importance across various contexts, with recent studies investigating their application for detecting people at sea. While research in this area is still limited, there is a clear trend towards further exploration. However, these models require large datasets, which are currently scarce in this field. The literature suggests that synthetic data can help create more robust models with limited real data, but there is still a lack of publicly available databases to support this development. In addition, there is a lack of a common and representative benchmark that may be used by all methods in order to make more effective comparisons with SOTA. The evaluation of the methods on different datasets makes it difficult to compare their results and draw conclusions from a wider perspective.

Another strategy to improve the detection of small objects is the integration of attention modules, which have recently been applied in the maritime context to capture deeper semantic information about people overboard. Modules such as \ac{SimAM}, C2f, \ac{CBAM}, and BiFormer have been used, and approaches incorporating these modules have outperformed previous results in key maritime datasets, positioning them as the current \ac{SOTA} in this domain. Thus, the application of these techniques has significantly advanced the field of SAR at sea.

A recurring concern in the literature is the development of robust models that can perform under adverse conditions. Several factors limit performance in such scenarios, but available benchmarks cover only a small range of situations compared to what models will encounter in real rescue operations. Variables such as waves, glare, reflections, time of day, lighting conditions, shadows, weather, camera perspective, and the presence of other elements, among others, complicate rescue tasks, making them a significant challenge. While some studies have attempted to address this, research remains limited and insufficient.

\subsection{Future directions}

As we have seen, there is a clear trend toward the use of \ac{DL} techniques. However, achieving robust systems with these techniques requires large amounts of training data, which are essential for reliable and effective models. So far, there are not many public datasets, especially for tracking tasks (see Section~\ref{sec:datasets}). Additionally, its variability is limited, covering very few scenarios that could occur in a real environment. These limitations pose a significant obstacle, hindering the development of new solutions. 
To advance the field, more diverse datasets are needed across all maritime SAR tasks, enabling the creation of increasingly accurate models. While additional datasets may emerge over time, data collection remains challenging due to high costs and the need for licenses and specialized equipment. Nonetheless, this is essential for continued progress.

Generating synthetic data offers a viable short-term solution to the current lack of data variability and volume. Modern tools can produce highly realistic, automatically labeled images, enabling the creation of extensive datasets without deploying teams to simulate accidents. Various strategies have been explored to integrate these data into the training process, such as the use of pre-trained models or the combination of both distributions---synthetic and real---yielding promising results that enhance the effectiveness of the models in real-world scenarios. In cases where data collection is impractical---such as in fog, storms, or other uncontrollable environments that could endanger drones and incur high costs---synthetic data can effectively complement real data, providing valuable training material with minimal risk and effort. These challenging conditions are critical for rescue operations, making synthetic data essential for training models that can generalize and perform well under such circumstances. Therefore, based on all the above, these needs may represent a future line of research to pursue in the coming years: exploring new techniques to effectively integrate synthetic and real data, as well as developing more powerful synthetic generators capable of producing data dynamically tailored to any type of required scenario.

On the other hand, domain shifts and adverse weather conditions such as night-time, rain, fog, or the state of the sea, present significant challenges for the detection of individuals at sea, as they introduce variability that may greatly affect the performance of detection systems. It has been observed that most of the reviewed papers propose supervised strategies that perform well within the domain they were trained on; however, when tested in different domains, they tend to fail. The systems developed in this field must become increasingly capable of overcoming these barriers and ensuring reliable performance under all conditions. To get close to this ideal situation, the development of robust techniques, such as ensemble learning or domain adaptation, is essential. Ensemble methods, for instance, could combine the strengths of multiple models trained on diverse scenarios, increasing system resilience across unseen conditions. Similarly, domain adaptation techniques can be employed to bridge the gap between different domains, leveraging unlabeled data to enhance model generalization. Unsupervised approaches, in particular, hold great promise by allowing models to adapt to new environments without requiring extensive annotations, which is especially beneficial in maritime SAR where data collecting and labeling are costly and sometimes impractical. 

Advanced techniques, often first applied in other fields, are increasingly adapted for maritime rescue. Trends like transformers and attention mechanisms show potential due to their effectiveness with large datasets. While labeled data for maritime SAR remains limited, synthetic data generation offers a promising solution, allowing transformers to be trained on extensive, diverse datasets and enabling better generalization to all types of real-world scenarios. 

Another point to consider is that most proposed methods rely on YOLO or \ac{FR-CNN}, leaving many other architectures with potential---such as SSD, RetinaNet, DINO, \ac{CR-CNN}, and CenterNet---insufficiently explored. Some of these models have been evaluated on limited datasets, but their ability to generalize or perform robustly in realistic scenarios remains unexplored. Expanding experiments to include a broader range of architectures and even exploring ensemble methods could provide valuable insights and improve performance.

As discussed throughout this review, improving efficiency is crucial, as time is a limiting factor that directly impacts survival chances. Efficiency not only refers to detection speed but also to reducing computational resource requirements without sacrificing accuracy. Many of the studies discussed in this review demonstrate high accuracy but low efficiency, while others are computationally intensive, making them impractical for real-time deployment on drones. This is particularly critical as drones often have limited hardware capabilities and must cover large areas with limited battery capacity, rendering such models unsuitable for effective real-time maritime SAR operations. Therefore, developing lightweight models that provide high accuracy and are capable of achieving reliable real-time performance is a key area of interest for future research. Additionally, task offloading is gaining importance, where computationally demanding tasks are offloaded to external systems like cloud servers or edge devices. This approach enables more complex operations without overloading onboard systems. Future research will likely focus on methods such as model optimization, edge computing, pruning, and quantization, along with leveraging advanced hardware to enhance detection performance. These advancements will help UAVs operate for longer periods, cover larger areas, and improve real-time responsiveness.

\section{Conclusions}
\label{sec:conclusions}

This review has provided a comprehensive analysis of the current state of methodologies for detecting people at sea using aerial images, with a particular focus on SAR operations. Several key conclusions can be highlighted.

The integration of UAVs with DL techniques has notably enhanced the efficiency of SAR missions. However, detecting individuals in maritime environments remains challenging due to factors like the small size of the bodies in aerial images, sea variability, and changing weather conditions. These issues often lead to false detections, highlighting the need for more robust models.

A major obstacle is the scarcity of annotated real-world data. Synthetic data generation has emerged as a practical solution to diversify training datasets, but there is still a performance gap when transitioning from synthetic to real-world scenarios. Bridging this domain gap requires advanced techniques to better simulate realistic conditions.

The field also lacks standardized evaluation metrics and comprehensive publicly available datasets. Although some datasets have been introduced, larger and more varied collections are needed to better reflect real-world conditions and to facilitate consistent benchmarking of different methods.

Future research should focus on developing more sophisticated models capable of handling the inherent variability of maritime environments. This includes exploring advanced architectures like attention mechanisms and transformer-based models, as well as leveraging metadata such as GPS and environmental conditions to improve robustness. Additionally, real-time processing and lightweight models suitable for UAV deployment will be crucial for practical application.

\section*{Acknowledgment}

This work was supported by the Generalitat Valenciana (GV) through the TADMar project, INVEST/2022/450.



\bibliographystyle{model5-names} 


\bibliography{paper}

\end{document}

%% file: acronims.tex
\usepackage{acronym}
\acrodef{AI}{Artificial Intelligence}
\acrodef{CNN}{Convolutional Neural Network}
\acrodef{RED-Net}{Very deep Residual Encoder-Decoder Network}
\acrodef{SAR}{Search and Rescue}
\acrodef{UAV}{Unmanned Aerial Vehicle}
\acrodef{USV}{Unmanned Surface Vessel}
\acrodef{USV}{Unmanned Surface Vehicle}
\acrodef{RS}{Remote Sensing}
\acrodef{CV}{Computer Vision}
\acrodef{ML}{Matching Learning}
\acrodef{DL}{Deep Learning}
\acrodef{IoU}{Intersection over Union}
\acrodef{DA}{Data Augmentation}
\acrodef{BB}{Bounding Boxes}

\acrodef{FCN}{Fully-Convolutional Network}
\acrodef{DNN}{Deep Neural Network}

\acrodef{GMM}{Gaussian Mixture Model}
\acrodef{IR}{infrared image}
\acrodef{SVD}{Singular Value Decomposition}
\acrodef{FFT}{Fast Fourier Transforms}
\acrodef{SOTA}{state-of-the-art}
\acrodef{YOLO}{You-Only-Look-Once}
\acrodef{SSD}{Single-Shot Detector}
\acrodef{GNSS}{Global Navigation Satelite System}
\acrodef{MMD}{Maximum Mean Discrepancy distance}
\acrodef{VAM}{Visual Attention Mechanism}
\acrodef{SimAM}{Simple Parameter-Free Attention Module}
\acrodef{ASFF}{Adaptive Feature Fusion Network}
\acrodef{MAEM}{Multiscale Attention Enhancement Module}
\acrodef{MFFM}{Multiscale Feature Fusion Module}
\acrodef{CBAM}{Convolutional Block Attention Module}
\acrodef{MCE-CBAM}{Multi-level Cascaded Enhanced Convolutional Block Attention Module}
\acrodef{C2fELAN}{Two Convolutions Efficient Layer Aggregation Networks}
\acrodef{ASPP}{Atrous Spatial Pyramid Pooling}
\acrodef{AFPN}{Asymptotic Feature Pyramid Network}
\acrodef{TSCODE}{Task-Specific COntext DEcoupling}
\acrodef{ReID}{Re-IDentification}
\acrodef{DETR}{DEtection TRansformer}
\acrodef{DCNv2}{Deformable Convolutional Networks v2}
\acrodef{SIMMIM}{SIMple framework for Masked Image Modeling}
\acrodef{ViT}{Vision Transformer}
\acrodef{UE}{Unreal Engine}
\acrodef{ROS}{Robot Operating System}
\acrodef{BLD}{Blended Latent Diffusion}
\acrodef{PSO}{Particle Swarm Optimization}
\acrodef{VSA}{Varied-Size window Attention}
\acrodef{HMM}{Hidden Markov Model}
\acrodef{FFTT}{First Frame of True Target Detection}
\acrodef{FA}{False Alarm}
\acrodef{CCR}{Correct Classification Rate}
\acrodef{TEFF}{Texture-Enhanced Feature Fusion}
\acrodef{CAM}{Channel Attention Mechanism}
\acrodef{BiFPN}{Bidirectional Feature Pyramid Network}

\acrodef{FR-CNN}{Faster R-CNN}
\acrodef{MR-CNN}{Mask R-CNN}
\acrodef{CR-CNN}{Cascade R-CNN}
\acrodef{Swin-T}{Swin-Transformer}

\acrodef{SDS}{SeaDronesSee}
\acrodef{MOBD}{MOBDrone}

%% file: table_methods.tex
\begin{table}[!ht]
    \centering
    \caption{Summary of methods for SAR operations at sea, organized by task. Each method is described in terms of the following columns: reference (\textbf{Ref.}), \textbf{year}, methodology \textbf{type}, primary \textbf{methods} used, system \textbf{input} and \textbf{output}, \textbf{code} availability, \textbf{corpora} used for experiments, and evaluation \textbf{metrics}. Additionally, two final columns summarize the accuracy (\textbf{Acc.}) and efficiency (\textbf{Eff.}) of each method, represented with up to three $\bullet$ to provide a visual and comparative estimation of their performance. Some papers are listed in multiple categories as they address more than one task.}
    \label{tab:methodology}
    
    \renewcommand{\arraystretch}{0.55}
    \setlength{\tabcolsep}{1pt}
    \resizebox{1\textwidth}{!}{
    \begin{tabular}{l cccccc l lclcl}
    
    \toprule[2pt]
       
    & \multicolumn{6}{c}{\textit{\textbf{Methodology}}} 
        & 
        & \multicolumn{5}{c}{\textit{\textbf{Experiments}}} \\
    
    \cmidrule{2-7} \cmidrule{9-13} 
    
    \textit{\textbf{Ref.}} & \textit{\textbf{Year}} & \textit{\textbf{Type}} & \textit{\textbf{Methods}} & \textit{\textbf{Input}} & \textit{\textbf{Output}} & \textit{\textbf{Code}} & & \textit{\textbf{Corpora}} & \textit{\textbf{Metrics}} &
    \textit{\textbf{Acc.}} && \textit{\textbf{Eff.}} \\
    
    \midrule
    \midrule
    \multicolumn{13}{l}{\textbf{\textit{Classification}}} \\ 
    \midrule

    Leira et al.\cite{leira2015automatic}                  
    & 2015
    & CV
    & NNC, Hu moments
    & Thermal
    & Hum./No hum.
    & \xmark
    &
    & Private
    & CCR
    & $\bullet$$\bullet$
    & \phantom{0}
    & $\bullet$ \\

    Rodin et al.~\cite{rodin2018object} 
    & 2018 
    & DL
    & CNN, GMM
    & Thermal
    & Hum./No hum.
    & \xmark
    &
    & Private
    & Prob, Acc
    & $\bullet$
    & \phantom{0}
    & $\bullet$$\bullet$ \\   

    Gallego et al.~\cite{gallego2019detection} 
    & 2019 
    & DL
    & MobileNet
    & Multisp.
    & Hum./No hum.
    & \xmark
    &
    & Private
    & P, R, F1, F2
    & $\bullet$$\bullet$
    & \phantom{0}
    & $\bullet$$\bullet$ \\

    \toprule[2pt] 
    \multicolumn{13}{l}{\textbf{\textit{Object segmentation}}} \\ 
    \midrule

    Sumimoto et al.~\cite{sumimoto1994machine}                
    & 1994 
    & CV
    & CV techniques
    & RGB
    & Bin. map
    & \xmark
    &
    & Private
    & Qualitative examples
    & $\bullet$
    & \phantom{0}
    & $\bullet$ \\    

    Yamamoto et al.~\cite{yamamoto1999optical}                
    & 1999
    & CV
    & CV techniques
    & IR
    & Bin. map
    & \xmark
    &
    & Private
    & Qualitative examples
    & $\bullet$
    & \phantom{0}
    & $\bullet$ \\

    Shi et al.~\cite{shi2008architecture}                
    & 2008
    & CV
    & CV techniques
    & RGB, IR
    & Bin. map
    & \xmark
    &
    & Private
    & Qualitative examples
    & $\bullet$
    & \phantom{0}
    & $\bullet$ \\

    Ran et al.~\cite{ran2010search}                
    & 2010
    & CV
    & CV techniques, saliency
    & RGB, IR
    & Bin. map
    & \xmark
    &
    & Private
    & Qualitative examples
    & $\bullet$
    & \phantom{0}
    & $\bullet$ \\

    Ren et al.~\cite{ren2011target}
    & 2011 
    & ML
    & SVD, saliency
    & RGB
    & Bin. map
    & \xmark
    &
    & Private
    & Qualitative examples
    & $\bullet$
    & \phantom{0}
    & $\bullet$ \\

    Ren et al.~\cite{ren2012target}
    & 2012
    & CV
    & CV techniques, saliency
    & RGB
    & Bin. map
    & \xmark
    &
    & Private
    & Qualitative examples
    & $\bullet$
    & \phantom{0}
    & $\bullet$ \\

    Kim et al.~\cite{kim2014small}                
    & 2014 
    & CV
    & CV techniques
    & IR
    & Bin. map
    & \xmark
    &
    & Private
    & DR, FAR 
    & $\bullet$$\bullet$
    & \phantom{0}
    & $\bullet$ \\

    Mendonça et al.~\cite{mendoncca2016cooperative}               
    & 2016
    & CV
    & CV techniques
    & Thermal
    & Bin. map
    &\xmark
    &
    & Private
    & Qualitative examples
    & $\bullet$$\bullet$
    & \phantom{0}
    & $\bullet$$\bullet$ \\

    Dinnbier et al.~\cite{dinnbier2017target}                
    & 2017 
    & ML
    & GMM, FFT
    & RGB
    & Bin. map
    & \xmark
    &
    & Private
    & TP, FP 
    & $\bullet$$\bullet$$\bullet$
    & \phantom{0}
    & $\bullet$$\bullet$ \\

    N-FINDR~\cite{park2023aerial} 
    & 2023
    & CV
    & N-FINDR
    & RGB, Hypersp.
    & Frac.map, Ellip.
    & \xmark
    &
    & Private
    & Localization error, size
    & $\bullet$$\bullet$$\bullet$
    & \phantom{0}
    & $\bullet$$\bullet$ \\

    \toprule[2pt] 
    \multicolumn{13}{l}{\textbf{\textit{Object detection}}} \\ 
    \midrule

    Leira et al.~\cite{leira2015automatic}  
    & 2015
    & CV
    & CV techniques
    & Thermal
    & BB
    & \xmark
    &
    & Private
    & \# det., \% det. 
    & $\bullet$$\bullet$
    & \phantom{0}
    & $\bullet$ \\

    Hoai et al.~\cite{hoai2017anomaly}        
    & 2017
    & CV
    & RX detector
    & Multiple CS 
    & BB
    & \xmark
    &
    & Private
    & ROC, AUC
    & $\bullet$
    & \phantom{0}
    & $\bullet$ \\

    Lygouras et al.~\cite{lygouras2018rolfer, lygouras2019unsupervised} 
    & 2018/19 
    & DL
    & YOLOv3, GNSS
    & RGB
    & BB
    & \xmark
    &
    & Swimmers
    & Acc, R, F1, mAP, FPS
    & $\bullet$
    & \phantom{0}
    & $\bullet$ \\

    Gallego et al.~\cite{gallego2019detection} 
    & 2019
    & DL
    & MobileNet
    & Multisp.
    & BB
    & \xmark
    &
    & Private
    & P, R, F1, F2, MAE
    & $\bullet$$\bullet$
    & \phantom{0}
    & $\bullet$ \\

    Moosbauer et al.~\cite{moosbauer2019benchmark}      
    & 2019 
    & DL
    & FR-CNN, MR-CNN
    & RGB
    & BB
    & \xmark
    &
    & SMD
    & F1
    & $\bullet$$\bullet$
    & \phantom{0}
    & $\bullet$ \\

    Feraru et al.~\cite{feraru2020towards} 
    & 2020
    & DL
    & FR-CNN, GNSS
    & Thermal
    & BB
    & \xmark
    &
    & Private
    & TP, FP, FN, P, R, AP, speed 
    & $\bullet$$\bullet$
    & \phantom{0}
    & $\bullet$ \\

    Gasienica et al.~\cite{gasienica2021ensemble} 
    & 2021
    & DL
    & FR-CNN,RetinaNet,SSD,YOLOv3/4
    & RGB
    & BB
    & \xmark
    &
    & AFO
    & mAP
    & $\bullet$
    & \phantom{0}
    & $\bullet$$\bullet$ \\

    TEFF~\cite{zhou2021texture} 
    & 2021
    & DL
    & FR-CNN, RetinaNet
    & RGB
    & BB
    & \xmark
    &
    & TinyPerson
    & mAP, MR 
    & $\bullet$$\bullet$
    & \phantom{0}
    & $\bullet$ \\

    DyHead~\cite{dai2021dynamic} 
    & 2021
    & DL
    & Swin-T
    & RGB
    & BB
    & \xmark
    &
    & SDS
    & mAP, FPS
    & $\bullet$$\bullet$$\bullet$
    & \phantom{0}
    & $\bullet$ \\

    Sharafaldeen et al.~\cite{sharafaldeen2022marine} 
    & 2022
    & DL
    & YOLOv4
    & RGB, IR
    & BB
    & \xmark
    &
    & \begin{tabular}[c]{@{}l@{}}SDS,MOBD,\\Swim.,AFO\end{tabular}
    & P, R, F1, IoU, mAP, FPS
    & $\bullet$$\bullet$
    & \phantom{0}
    & $\bullet$$\bullet$ \\

    Gonçalves et al.~\cite{gonccalves2022automatic} 
    & 2022
    & DL
    & YOLOv4-tiny
    & RGB
    & BB
    & \xmark
    &
    & Private
    & P, R
    & $\bullet$
    & \phantom{0}
    & $\bullet$$\bullet$ \\


    Rizk et al.~\cite{rizk2022optimization,rizk2022towards} 
    & 2022
    & DL
    & YOLOv4, YOLOv4-Tiny
    & RGB
    & BB
    & \xmark
    &
    & Swim.,Custom
    & P,R,F1,mAP,IoU,Params,FPS,FLOPS
    & $\bullet$$\bullet$
    & \phantom{0}
    & $\bullet$$\bullet$ \\

    Bhuiya et al.~\cite{bhuiya2022surveillance} 
    & 2022
    & DL
    & VGG16, PSO
    & RGB
    & BB
    & \xmark
    &
    & Swimmers
    & Acc, IoU, time
    & $\bullet$$\bullet$
    & \phantom{0}
    & $\bullet$ \\

    Bai et al.~\cite{bai2022detection} 
    & 2022 
    & DL
    & YOLOv5, C3Ghost
    & RGB
    & BB
    & \xmark
    &
    & Private
    & mAP, speed, FLOPs, Params, weight  
    & $\bullet$$\bullet$
    & \phantom{0}
    & $\bullet$$\bullet$$\bullet$\\

    Cafarelli et al.~\cite{cafarelli2022mobdrone} 
    & 2022 
    & DL
    & \begin{tabular}[c]{@{}l@{}}Varifocal.,TOOD,YOLOX,FR-CNN,\\CenterNet,MR-CNN,YOLOv3,DETR\end{tabular}
    & RGB
    & BB
    & \xmark
    &
    & MOBD
    & TPR, R, F1, mAP  
    & $\bullet$
    & \phantom{0}
    & $\bullet$ \\

    Gao et al.~\cite{gao2022multiscale} 
    & 2022
    & DL
    & Swin-T, MAEM, MFFM
    & RGB
    & BB
    & \checkmark
    &
    & TinyPerson
    & mAP, FLOPs, Params 
    & $\bullet$$\bullet$
    & \phantom{0}
    & $\bullet$\\

    Zhang et al.~\cite{zhang2023semi} 
    & 2023
    & DL
    & YOLOv5, MMD
    & RGB
    & BB
    & \xmark
    &
    & AFO,TinyPerson
    & P, R, mAP
    & $\bullet$$\bullet$$\bullet$
    & \phantom{0}
    & $\bullet$$\bullet$\\
    
    Falcon~\cite{oda2023falcon} 
    & 2023 
    & DL
    & YOLOv5
    & RGB
    & BB
    & \xmark
    &
    & MOBD, private 
    & P, R, AP, Compression Rate 
    & $\bullet$
    & \phantom{0}
    & $\bullet$$\bullet$\\

    Fernandes et al.~\cite{fernandes2023enhancing} 
    & 2023
    & DL
    & YOLOv7
    & RGB
    & BB
    & \checkmark
    &
    & SDS
    & AR, mAP 
    & $\bullet$$\bullet$
    & \phantom{0}
    & $\bullet$$\bullet$ \\

    Lu et al.~\cite{lu2023high} 
    & 2023 
    & DL
    & SIMMIM, ViT, DINO
    & RGB
    & BB
    & \xmark
    &
    & AFO
    & P, R, mAP, FLOPs, Params
    & $\bullet$$\bullet$$\bullet$
    & \phantom{0}
    & $\bullet$ \\
    
    YOLOv7-sea~\cite{zhao2023yolov7}                  
    & 2023
    & DL
    & YOLOv7, SimAM
    & RGB
    & BB
    & \xmark
    &
    & SDS
    & AR, mAP
    & $\bullet$$\bullet$$\bullet$
    & \phantom{0}
    & $\bullet$ \\

    Maritime-VSA~\cite{kiefer20231st}                  
    & 2023 
    & DL
    & Swin-T, CBNetV2, CR-CNN, VSA
    & RGB
    & BB
    & \xmark
    &
    & SDS
    & mAP, FPS
    & $\bullet$$\bullet$$\bullet$
    & \phantom{0}
    & $\bullet$ \\

    DetectorRS~\cite{kiefer20231st}                  
    & 2023
    & DL
    & DetectoRS, CR-CNN
    & RGB
    & BB
    & \checkmark
    &
    & SDS
    & mAP, FPS
    & $\bullet$$\bullet$$\bullet$
    & \phantom{0}
    & $\bullet$ \\

    YOLOv7-CSAW~\cite{zhu2023yolov7}                  
    & 2023 
    & DL
    & YOLOv7,K-means++,C2f mod.,SimAM
    & RGB
    & BB
    & \xmark
    &
    & AFO
    & mAP, FLOPs, Params
    & $\bullet$$\bullet$$\bullet$
    & \phantom{0}
    & $\bullet$ \\

    YOLO-BEV~\cite{yang2023high} 
    & 2023
    & DL
    & YOLOv5, CSPDarknet53, PAN+
    & RGB
    & BB
    & \xmark
    &
    & MOBD
    & mAP, mAR, FLOPs, Params
    & $\bullet$$\bullet$$\bullet$
    & \phantom{0}
    & $\bullet$$\bullet$$\bullet$ \\

    Sea-YOLOv5s~\cite{wang2023sea} 
    & 2023 
    & DL
    & YOLOv5, Swin-T, CBAM
    & RGB
    & BB
    & \xmark
    &
    & SDS
    & mAP, mAR
    & $\bullet$
    & \phantom{0}
    & $\bullet$$\bullet$ \\

    SG-Det~\cite{zhang2023sg} 
    & 2023
    & DL
    & BiFPN-tiny, ASPP module
    & RGB
    & BB
    & \xmark
    &
    & AFO
    & mAP, FPS, FLOPs, Params
    & $\bullet$$\bullet$
    & \phantom{0}
    & $\bullet$$\bullet$$\bullet$ \\

    ABT-YOLOv7~\cite{zhang2023enhanced} 
    & 2023
    & DL
    & AFPN, BiFormer
    & RGB
    & BB
    & \xmark
    &
    & SDS, MOBD
    & P, R, mAP, FPS, Params
    & $\bullet$$\bullet$$\bullet$
    & \phantom{0}
    & $\bullet$ \\

    YOLOv5s-SwinDS~\cite{liu2024yolov5s} 
    & 2024 
    & DL
    & YOLOv5, Swin-T
    & RGB
    & BB
    & \xmark
    &
    & SDS
    & P, R, mAP, FLOPs
    & $\bullet$$\bullet$
    & \phantom{0}
    & $\bullet$$\bullet$ \\

    Zhao et al.~\cite{zhao2024heuristic} 
    & 2024
    & DL
    & FR-CNN, CR-CNN, RetinaNet
    & RGB
    & BB
    & \checkmark
    &
    & SDS
    & mAP
    & $\bullet$$\bullet$
    & \phantom{0}
    & $\bullet$ \\

    MTP-YOLO~\cite{shi2024mtp} 
    & 2024
    & DL
    & YOLOv9, MCE-CBAM, C2fELAN
    & RGB
    & BB
    & \xmark
    &
    & SeaPerson
    & P, R, mAP, FLOPS, Params
    & $\bullet$$\bullet$$\bullet$
    & \phantom{0}
    & $\bullet$ \\

    \toprule[2pt] 
    \multicolumn{13}{l}{\textbf{\textit{Tracking}}} \\ 
    \midrule

    Westall et al.~\cite{Westall2008,Westall2009}
    & 2008/09 
    & CV
    & DP, HMM
    & Multiple CS 
    & \# det.
    & \xmark
    &
    & Private
    & FPPI, MR, FFTT, False Alarms 
    & $\bullet$
    & \phantom{0}
    & $\bullet$ \\


    Kaiyu \& Chaojian\cite{kaiyu2009vision}
    & 2009 
    & CV
    & CV techniques
    & IR
    & Position
    & \xmark
    &
    & Private
    & \# det. 
    & $\bullet$
    & \phantom{0}
    & $\bullet$ \\

    Leira et al.~\cite{leira2015automatic}
    & 2015
    & ML
    & Kalman Filter
    & Thermal
    & Path,Correct/Not
    & \xmark
    &
    & Private
    & \# det. 
    & $\bullet$$\bullet$
    & \phantom{0}
    & $\bullet$ \\

    YOLOv7-FSB~\cite{zhang2023lightweight} 
    & 2023 
    & DL
    & YOLOv7, ByteTrack, SimAM, BiFPN
    & RGB
    & BB
    & \xmark
    &
    & SDS, MOBD 
    & MOTA, P, R, mAP, FPS, Params
    & $\bullet$$\bullet$$\bullet$
    & \phantom{0}
    & $\bullet$$\bullet$$\bullet$ \\

    ReIDTracker\_Sea~\cite{huang2023reidtracker} 
    & 2023
    & DL
    & Swin-T,  MoCo-v2, ReID
    & RGB
    & Paths, BB
    & \xmark
    &
    & SDS
    & \begin{tabular}[c]{@{}l@{}}MOTA,MOTP,HOTA,IDF1,\\FP,FN,FPS,FLOPs\end{tabular}
    & $\bullet$$\bullet$
    & \phantom{0}
    & $\bullet$ \\

    Kiefer et al.~\cite{kiefer2023fast} 
    & 2023
    & DL
    & Autoencoder
    & RGB
    & Regions, BB
    & \checkmark
    &
    & SDS
    & R, AR, Reconstruction error, FPS
    & $\bullet$$\bullet$
    & \phantom{0}
    & $\bullet$$\bullet$ \\

    Kiefer et al.~\cite{kiefer2023memory} 
    & 2023
    & DL
    & \begin{tabular}[c]{@{}l@{}}Autoencoder, YOLOv7-Tiny,\\DeepSORT, memory maps\end{tabular}
    & RGB
    & Heatmaps, BB
    & \xmark
    &
    & SDS
    & MOTA, HOTA, IDs, Frag, AP, AR, FPS
    & $\bullet$$\bullet$
    & \phantom{0}
    & $\bullet$$\bullet$ \\

    MG-MOT~\cite{yang2024sea} 
    & 2024 
    & DL
    & YOLOv8-x, ReID
    & RGB
    & BB
    & \xmark
    &
    & SDS
    & \begin{tabular}[c]{@{}l@{}}MOTA, MOTP, HOTA, IDF1, IDs,\\ Frag, MT, ML, FP, FN, P, R\end{tabular} 
    & $\bullet$$\bullet$$\bullet$
    & \phantom{0}
    & $\bullet$$\bullet$ \\

    \toprule[2pt] 
    \multicolumn{13}{l}{\textbf{\textit{Synthetic Data Generation}}} \\ 
    \midrule

    Yun et al.~\cite{yun2019small} 
    & 2019
    & DL
    & ARMA3, SSD
    & RGB
    & BB
    & \xmark
    &
    & Private
    & AP, AR, mAP 
    & $\bullet$$\bullet$
    & \phantom{0}
    & $\bullet$ \\
    
    Usilin et al.~\cite{usilin2020training} 
    & 2020
    & ML
    & Blender, Viola-Jones
    & Grayscale
    & BB
    & \xmark
    &
    & Private
    & P, R, F1 
    & $\bullet$$\bullet$
    & \phantom{0}
    & $\bullet$ \\

    Do et al.~\cite{do2021novelty} 
    & 2021 
    & DL
    & AirSim, UE, FR-CNN, GPS
    & RGB
    & BB
    & \xmark
    &
    & Victims dataset
    & Acc
    & $\bullet$$\bullet$$\bullet$
    & \phantom{0}
    & $\bullet$ \\

    DeepGTAV~\cite{kiefer2022leveraging} 
    & 2022 
    & DL
    & EfficientDet-D0, YOLOv5
    & RGB
    & BB
    & \checkmark
    &
    & SDS, DGTA-SDS
    & mAP
    & $\bullet$$\bullet$$\bullet$
    & \phantom{0}
    & $\bullet$$\bullet$ \\

    Poudel et al.~\cite{poudel2023novel} 
    & 2023 
    & DL
    & AirSim, UE, ROS, YOLOv7
    & RGB
    & BB
    & \xmark
    &
    & Private
    & Acc 
    & $\bullet$$\bullet$$\bullet$
    & \phantom{0}
    & $\bullet$$\bullet$ \\

    POSEIDON~\cite{ruiz2023poseidon} 
    & 2023
    & DL
    & YOLOv5, YOLOv8
    & RGB
    & BB
    & \checkmark
    &
    & SDS
    & P, R, mAP 
    & $\bullet$$\bullet$
    & \phantom{0}
    & $\bullet$$\bullet$ \\

    SeaDronesSim~\cite{lin2023seadronesim} 
    & 2023
    & DL
    & Blender, FR-CNN
    & RGB
    & BB
    & \xmark
    &
    & Private
    & mAP 
    & $\bullet$$\bullet$$\bullet$
    & \phantom{0}
    & $\bullet$ \\

    Mohammadi et al.~\cite{khan2024investigating} 
    & 2024
    & DL
    & YOLO architectures
    & RGB
    & BB
    & \xmark
    &
    & Private
    & mAP 
    & $\bullet$$\bullet$$\bullet$
    & \phantom{0}
    & $\bullet$$\bullet$ \\
    
    SafeSea~\cite{tran2024safesea} 
    & 2024 
    & DL
    & YOLOv5, BLD
    & RGB
    & BB
    & \checkmark
    &
    & SDS
    & mAP 
    & $\bullet$$\bullet$
    & \phantom{0}
    & $\bullet$$\bullet$ \\

    Martinez et al.~\cite{martinez2025use} 
    & 2024 
    & DL
    & UE, YOLOv5
    & RGB
    & BB
    & \checkmark
    &
    & SDS
    & mAP, FPS, Params
    & $\bullet$$\bullet$
    & \phantom{0}
    & $\bullet$$\bullet$ \\

    \toprule[2pt] 
    \multicolumn{13}{l}{\textbf{\textit{General detection approaches applied to maritime SAR}}} \\ 
    \midrule

    SSPNet~\cite{hong2021sspnet} 
    & 2021
    & DL
    & FR-CNN
    & RGB
    & BB
    & \xmark
    &
    & TinyPerson
    & mAP 
    & $\bullet$$\bullet$
    & \phantom{0}
    & $\bullet$ \\

    DCLANet.~\cite{zhang2022finding} 
    & 2022 
    & DL
    & YOLOv5
    & RGB
    & BB
    & \xmark
    &
    & TinyPerson
    & P, R, mAP
    & $\bullet$$\bullet$$\bullet$
    & \phantom{0}
    & $\bullet$$\bullet$ \\

    SODNet~\cite{qi2022small} 
    & 2022
    & DL
    & ASPConv, FPN
    & RGB
    & BB
    & \xmark
    &
    & TinyPerson
    & mAP, MR, FPS
    & $\bullet$$\bullet$
    & \phantom{0}
    & $\bullet$$\bullet$$\bullet$ \\

    SF-YOLOv5~\cite{liu2022sf}
    & 2022
    & DL
    & YOLOv5, PB-FPN
    & RGB
    & BB
    & \xmark
    &
    & TinyPerson
    & mAP, FLOPs, Params, Time
    & $\bullet$
    & \phantom{0}
    & $\bullet$$\bullet$$\bullet$ \\

    MCGR~\cite{wang2022remote} 
    & 2022 
    & DL
    & YOLOv5, MCGR, SRGAN
    & RGB
    & BB
    & \xmark
    &
    & AFO
    & P, R, mAP 
    & $\bullet$$\bullet$$\bullet$
    & \phantom{0}
    & $\bullet$ \\

    YOLOv7-UAV~\cite{zeng2023yolov7} 
    & 2023
    & DL
    & YOLOv7, PANet
    & RGB
    & BB
    & \xmark
    &
    & TinyPerson
    & mAP 
    & $\bullet$
    & \phantom{0}
    & $\bullet$$\bullet$$\bullet$ \\

    LAI-YOLOv5s~\cite{deng2023lightweight}
    & 2023
    & DL
    & YOLOv5
    & RGB
    & BB
    & \xmark
    &
    & TinyPerson
    & mAP, FPS, speed
    & $\bullet$
    & \phantom{0}
    & $\bullet$$\bullet$$\bullet$ \\

    HANet~\cite{guo2023save}
    & 2023
    & DL
    & CenterNet, SSD
    & RGB
    & BB
    & \xmark
    &
    & TinyPerson
    & mAP, Params
    & $\bullet$$\bullet$
    & \phantom{0}
    & $\bullet$ \\

    TO-YOLOX~\cite{chen2023yolox} 
    & 2023 
    & DL
    & YOLOX, CBAM
    & RGB
    & BB
    & \xmark
    &
    & TinyPerson
    & mAP, FPS, Params
    & $\bullet$
    & \phantom{0}
    & $\bullet$$\bullet$$\bullet$ \\

    TOD-YOLOv7~\cite{tang2023long} 
    & 2023 
    & DL
    & YOLOv7
    & RGB
    & BB
    & \xmark
    &
    & TinyPerson
    & P, R, mAP, FPS, Params
    & $\bullet$
    & \phantom{0}
    & $\bullet$$\bullet$$\bullet$ \\

    GCRN~\cite{chen2023graphormer} 
    & 2023
    & DL
    & FR-CNN, Graphormer
    & RGB
    & BB
    & \xmark
    &
    & TinyPerson
    & mAP, MR
    & $\bullet$$\bullet$
    & \phantom{0}
    & $\bullet$ \\

    Swin-T-EFPNet~\cite{zhang2023efpnet}
    & 2023 
    & DL
    & Swin-T
    & RGB
    & BB
    & \xmark
    &
    & TinyPerson
    & mAP
    & $\bullet$$\bullet$$\bullet$
    & \phantom{0}
    & $\bullet$ \\

    Faster R-CNN-FSPN\cite{hu2023fspn}
    & 2023
    & DL
    & FR-CNN, FPN
    & RGB
    & BB
    & \xmark
    &
    & TinyPerson
    & mAP, MR
    & $\bullet$$\bullet$
    & \phantom{0}
    & $\bullet$ \\

    FG-RCNN~\cite{shao2024small}
    & 2024
    & DL
    & FR-CNN
    & RGB
    & BB
    & \xmark
    &
    & TinyPerson
    & mAP, MR
    & $\bullet$$\bullet$
    & \phantom{0}
    & $\bullet$ \\
    
\bottomrule[1pt]

\multicolumn{13}{c}{ 
    \makecell{ \footnotesize
        \footnotesize
        Abbreviations: 
            BB: Bounding Box;
            NNC: Nearest Neighbor Classifier;
            CS: Color Spaces;
            CCR: Correct Classification Rate;
            Hum.: Human;
            Bin.: Binary.
        \\
        \footnotesize
            Frac.: Fraction;
            Ellip.: Ellipses;
            Prob.: Probability;
            Swim.: Swimmers;
            Det.: Detections.
    }
}\\

\end{tabular}
}        
\end{table}

%% file: table_corpora.tex
\begin{table}[!ht]
    \centering
    \caption{Detailed comparison of features of the corpora available for maritime SAR grouped by real and synthetic datasets. The columns refer to: the name of the dataset and reference (\textbf{Dataset}); year of publication (\textbf{Year}); the format of the data if can be used for image-based detection, video, or both (\textbf{Format}); type of tasks including object detection (OD), tracking, and segmentation  (\textbf{Task}); the place where the images were captured (\textbf{Scenario}); image size (\textbf{Resolution}); number of images (\textbf{\# images}), classes (\textbf{\# classes}), annotations (\textbf{\# annots.}), and individuals (\textbf{\# persons}); and the availability of the corpus (\textbf{Avail.}).  Note: ``--'' indicates an unknown term.}
    \label{tab:datatable_sea}
    \renewcommand{\arraystretch}{0.925}
    \setlength{\tabcolsep}{3pt}
    \resizebox{\textwidth}{!}{
    \begin{tabular}{lc  cccccccccc}
    \toprule[1pt]

        \textbf{Dataset}
        &
        &\textbf{Year} 
        &\textbf{Format}
        &\textbf{Task}
        &\textbf{Scenario}
        &\textbf{Resolution}
        &\textbf{\# images}
        &\textbf{\# classes}  
        &\textbf{\# annots.} 
        &\textbf{\# persons}
        &\textbf{Avail.} \\ 
    \midrule
    \midrule

    \multicolumn{12}{l}{\textbf{\textit{Real}}} \\ 
    \midrule
 
    SeaDronesSee~\cite{varga2022seadronessee}    
            &
            & 2022  
            & Img,Vid
            & OD,Track
            & Sea
            & 1,280$\times$960 to  5,456$\times$3,632
            & $\sim$54,000
            & 5
            & $\sim$400,000
            & $\sim$150,000 
            & \href{https://seadronessee.cs.uni-tuebingen.de/}{Link} \\ 

    MOBDrone~\cite{cafarelli2022mobdrone}        
            &
            & 2022    
            & Img,Vid
            & OD,Track
            & Sea
            & 1,080$\times$1,080
            & 126,170
            & 5        
            & 181,689
            & 113,408
            & \href{https://aimh.isti.cnr.it/dataset/mobdrone/}{Link} \\

    Swimmers dataset~\cite{lygouras2019unsupervised}        
            &
            & 2019   
            & Img
            & OD
            & Sea, Coast
            & Up to 1,920$\times$1,080
            & 9,000
            & 1
            & --
            & --
            & \href{https://robotics.pme.duth.gr/swimmers_dataset/}{Link} \\ 

    AFO~\cite{gasienica2021ensemble}        
            &
            & 2021  
            & Img,Vid
            & OD
            & Sea, Coast
            & 1,280$\times$720 to 3,840$\times$2,160
            & 3,647
            & 6         
            & 39,991
            & 33,174
            &\href{https://www.kaggle.com/datasets/jangsienicajzkowy/afo-aerial-dataset-of-floating-objects}{Link} \\ 

     TinyPerson~\cite{yu2020scale}        
            &
            & 2020
            & Img
            & OD, Seg
            & Sea, Beach
            & 1,920$\times$1,080
            & 1,610
            & 3
            & 72,651
            & 42,256
            & \href{https://github.com/ucas-vg/PointTinyBenchmark/tree/master/dataset}{Link} \\ 

    SeaPerson (TinyPersonV2)\cite{yu2022object} 
            &
            & 2022 
            & Img
            & OD, Seg
            & Sea, Beach
            & 1,920$\times$1,080
            & 12,032
            & 3
            & 619,627 
            & -- 
            & \href{https://github.com/ucas-vg/PointTinyBenchmark/tree/master/dataset}{Link} \\

    \toprule[1pt] 
    \multicolumn{12}{l}{\textbf{\textit{Synthetic}}} \\ 

    \midrule

    Victims on Ocean~\cite{do2021novelty}
            &
            & 2021    
            & Img
            & OD
            & Sea
            &  512$\times$288
            & 3,000
            & 2         
            & -- 
            & --
            & \href{https://github.com/winter2897/Airsim-Search-and-Rescue-SAR-at-sea-with-UAV/blob/master/README.md}{Link} \\ 

    DGTA-SeaDronesSee~\cite{kiefer2022leveraging}
            &
            & 2022 
            & Img
            & OD
            & Sea
            & 3,840$\times$3,840
            & 100,000
            & 5           
            & --
            & --
            & \href{https://cogsys.cs.uni-tuebingen.de/webprojects/DGTA-2023-Datasets/}{Link} \\ 

    SynBASe~\cite{martinez2025use}
            &
            & 2024  
            & Img
            & OD 
            & Sea
            & 1,200$\times$1,200
            & 1,295
            & 1          
            & 3,415 
            & 3,415 
            & \href{https://www.dlsi.ua.es/~jgallego/datasets/synbase/index2.html}{Link} \\ 
            
\bottomrule[1pt]
\end{tabular}
}        
\end{table}

%% file: table_results_image.tex
\begin{table}[!ht]
    \centering
    \caption{
    Comparison of results from state-of-the-art object detection methods on the most widely used public datasets. The methods are organized according to the dataset on which they have been evaluated, including (in columns) their name and reference (\textbf{Method}), their base architecture (\textbf{Base arch.}), information about the model input (\textbf{Input}), and results in terms of key metrics: mAP@0.5, mAP@0.5-0.95, Frames per Second (\textbf{FPS}), and number of parameters in millions (\textbf{Params (M)})---an indicator of the complexity of the methods and resource demands. Two additional ratios are included: the accuracy ratio (\textbf{Acc.}), calculated as mAP@0.5 divided by the number of parameters, and the efficiency ratio (\textbf{Eff.}), calculated as FPS divided by the number of parameters. These ratios summarize the performance of each method in relation to its size (higher values are better).   
    }
    \label{tab:results_image}
    \renewcommand{\arraystretch}{0.75}
    \setlength{\tabcolsep}{2pt}
    \resizebox{\textwidth}{!}{
    \begin{tabular}{l lcc l ccccc l cc}
    \toprule[1pt]
    
    \multicolumn{2}{l}{\textbf{\textit{Dataset}}} & \multicolumn{2}{c}{\textit{\textbf{Method information}}} &&  \multicolumn{5}{c}{\textit{\textbf{Metrics}}} && \multicolumn{2}{c}{\textit{\textbf{Ratio}}}  \\
    
     \cmidrule{3-4} \cmidrule{6-10} \cmidrule{12-13} 
     
        \phantom{----} 
        & \textbf{Method} 
        & \textbf{Base arch.}
        & \textbf{Input} 
        &
        & \textbf{F1} 
        & \textbf{\textbf{mAP@0.5}}
        & \textbf{\textbf{mAP@0.5-0.95}}
        & \textbf{FPS}
        & \textbf{Params (M)}
        &
        & \textbf{Acc.} 
        & \textbf{Eff.} 
        \\
    \midrule
    \midrule

    \multicolumn{2}{l}{\textbf{SeaDronesSee \cite{varga2022seadronessee}}} \\ 
               
        &   Sea-YOLOv5s\cite{wang2023sea}
            & YOLOv5s
            & $640\times640$
            &
            & --
            & 58.2         
            & 33.7
            & --
            & -- 
            &
            & --    
            & --    
            \\ 
            
        &   YOLOv7-sea~\cite{zhao2023yolov7}          
            & YOLOv7
            & $2400\times2400$
            &
            & --
            & 90.7          
            & 59.0
            & 1.0
            & -- 
            &
            & --    
            & --    
            \\        
    
            
        &   ABT-YOLOv7~\cite{zhang2023enhanced}\textsuperscript{\ddag}         
            & YOLOv7                   
            & $1280\times1280$ 
            &
            & 92.0*  
            & 91.6 
            & --
            & 7.5
            & 52.4 
            &
            & 1.8    
            & 0.1    
            \\

        &   DetectoRS~\cite{kiefer20231st}         
            & CR-CNN                   
            & Multiple resol.
            &
            & --
            & 90.0 
            & 60.0
            & 1.0
            & -- 
            &
            & --    
            & --    
            \\ 

        &   Maritime-VSA~\cite{kiefer20231st}         
            & DB-Swin-S                    
            & $640\times640$          
            &        
            & --
            & 91.0 
            & 62.0
            & 1.0
            & -- 
            &
            & --    
            & --    
            \\

        &   DyHead~\cite{dai2021dynamic}         
            & Swin-L                   
            & $640\times640$          
            &        
            & --
            & 89.0 
            & 57.0
            & 1.0
            & -- 
            &
            & --    
            & --    
            \\

        &   YOLOv5s-SwinDS~\cite{liu2024yolov5s}         
            & Swin-T                    
            & Multiple resol.
            &         
            & 81.4*
            & 79.1
            & 42.9
            & --
            & -- 
            &
            & --    
            & --    
            \\

        &   YOLOv7-FSB~\cite{zhang2023lightweight}\textsuperscript{\ddag}
            & YOLOv7                    
            & $1280\times1280$          
            &        
            & --       
            & 89.2       
            & --       
            & 96.5     
            & 5.8 
            &
            & 15.3    
            & 16.6    
            \\    

        &   Fernandes et al.~\cite{fernandes2023enhancing}  
            & YOLOv7                    
            & $1280\times1280$         
            &         
            & --       
            & 74.6       
            & 44.0       
            & --       
            & 14.17 
            &
            & 5.2    
            & --    
            \\

        &   Memory Map\cite{kiefer2023memory}           
            & YOLOv7-Tiny
            & --         
            & 
            & --
            & 72.8            
            & -- 
            & 25.1
            & -- 
            &
            & --    
            & --    
            \\

        &   Zhao et al.~\cite{zhao2024heuristic} 
            & FR-CNN, CR-CNN
            & --
            &         
            & --       
            & 71.8       
            & 45.4       
            & --       
            & -- 
            &
            & --    
            & --    
            \\

    \midrule

    \multicolumn{2}{l}{\textbf{MOBDrone~\cite{cafarelli2022mobdrone}}} \\ 
        &   YOLO-BEV~\cite{yang2023high}           
            & YOLOv5   
            & --         
            &
            & --           
            & 97.1  
            & 56.4
            & 48.0 
            & 6.9 
            &
            & 14.1    
            & 7.0    
            \\
            
        &   ABT-YOLOv7~\cite{zhang2023enhanced}\textsuperscript{\ddag} 
            & YOLOv7                   
            & $1280\times1280$            
            &       
            & 92.0* 
            & 91.6 
            & --
            & 7.5
            & 52.4 
            &
            & 1.8    
            & 0.1    
            \\

        &   YOLOv7-FSB~\cite{zhang2023lightweight}\textsuperscript{\ddag}
            & YOLOv7                    
            & $1280\times1280$         
            &         
            & --       
            & 89.2       
            & --       
            & 96.5       
            & 5.8 
            &
            & 15.3    
            & 16.6    
            \\

    \midrule

    \multicolumn{2}{l}{\textbf{AFO~\cite{gasienica2021ensemble}}} \\ 
        &   Gasienica et al.\cite{gasienica2021ensemble}  
            & FR-CNN, RetinaNet
            & $1333\times750$          
            &     
            & --  
            & 82.2  
            & 37.7
            & --  
            & -- 
            &
            & --    
            & --    
            \\
            
        &   SG-Det\cite{zhang2023sg}  
            & GhostNet
            & $416\times416$          
            &     
            & --  
            & 87.5 
            & --
            & 31.9
            & -- 
            &
            & --    
            & --    
            \\

        &   MCGR~\cite{wang2022remote}  
            & EDSR
            & $1000\times1000$         
            &       
            & --  
            & 84.5 
            & --
            & --  
            & -- 
            &
            & --    
            & --    
            \\

        &   YOLOv7-CSAW~\cite{zhu2023yolov7}  
            & YOLOv7
            & $1280\times1280$         
            &      
            & --  
            & 86.4  
            & --
            & --  
            & 58.7 
            &
            & 1.5    
            & --    
            \\

        &   Zhang et al.~\cite{zhang2023semi}  
            & YOLOv5s
            & $1536\times1536$         
            &   
            & 94.5*  
            & 96.3  
            & --   
            & --  
            & --   
            &
            & --    
            & --    
            \\

        &   Lu et al.~\cite{lu2023high}  
            & DINO
            & $448\times448$         
            &   
            & --  
            & 89.7  
            & 57.6   
            & --  
            & 57.7  
            &
            & 1.6    
            & --    
            \\

    \midrule

        \multicolumn{2}{l}{\textbf{TinyPerson~\cite{yu2020scale}}} \\ 
            
            & Gao et al.~\cite{gao2022multiscale}
            & Swin-T
            & $640\times512$         
            &  
            & --  
            & 62.0  
            & --   
            & --  
            & 51.6  
            &
            & 1.2    
            & --    
            \\

        &   TEFF~\cite{zhou2021texture}  
            & FR-CNN
            & $640\times512$         
            &   
            & --  
            & 56.3  
            & --   
            & --  
            & --  
            &
            & --    
            & --    
            \\

        &   Zhang et al.~\cite{zhang2023semi}
            & YOLOv5s
            & $1536\times1536$         
            &       
            & 65.4*  
            & 67.5  
            & --
            & --  
            & -- 
            &
            & --    
            & --    
            \\

        &   DCLANet~\cite{zhang2022finding}\textsuperscript{$\dagger$}  
            & YOLOv5
            & $1600\times1600$         
            & 
            & 63.7*  
            & 66.6 
            & 27.2
            & --
            & -- 
            &
            & --    
            & --    
            \\

            & SODNet~\cite{qi2022small}\textsuperscript{$\dagger$}  
            & YOLOv5s
            & $640\times640$          
            &  
            & --  
            & 55.6  
            & 17.1   
            & 91  
            & --  
            &
            & --    
            & --    
            \\

        &   SF-YOLOv5~\cite{liu2022sf}\textsuperscript{$\dagger$}  
            & YOLOv5
            & $640\times640$          
            & 
            & --  
            & 20.0  
            & 6.5   
            & --  
            & 2.2  
            &
            & 9.0    
            & --    
            \\

        &   SSPNet~\cite{hong2021sspnet}\textsuperscript{$\dagger$}  
            & FR-CNN
            & $640\times512$          
            & 
            & --  
            & 59.1  
            & --   
            & --  
            & --  
            &
            & --    
            & --    
            \\

        &   YOLOv7-UAV~\cite{zeng2023yolov7}\textsuperscript{$\dagger$}  
            & YOLOv7
            & $960\times960$          
            & 
            & --  
            & 20.6  
            & 6.0   
            & 86.0  
            & 3.1 
            &
            & 6.6    
            & 27.7    
            \\

        &   LAI-YOLOv5s~\cite{deng2023lightweight}\textsuperscript{$\dagger$}  
            & YOLOv5
            & $640\times640$          
            &
            & -- 
            & 20.8  
            & 6.5   
            & 94.3  
            & 6.3   
            &
            & 3.3    
            & 15.0    
            \\

        &   HANet~\cite{guo2023save}\textsuperscript{$\dagger$}  
            & CenterNet
            & $640\times512$           
            &
            & --  
            & 58.5  
            & --   
            & --  
            & 24.4  
            &
            & 2.4    
            & --    
            \\

        &   TO-YOLOX~\cite{chen2023yolox}\textsuperscript{$\dagger$}  
            & YOLOX
            & $640\times640$           
            &
            & --  
            & 29.2  
            & --   
            & 103.0  
            & 6.7  
            &
            & 4.4    
            & 15.4    
            \\

        &   FG-RCNN~\cite{shao2024small}\textsuperscript{$\dagger$}  
            & FR-CNN
            & $640\times640$           
            &
            & --  
            & 56.8  
            & --   
            & --  
            & --  
            &
            & --    
            & --    
            \\

        &   TOD-YOLOv7~\cite{tang2023long}\textsuperscript{$\dagger$}
            & YOLOv7
            & --            
            &
            & 39.2*  
            & 30.0  
            & 9.5   
            & 208.0  
            & 38.0  
            &
            & 0.8    
            & 5.5    
            \\

            &   GCRN~\cite{chen2023graphormer}\textsuperscript{$\dagger$}  
            & FR-CNN
            & $512\times512$           
            & 
            & --  
            & 53.7  
            & --   
            & --  
            & --  
            &
            & --    
            & --    
            \\

        &   Swin-T-EFPNet~\cite{zhang2023efpnet}\textsuperscript{$\dagger$}  
            & Swin T
            & $680\times510$          
            & 
            & --  
            & 60.7  
            & --   
            & --  
            & --  
            &
            & --    
            & --    
            \\

        &   Faster R-CNN-FSPN~\cite{hu2023fspn}\textsuperscript{$\dagger$}  
            & FR-CNN
            & --           
            &
            & --  
            & 57.9  
            & --   
            & --  
            & --  
            &
            & --    
            & --    
            \\

    \midrule

    \multicolumn{2}{l}{\textbf{Swimmers~\cite{lygouras2019unsupervised}}} \\ 
        
        &   Tiny YOLOv3~\cite{lygouras2019unsupervised}  
            & Tiny YOLOv3
            & $608\times608$         
            &
            & 68.0 
            & 67.0 
            & --
            & 12.0
            & --
            &
            & --    
            & --    
            \\

        &   YOLOv4 Large~\cite{rizk2022towards}\textsuperscript{\ddag}  
            & YOLOv4
            & $608\times608$          
            & 
            & 70.4  
            & 69.4  
            & --   
            & 5.7  
            & --  
            &
            & --    
            & --    
            \\

        &   YOLOv4 Tiny~\cite{rizk2022towards}\textsuperscript{\ddag}  
            & YOLOv4
            & $608\times608$           
            & 
            & 62.7  
            & 62.5  
            & --   
            & 45.6  
            & --   
            &
            & --    
            & --    
            \\

        &   Pruned YOLOv4~\cite{rizk2022optimization}\textsuperscript{\ddag} 
            & YOLOv4
            & $608\times608$           
            &
            & 70.5  
            & 72.1  
            & --   
            & 69.0  
            & 1.03  
            &
            & 70.0    
            & 67.0    
            \\
            
\bottomrule[1pt]

\multicolumn{13}{c}{ 
    \makecell{ \footnotesize
        \footnotesize
        Notes: 
        ``–'' denotes an unspecified value; 
        ``\textsuperscript{$\dagger$}''  indicates methods not specific to maritime person detection; 
        ``\textsuperscript{\ddag}'' marks methods that combine datasets;
        \\
        \footnotesize
        and ``*" in the F$_1$ metric signifies a value manually calculated from the provided precision and recall.
    }
}

\end{tabular}
}        
\end{table}

%% file: table_results_video.tex

\begin{table}[!ht]
    \centering
    \caption{Detailed results obtained with different methods based on tracking in the two existing public databases for this task. Note: ``--'' denotes a non-specified value and ``\textsuperscript{\ddag}'' indicates that the method mixes images from various datasets.}
    \label{tab:results_video}
    \renewcommand{\arraystretch}{0.925}
    \setlength{\tabcolsep}{2pt}
    \resizebox{\textwidth}{!}{
    \begin{tabular}{l ccc l cccccccc}
    
    \toprule[1pt]
    
    & \multicolumn{3}{c}{\textit{\textbf{Method}}} 
        & &  \multicolumn{8}{c}{\textit{\textbf{Metrics}}}  \\
    
    \cmidrule{2-4} \cmidrule{6-13} 

    \textbf{\textit{Corpora}} 
        & \textbf{Name}
        & \textbf{Base architecture}
        & \textbf{Input}
        &
        & \textbf{HOTA}
        & \textbf{MOTA} 
        & \textbf{MOTP} 
        & \textbf{IDs} 
        & \textbf{IDF$_1$}
        & \textbf{R}
        & \textbf{Frags}
        & \textbf{FPS} \\ 
    \midrule
    \midrule

    \multirow{5}{*}{SeaDronesSee \cite{varga2022seadronessee}}    
            
        &   YOLOv7-FSB~\cite{zhang2023lightweight}\textsuperscript{\ddag}
            & YOLOv7
            & $1280\times1280$ 
            &
            & --     
            & 87.6       
            & -- 
            & 18 
            & -- 
            & --       
            & --       
            & 96.5 \\    

        &   ReIDTracker\_Sea~\cite{huang2023reidtracker}  
            & Swin-T 
            & $2880\times1920$
            &
            & 62.4 
            & 78.1 
            & 20.4 
            & 178 
            & 71.3 
            & 88.3 
            & 1343 
            & 3 \\ 

        &   Memory Map~\cite{kiefer2023memory}           
            & YOLOv7-Tiny 
            & $3840\times2160$
            &
            & 67.2 
            & 80.8 
            & -- 
            & 35 
            & -- 
            & -- 
            & 721 
            & 27 \\ 

        &   Kiefer et al.~\cite{kiefer2023fast}           
            & Autoencoder 
            & --
            &
            & -- 
            & -- 
            & -- 
            & --   
            & -- 
            & 86.3 
            & -- 
            & 48 \\ 

        &   MG-MOT~\cite{yang2024sea}           
            & YOLOv8-x 
            & $1280\times1280$ 
            &
            & 69.5 
            & 77.9 
            & 20.7 
            & 18   
            & 85.9 
            & 88.1   
            & 784   
            & -- \\ 
    
    \midrule

    MOBDrone~\cite{cafarelli2022mobdrone}
    
        &   YOLOv7-FSB~\cite{zhang2023lightweight}\textsuperscript{\ddag}
            & YOLOv7 
            & $1280\times1280$
            &
            & --     
            & 87.6       
            & --       
            & 18   
            & -- 
            & --       
            & --       
            & 96.5 \\    
            
\bottomrule[1pt]
\end{tabular}
}        
\end{table}

%% file: paper.bbl
\begin{thebibliography}{176}
\expandafter\ifx\csname natexlab\endcsname\relax\def\natexlab#1{#1}\fi
\providecommand{\url}[1]{\texttt{#1}}
\providecommand{\href}[2]{#2}
\providecommand{\path}[1]{#1}
\providecommand{\DOIprefix}{doi:}
\providecommand{\ArXivprefix}{arXiv:}
\providecommand{\URLprefix}{URL: }
\providecommand{\Pubmedprefix}{pmid:}
\providecommand{\doi}[1]{\href{http://dx.doi.org/#1}{\path{#1}}}
\providecommand{\Pubmed}[1]{\href{pmid:#1}{\path{#1}}}
\providecommand{\bibinfo}[2]{#2}
\ifx\xfnm\relax \def\xfnm[#1]{\unskip,\space#1}\fi
\bibitem[{Al-Jarrah et~al.(2015)Al-Jarrah, Yoo, Muhaidat, Karagiannidis \&
  Taha}]{al2015efficient}
\bibinfo{author}{Al-Jarrah, O.~Y.}, \bibinfo{author}{Yoo, P.~D.},
  \bibinfo{author}{Muhaidat, S.}, \bibinfo{author}{Karagiannidis, G.~K.}, \&
  \bibinfo{author}{Taha, K.} (\bibinfo{year}{2015}).
\newblock \bibinfo{title}{Efficient machine learning for big data: A review}.
\newblock {\it \bibinfo{journal}{Big Data Research}\/},  {\it
  \bibinfo{volume}{2}\/}, \bibinfo{pages}{87--93}.
\bibitem[{Al-Mter \& Lu(2016)}]{al2016modified}
\bibinfo{author}{Al-Mter, A.~H.}, \& \bibinfo{author}{Lu, S.-F.}
  (\bibinfo{year}{2016}).
\newblock \bibinfo{title}{A modified particle swarm optimization algorithm
  using uniform design}.
\newblock In {\it \bibinfo{booktitle}{2016 International Conference on
  Information System and Artificial Intelligence (ISAI)}\/} (pp.
  \bibinfo{pages}{432--435}).
\newblock \bibinfo{organization}{IEEE}.
\bibitem[{Andriyanov \& Andriyanov(2020)}]{andriyanov2020using}
\bibinfo{author}{Andriyanov, N.}, \& \bibinfo{author}{Andriyanov, D.}
  (\bibinfo{year}{2020}).
\newblock \bibinfo{title}{The using of data augmentation in machine learning in
  image processing tasks in the face of data scarcity}.
\newblock In {\it \bibinfo{booktitle}{Journal of Physics: Conference Series}\/}
  (p. \bibinfo{pages}{012018}).
\newblock \bibinfo{organization}{IOP Publishing}.
\bibitem[{Bai et~al.(2023)Bai, Bai \& Wu}]{bai2023review}
\bibinfo{author}{Bai, C.}, \bibinfo{author}{Bai, X.}, \& \bibinfo{author}{Wu,
  K.} (\bibinfo{year}{2023}).
\newblock \bibinfo{title}{A review: Remote sensing image object detection
  algorithm based on deep learning}.
\newblock {\it \bibinfo{journal}{Electronics}\/},  {\it
  \bibinfo{volume}{12}\/}, \bibinfo{pages}{4902}.
\bibitem[{Bai et~al.(2022)Bai, Dai, Wang \& Yang}]{bai2022detection}
\bibinfo{author}{Bai, J.}, \bibinfo{author}{Dai, J.}, \bibinfo{author}{Wang,
  Z.}, \& \bibinfo{author}{Yang, S.} (\bibinfo{year}{2022}).
\newblock \bibinfo{title}{A detection method of the rescue targets in the
  marine casualty based on improved {YOLOv5s}}.
\newblock {\it \bibinfo{journal}{Frontiers in Neurorobotics}\/},  {\it
  \bibinfo{volume}{16}\/}, \bibinfo{pages}{1053124}.
\bibitem[{Bansal et~al.(2022)Bansal, Sharma \& Kathuria}]{bansal2022systematic}
\bibinfo{author}{Bansal, M.~A.}, \bibinfo{author}{Sharma, D.~R.}, \&
  \bibinfo{author}{Kathuria, D.~M.} (\bibinfo{year}{2022}).
\newblock \bibinfo{title}{A systematic review on data scarcity problem in deep
  learning: solution and applications}.
\newblock {\it \bibinfo{journal}{ACM Computing Surveys (CSUR)}\/},  {\it
  \bibinfo{volume}{54}\/}, \bibinfo{pages}{1--29}.
\bibitem[{Bejiga et~al.(2017)Bejiga, Zeggada, Nouffidj \&
  Melgani}]{bejiga2017convolutional}
\bibinfo{author}{Bejiga, M.~B.}, \bibinfo{author}{Zeggada, A.},
  \bibinfo{author}{Nouffidj, A.}, \& \bibinfo{author}{Melgani, F.}
  (\bibinfo{year}{2017}).
\newblock \bibinfo{title}{A convolutional neural network approach for assisting
  avalanche search and rescue operations with {UAV} imagery}.
\newblock {\it \bibinfo{journal}{Remote Sensing}\/},  {\it
  \bibinfo{volume}{9}\/}, \bibinfo{pages}{100}.
\bibitem[{Berg et~al.(2022)Berg, Santana~Maia, Pham \&
  Lef{\`e}vre}]{berg2022weakly}
\bibinfo{author}{Berg, P.}, \bibinfo{author}{Santana~Maia, D.},
  \bibinfo{author}{Pham, M.-T.}, \& \bibinfo{author}{Lef{\`e}vre, S.}
  (\bibinfo{year}{2022}).
\newblock \bibinfo{title}{Weakly supervised detection of marine animals in high
  resolution aerial images}.
\newblock {\it \bibinfo{journal}{Remote Sensing}\/},  {\it
  \bibinfo{volume}{14}\/}, \bibinfo{pages}{339}.
\bibitem[{Bhuiya et~al.(2022)Bhuiya, Islam, Drishty, Akash, Saha, Chakrabarty
  \& Hossain}]{bhuiya2022surveillance}
\bibinfo{author}{Bhuiya, M. S.~R.}, \bibinfo{author}{Islam, N.},
  \bibinfo{author}{Drishty, A.~S.}, \bibinfo{author}{Akash, U.~D.},
  \bibinfo{author}{Saha, S.~S.}, \bibinfo{author}{Chakrabarty, A.}, \&
  \bibinfo{author}{Hossain, S.} (\bibinfo{year}{2022}).
\newblock \bibinfo{title}{Surveillance in maritime scenario using deep learning
  and swarm intelligence}.
\newblock In {\it \bibinfo{booktitle}{2022 25th International Conference on
  Computer and Information Technology (ICCIT)}\/} (pp.
  \bibinfo{pages}{569--574}).
\newblock \bibinfo{organization}{IEEE}.
\bibitem[{Bochkovskiy et~al.(2020)Bochkovskiy, Wang \&
  Liao}]{bochkovskiy2020yolov4}
\bibinfo{author}{Bochkovskiy, A.}, \bibinfo{author}{Wang, C.-Y.}, \&
  \bibinfo{author}{Liao, H.-Y.~M.} (\bibinfo{year}{2020}).
\newblock \bibinfo{title}{Yolov4: Optimal speed and accuracy of object
  detection}.
\newblock {\it \bibinfo{journal}{arXiv preprint arXiv:2004.10934}\/}, .
\bibitem[{Breivik et~al.(2013)Breivik, Allen, Maisondieu \&
  Olagnon}]{breivik2013advances}
\bibinfo{author}{Breivik, {\O}.}, \bibinfo{author}{Allen, A.~A.},
  \bibinfo{author}{Maisondieu, C.}, \& \bibinfo{author}{Olagnon, M.}
  (\bibinfo{year}{2013}).
\newblock \bibinfo{title}{Advances in search and rescue at sea}.
\newblock {\it \bibinfo{journal}{Ocean Dynamics}\/},  {\it
  \bibinfo{volume}{63}\/}, \bibinfo{pages}{83--88}.
\bibitem[{Cafarelli et~al.(2022)Cafarelli, Ciampi, Vadicamo, Gennaro, Berton,
  Paterni, Benvenuti, Passera \& Falchi}]{cafarelli2022mobdrone}
\bibinfo{author}{Cafarelli, D.}, \bibinfo{author}{Ciampi, L.},
  \bibinfo{author}{Vadicamo, L.}, \bibinfo{author}{Gennaro, C.},
  \bibinfo{author}{Berton, A.}, \bibinfo{author}{Paterni, M.},
  \bibinfo{author}{Benvenuti, C.}, \bibinfo{author}{Passera, M.}, \&
  \bibinfo{author}{Falchi, F.} (\bibinfo{year}{2022}).
\newblock \bibinfo{title}{{MOBDrone}: A drone video dataset for man overboard
  rescue}.
\newblock In {\it \bibinfo{booktitle}{International Conference on Image
  Analysis and Processing}\/} (pp. \bibinfo{pages}{633--644}).
\newblock \bibinfo{organization}{Springer}.
\bibitem[{Cai \& Vasconcelos(2018)}]{cai2018cascade}
\bibinfo{author}{Cai, Z.}, \& \bibinfo{author}{Vasconcelos, N.}
  (\bibinfo{year}{2018}).
\newblock \bibinfo{title}{Cascade r-cnn: Delving into high quality object
  detection}.
\newblock In {\it \bibinfo{booktitle}{Proceedings of the IEEE Conference on
  Computer Vision and Pattern Recognition}\/} (pp.
  \bibinfo{pages}{6154--6162}).
\bibitem[{Carion et~al.(2020)Carion, Massa, Synnaeve, Usunier, Kirillov \&
  Zagoruyko}]{carion2020end}
\bibinfo{author}{Carion, N.}, \bibinfo{author}{Massa, F.},
  \bibinfo{author}{Synnaeve, G.}, \bibinfo{author}{Usunier, N.},
  \bibinfo{author}{Kirillov, A.}, \& \bibinfo{author}{Zagoruyko, S.}
  (\bibinfo{year}{2020}).
\newblock \bibinfo{title}{End-to-end object detection with transformers}.
\newblock In {\it \bibinfo{booktitle}{European Conference on Computer
  Vision}\/} (pp. \bibinfo{pages}{213--229}).
\newblock \bibinfo{organization}{Springer}.
\bibitem[{Cazzato et~al.(2020)Cazzato, Cimarelli, Sanchez-Lopez, Voos \&
  Leo}]{cazzato2020survey}
\bibinfo{author}{Cazzato, D.}, \bibinfo{author}{Cimarelli, C.},
  \bibinfo{author}{Sanchez-Lopez, J.~L.}, \bibinfo{author}{Voos, H.}, \&
  \bibinfo{author}{Leo, M.} (\bibinfo{year}{2020}).
\newblock \bibinfo{title}{A survey of computer vision methods for 2d object
  detection from unmanned aerial vehicles}.
\newblock {\it \bibinfo{journal}{Journal of Imaging}\/},  {\it
  \bibinfo{volume}{6}\/}, \bibinfo{pages}{78}.
\bibitem[{Ch et~al.(2023)Ch, Navena et~al.}]{ch2023classification}
\bibinfo{author}{Ch, R.}, \bibinfo{author}{Navena, M.} et~al.
  (\bibinfo{year}{2023}).
\newblock \bibinfo{title}{Classification and segmentation of marine related
  remote sensing imagery data using deep learning}.
\newblock In {\it \bibinfo{booktitle}{2023 2nd International Conference on
  Vision Towards Emerging Trends in Communication and Networking Technologies
  (ViTECoN)}\/} (pp. \bibinfo{pages}{1--5}).
\newblock \bibinfo{organization}{IEEE}.
\bibitem[{Chen et~al.(2023{\natexlab{a}})Chen, Li, Ou, Hu \&
  Peng}]{chen2023graphormer}
\bibinfo{author}{Chen, J.}, \bibinfo{author}{Li, X.}, \bibinfo{author}{Ou, Y.},
  \bibinfo{author}{Hu, X.}, \& \bibinfo{author}{Peng, T.}
  (\bibinfo{year}{2023}{\natexlab{a}}).
\newblock \bibinfo{title}{Graphormer-based contextual reasoning network for
  small object detection}.
\newblock In {\it \bibinfo{booktitle}{Chinese Conference on Pattern Recognition
  and Computer Vision (PRCV)}\/} (pp. \bibinfo{pages}{294--305}).
\newblock \bibinfo{organization}{Springer}.
\bibitem[{Chen et~al.(2023{\natexlab{b}})Chen, Liang, Yu, Xu, Ji, Zhang, Zhang,
  Cui, He, Chang et~al.}]{chen2023yolox}
\bibinfo{author}{Chen, Z.}, \bibinfo{author}{Liang, Y.}, \bibinfo{author}{Yu,
  Z.}, \bibinfo{author}{Xu, K.}, \bibinfo{author}{Ji, Q.},
  \bibinfo{author}{Zhang, X.}, \bibinfo{author}{Zhang, Q.},
  \bibinfo{author}{Cui, Z.}, \bibinfo{author}{He, Z.}, \bibinfo{author}{Chang,
  R.} et~al. (\bibinfo{year}{2023}{\natexlab{b}}).
\newblock \bibinfo{title}{{TO}--{YOLOX}: a pure {CNN} tiny object detection
  model for remote sensing images}.
\newblock {\it \bibinfo{journal}{International Journal of Digital Earth}\/},
  {\it \bibinfo{volume}{16}\/}, \bibinfo{pages}{3882--3904}.
\bibitem[{Chien et~al.(2022)Chien, Chen \& Chen}]{chien2022study}
\bibinfo{author}{Chien, T.-Y.}, \bibinfo{author}{Chen, C.-F.}, \&
  \bibinfo{author}{Chen, Z.-Y.} (\bibinfo{year}{2022}).
\newblock \bibinfo{title}{Study of taiwanese white dolphins detection and
  tracking techniques by utilizing autonomous unmanned aerial vehicles}.
\newblock In {\it \bibinfo{booktitle}{OCEANS 2022, Hampton Roads}\/} (pp.
  \bibinfo{pages}{1--10}).
\newblock \bibinfo{organization}{IEEE}.
\bibitem[{Chollet(2017)}]{chollet2017xception}
\bibinfo{author}{Chollet, F.} (\bibinfo{year}{2017}).
\newblock \bibinfo{title}{Xception: Deep learning with depthwise separable
  convolutions}.
\newblock In {\it \bibinfo{booktitle}{Proceedings of the IEEE conference on
  computer vision and pattern recognition}\/} (pp.
  \bibinfo{pages}{1251--1258}).
\bibitem[{Cruz \& Bernardino(2016)}]{cruz2016aerial}
\bibinfo{author}{Cruz, G.}, \& \bibinfo{author}{Bernardino, A.}
  (\bibinfo{year}{2016}).
\newblock \bibinfo{title}{Aerial detection in maritime scenarios using
  convolutional neural networks}.
\newblock In {\it \bibinfo{booktitle}{International Conference on Advanced
  Concepts for Intelligent Vision Systems}\/} (pp. \bibinfo{pages}{373--384}).
\newblock \bibinfo{organization}{Springer}.
\bibitem[{Dai et~al.(2021)Dai, Chen, Xiao, Chen, Liu, Yuan \&
  Zhang}]{dai2021dynamic}
\bibinfo{author}{Dai, X.}, \bibinfo{author}{Chen, Y.}, \bibinfo{author}{Xiao,
  B.}, \bibinfo{author}{Chen, D.}, \bibinfo{author}{Liu, M.},
  \bibinfo{author}{Yuan, L.}, \& \bibinfo{author}{Zhang, L.}
  (\bibinfo{year}{2021}).
\newblock \bibinfo{title}{Dynamic head: Unifying object detection heads with
  attentions}.
\newblock In {\it \bibinfo{booktitle}{Proceedings of the IEEE/CVF conference on
  computer vision and pattern recognition}\/} (pp.
  \bibinfo{pages}{7373--7382}).
\bibitem[{Deng et~al.(2023)Deng, Bi, Li, Chen, Duan, Lou, Zhang, Bi \&
  Liu}]{deng2023lightweight}
\bibinfo{author}{Deng, L.}, \bibinfo{author}{Bi, L.}, \bibinfo{author}{Li, H.},
  \bibinfo{author}{Chen, H.}, \bibinfo{author}{Duan, X.}, \bibinfo{author}{Lou,
  H.}, \bibinfo{author}{Zhang, H.}, \bibinfo{author}{Bi, J.}, \&
  \bibinfo{author}{Liu, H.} (\bibinfo{year}{2023}).
\newblock \bibinfo{title}{Lightweight aerial image object detection algorithm
  based on improved {YOLOv5s}}.
\newblock {\it \bibinfo{journal}{Scientific Reports}\/},  {\it
  \bibinfo{volume}{13}\/}, \bibinfo{pages}{7817}.
\bibitem[{Dinnbier et~al.(2017)Dinnbier, Thueux, Savvaris \&
  Tsourdos}]{dinnbier2017target}
\bibinfo{author}{Dinnbier, N.~M.}, \bibinfo{author}{Thueux, Y.},
  \bibinfo{author}{Savvaris, A.}, \& \bibinfo{author}{Tsourdos, A.}
  (\bibinfo{year}{2017}).
\newblock \bibinfo{title}{Target detection using gaussian mixture models and
  fourier transforms for {UAV} maritime search and rescue}.
\newblock In {\it \bibinfo{booktitle}{2017 International Conference on Unmanned
  Aircraft Systems (ICUAS)}\/} (pp. \bibinfo{pages}{1418--1424}).
\newblock \bibinfo{organization}{IEEE}.
\bibitem[{Do~Trong et~al.(2021)Do~Trong, Hai, Duc \& Thanh}]{do2021novelty}
\bibinfo{author}{Do~Trong, T.}, \bibinfo{author}{Hai, Q.~T.},
  \bibinfo{author}{Duc, N.~T.}, \& \bibinfo{author}{Thanh, H.~T.}
  (\bibinfo{year}{2021}).
\newblock \bibinfo{title}{A novelty approach to emulate field data captured by
  unmanned aerial vehicles for training deep learning algorithms used for
  search-and-rescue activities at sea}.
\newblock In {\it \bibinfo{booktitle}{2020 IEEE Eighth International Conference
  on Communications and Electronics (ICCE)}\/} (pp. \bibinfo{pages}{288--293}).
\newblock \bibinfo{organization}{IEEE}.
\bibitem[{Dong et~al.(2021)Dong, Ota \& Dong}]{dong2021uav}
\bibinfo{author}{Dong, J.}, \bibinfo{author}{Ota, K.}, \&
  \bibinfo{author}{Dong, M.} (\bibinfo{year}{2021}).
\newblock \bibinfo{title}{{UAV}-based real-time survivor detection system in
  post-disaster search and rescue operations}.
\newblock {\it \bibinfo{journal}{IEEE Journal on Miniaturization for Air and
  Space Systems}\/},  {\it \bibinfo{volume}{2}\/}, \bibinfo{pages}{209--219}.
\bibitem[{Dosovitskiy et~al.(2020)Dosovitskiy, Beyer, Kolesnikov, Weissenborn,
  Zhai, Unterthiner, Dehghani, Minderer, Heigold, Gelly
  et~al.}]{dosovitskiy2020image}
\bibinfo{author}{Dosovitskiy, A.}, \bibinfo{author}{Beyer, L.},
  \bibinfo{author}{Kolesnikov, A.}, \bibinfo{author}{Weissenborn, D.},
  \bibinfo{author}{Zhai, X.}, \bibinfo{author}{Unterthiner, T.},
  \bibinfo{author}{Dehghani, M.}, \bibinfo{author}{Minderer, M.},
  \bibinfo{author}{Heigold, G.}, \bibinfo{author}{Gelly, S.} et~al.
  (\bibinfo{year}{2020}).
\newblock \bibinfo{title}{An image is worth 16x16 words: Transformers for image
  recognition at scale}.
\newblock {\it \bibinfo{journal}{arXiv preprint arXiv:2010.11929}\/}, .
\bibitem[{Dujon et~al.(2021)Dujon, Ierodiaconou, Geeson, Arnould, Allan,
  Katselidis \& Schofield}]{dujon2021machine}
\bibinfo{author}{Dujon, A.~M.}, \bibinfo{author}{Ierodiaconou, D.},
  \bibinfo{author}{Geeson, J.~J.}, \bibinfo{author}{Arnould, J.~P.},
  \bibinfo{author}{Allan, B.~M.}, \bibinfo{author}{Katselidis, K.~A.}, \&
  \bibinfo{author}{Schofield, G.} (\bibinfo{year}{2021}).
\newblock \bibinfo{title}{Machine learning to detect marine animals in {UAV}
  imagery: Effect of morphology, spacing, behaviour and habitat}.
\newblock {\it \bibinfo{journal}{Remote Sensing in Ecology and
  Conservation}\/},  {\it \bibinfo{volume}{7}\/}, \bibinfo{pages}{341--354}.
\bibitem[{EMSA(2018)}]{emsa2018annual}
\bibinfo{author}{EMSA, P.} (\bibinfo{year}{2018}).
\newblock \bibinfo{title}{Annual overview of marine casualties and incidents}.
\bibitem[{Fefilatyev et~al.(2012)Fefilatyev, Goldgof, Shreve \&
  Lembke}]{fefilatyev2012detection}
\bibinfo{author}{Fefilatyev, S.}, \bibinfo{author}{Goldgof, D.},
  \bibinfo{author}{Shreve, M.}, \& \bibinfo{author}{Lembke, C.}
  (\bibinfo{year}{2012}).
\newblock \bibinfo{title}{Detection and tracking of ships in open sea with
  rapidly moving buoy-mounted camera system}.
\newblock {\it \bibinfo{journal}{Ocean Engineering}\/},  {\it
  \bibinfo{volume}{54}\/}, \bibinfo{pages}{1--12}.
\bibitem[{Feng et~al.(2021)Feng, Zhong, Gao, Scott \& Huang}]{feng2021tood}
\bibinfo{author}{Feng, C.}, \bibinfo{author}{Zhong, Y.}, \bibinfo{author}{Gao,
  Y.}, \bibinfo{author}{Scott, M.~R.}, \& \bibinfo{author}{Huang, W.}
  (\bibinfo{year}{2021}).
\newblock \bibinfo{title}{{TOOD}: Task-aligned one-stage object detection}.
\newblock In {\it \bibinfo{booktitle}{2021 IEEE/CVF International Conference on
  Computer Vision (ICCV)}\/} (pp. \bibinfo{pages}{3490--3499}).
\newblock \bibinfo{organization}{IEEE Computer Society}.
\bibitem[{Feraru et~al.(2020)Feraru, Andersen \& Boukas}]{feraru2020towards}
\bibinfo{author}{Feraru, V.~A.}, \bibinfo{author}{Andersen, R.~E.}, \&
  \bibinfo{author}{Boukas, E.} (\bibinfo{year}{2020}).
\newblock \bibinfo{title}{Towards an autonomous {UAV}-based system to assist
  search and rescue operations in man overboard incidents}.
\newblock In {\it \bibinfo{booktitle}{2020 IEEE International Symposium on
  Safety, Security, and Rescue Robotics (SSRR)}\/} (pp.
  \bibinfo{pages}{57--64}).
\newblock \bibinfo{organization}{IEEE}.
\bibitem[{Fernandes et~al.(2023)Fernandes, Bispo, Bento \&
  Figueiredo}]{fernandes2023enhancing}
\bibinfo{author}{Fernandes, D.~S.}, \bibinfo{author}{Bispo, J.},
  \bibinfo{author}{Bento, L.~C.}, \& \bibinfo{author}{Figueiredo, M.}
  (\bibinfo{year}{2023}).
\newblock \bibinfo{title}{Enhancing object detection in maritime environments
  using metadata}.
\newblock In {\it \bibinfo{booktitle}{Iberoamerican Congress on Pattern
  Recognition}\/} (pp. \bibinfo{pages}{76--89}).
\newblock \bibinfo{organization}{Springer}.
\bibitem[{Gallego et~al.(2018)Gallego, Pertusa \& Gil}]{gallego2018automatic}
\bibinfo{author}{Gallego, A.-J.}, \bibinfo{author}{Pertusa, A.}, \&
  \bibinfo{author}{Gil, P.} (\bibinfo{year}{2018}).
\newblock \bibinfo{title}{Automatic ship classification from optical aerial
  images with convolutional neural networks}.
\newblock {\it \bibinfo{journal}{Remote Sensing}\/},  {\it
  \bibinfo{volume}{10}\/}, \bibinfo{pages}{511}.
\bibitem[{Gallego et~al.(2019)Gallego, Pertusa, Gil \&
  Fisher}]{gallego2019detection}
\bibinfo{author}{Gallego, A.-J.}, \bibinfo{author}{Pertusa, A.},
  \bibinfo{author}{Gil, P.}, \& \bibinfo{author}{Fisher, R.~B.}
  (\bibinfo{year}{2019}).
\newblock \bibinfo{title}{Detection of bodies in maritime rescue operations
  using unmanned aerial vehicles with multispectral cameras}.
\newblock {\it \bibinfo{journal}{Journal of Field Robotics}\/},  {\it
  \bibinfo{volume}{36}\/}, \bibinfo{pages}{782--796}.
\bibitem[{Gao et~al.(2022)Gao, Liu, Zhang, Zhou \& Qiu}]{gao2022multiscale}
\bibinfo{author}{Gao, S.}, \bibinfo{author}{Liu, C.}, \bibinfo{author}{Zhang,
  H.}, \bibinfo{author}{Zhou, Z.}, \& \bibinfo{author}{Qiu, J.}
  (\bibinfo{year}{2022}).
\newblock \bibinfo{title}{Multiscale attention-based detection of tiny targets
  in aerial beach images}.
\newblock {\it \bibinfo{journal}{Frontiers in Marine Science}\/},  {\it
  \bibinfo{volume}{9}\/}, \bibinfo{pages}{1073615}.
\bibitem[{Gasienica-Jozkowy et~al.(2021)Gasienica-Jozkowy, Knapik \&
  Cyganek}]{gasienica2021ensemble}
\bibinfo{author}{Gasienica-Jozkowy, J.}, \bibinfo{author}{Knapik, M.}, \&
  \bibinfo{author}{Cyganek, B.} (\bibinfo{year}{2021}).
\newblock \bibinfo{title}{An ensemble deep learning method with optimized
  weights for drone-based water rescue and surveillance}.
\newblock {\it \bibinfo{journal}{Integrated Computer-Aided Engineering}\/},
  {\it \bibinfo{volume}{28}\/}, \bibinfo{pages}{221--235}.
\bibitem[{Gaur et~al.(2023)Gaur, Liu, Lin, Karapetyan \&
  Aloimonos}]{gaur2023whale}
\bibinfo{author}{Gaur, A.}, \bibinfo{author}{Liu, C.}, \bibinfo{author}{Lin,
  X.}, \bibinfo{author}{Karapetyan, N.}, \& \bibinfo{author}{Aloimonos, Y.}
  (\bibinfo{year}{2023}).
\newblock \bibinfo{title}{Whale detection enhancement through synthetic
  satellite images}.
\newblock In {\it \bibinfo{booktitle}{OCEANS 2023-MTS/IEEE US Gulf Coast}\/}
  (pp. \bibinfo{pages}{1--7}).
\newblock \bibinfo{organization}{IEEE}.
\bibitem[{Geraldes et~al.(2019)Geraldes, Goncalves, Lai, Villerabel, Deng,
  Salta, Nakayama, Matsuo \& Prendinger}]{geraldes2019uav}
\bibinfo{author}{Geraldes, R.}, \bibinfo{author}{Goncalves, A.},
  \bibinfo{author}{Lai, T.}, \bibinfo{author}{Villerabel, M.},
  \bibinfo{author}{Deng, W.}, \bibinfo{author}{Salta, A.},
  \bibinfo{author}{Nakayama, K.}, \bibinfo{author}{Matsuo, Y.}, \&
  \bibinfo{author}{Prendinger, H.} (\bibinfo{year}{2019}).
\newblock \bibinfo{title}{{UAV}-based situational awareness system using deep
  learning}.
\newblock {\it \bibinfo{journal}{IEEE Access}\/},  {\it \bibinfo{volume}{7}\/},
  \bibinfo{pages}{122583--122594}.
\bibitem[{Ghahremani et~al.(2018)Ghahremani, Bondarev \&
  De~With}]{ghahremani2018cascaded}
\bibinfo{author}{Ghahremani, A.}, \bibinfo{author}{Bondarev, E.}, \&
  \bibinfo{author}{De~With, P.~H.} (\bibinfo{year}{2018}).
\newblock \bibinfo{title}{Cascaded {CNN} method for far object detection in
  outdoor surveillance}.
\newblock In {\it \bibinfo{booktitle}{2018 14th International Conference on
  Signal-Image Technology \& Internet-Based Systems (SITIS)}\/} (pp.
  \bibinfo{pages}{40--47}).
\newblock \bibinfo{organization}{IEEE}.
\bibitem[{Girshick(2015)}]{girshick2015fast}
\bibinfo{author}{Girshick, R.} (\bibinfo{year}{2015}).
\newblock \bibinfo{title}{Fast r-cnn}.
\newblock In {\it \bibinfo{booktitle}{Proceedings of the IEEE International
  Conference on Computer Vision}\/} (pp. \bibinfo{pages}{1440--1448}).
\bibitem[{Gon{\c{c}}alves \& Damas(2022)}]{gonccalves2022automatic}
\bibinfo{author}{Gon{\c{c}}alves, L.}, \& \bibinfo{author}{Damas, B.}
  (\bibinfo{year}{2022}).
\newblock \bibinfo{title}{Automatic detection of rescue targets in maritime
  search and rescue missions using uavs}.
\newblock In {\it \bibinfo{booktitle}{2022 International Conference on Unmanned
  Aircraft Systems (ICUAS)}\/} (pp. \bibinfo{pages}{1638--1643}).
\newblock \bibinfo{organization}{IEEE}.
\bibitem[{Gray et~al.(2019)Gray, Fleishman, Klein, McKown, Bezy, Lohmann \&
  Johnston}]{gray2019convolutional}
\bibinfo{author}{Gray, P.~C.}, \bibinfo{author}{Fleishman, A.~B.},
  \bibinfo{author}{Klein, D.~J.}, \bibinfo{author}{McKown, M.~W.},
  \bibinfo{author}{Bezy, V.~S.}, \bibinfo{author}{Lohmann, K.~J.}, \&
  \bibinfo{author}{Johnston, D.~W.} (\bibinfo{year}{2019}).
\newblock \bibinfo{title}{A convolutional neural network for detecting sea
  turtles in drone imagery}.
\newblock {\it \bibinfo{journal}{Methods in Ecology and Evolution}\/},  {\it
  \bibinfo{volume}{10}\/}, \bibinfo{pages}{345--355}.
\bibitem[{Guo et~al.(2023)Guo, Chen, Yu, Han, Ye \& Gao}]{guo2023save}
\bibinfo{author}{Guo, G.}, \bibinfo{author}{Chen, P.}, \bibinfo{author}{Yu,
  X.}, \bibinfo{author}{Han, Z.}, \bibinfo{author}{Ye, Q.}, \&
  \bibinfo{author}{Gao, S.} (\bibinfo{year}{2023}).
\newblock \bibinfo{title}{Save the tiny, save the all: Hierarchical activation
  network for tiny object detection}.
\newblock {\it \bibinfo{journal}{IEEE Transactions on Circuits and Systems for
  Video Technology}\/}, .
\bibitem[{Han et~al.(2020)Han, Wang, Tian, Guo, Xu \& Xu}]{han2020ghostnet}
\bibinfo{author}{Han, K.}, \bibinfo{author}{Wang, Y.}, \bibinfo{author}{Tian,
  Q.}, \bibinfo{author}{Guo, J.}, \bibinfo{author}{Xu, C.}, \&
  \bibinfo{author}{Xu, C.} (\bibinfo{year}{2020}).
\newblock \bibinfo{title}{Ghostnet: More features from cheap operations}.
\newblock In {\it \bibinfo{booktitle}{Proceedings of the IEEE/CVF conference on
  computer vision and pattern recognition}\/} (pp.
  \bibinfo{pages}{1580--1589}).
\bibitem[{Han et~al.(2024)Han, Fu \& Han}]{han2024detection}
\bibinfo{author}{Han, X.}, \bibinfo{author}{Fu, S.}, \& \bibinfo{author}{Han,
  J.} (\bibinfo{year}{2024}).
\newblock \bibinfo{title}{Detection and tracking of low-frame-rate water
  surface dynamic multi-target based on the yolov7-deepsort fusion algorithm}.
\newblock {\it \bibinfo{journal}{Journal of Marine Science and Engineering}\/},
   {\it \bibinfo{volume}{12}\/}, \bibinfo{pages}{1528}.
\bibitem[{Hasan et~al.(2021)Hasan, Rahman, Sedigh, Khasanah, Asyhari, Tao \&
  Bakar}]{hasan2021search}
\bibinfo{author}{Hasan, M.~M.}, \bibinfo{author}{Rahman, M.~A.},
  \bibinfo{author}{Sedigh, A.}, \bibinfo{author}{Khasanah, A.~U.},
  \bibinfo{author}{Asyhari, A.~T.}, \bibinfo{author}{Tao, H.}, \&
  \bibinfo{author}{Bakar, S.~A.} (\bibinfo{year}{2021}).
\newblock \bibinfo{title}{Search and rescue operation in flooded areas: A
  survey on emerging sensor networking-enabled {IoT}-oriented technologies and
  applications}.
\newblock {\it \bibinfo{journal}{Cognitive Systems Research}\/},  {\it
  \bibinfo{volume}{67}\/}, \bibinfo{pages}{104--123}.
\bibitem[{He et~al.(2017)He, Gkioxari, Doll{\'a}r \& Girshick}]{he2017mask}
\bibinfo{author}{He, K.}, \bibinfo{author}{Gkioxari, G.},
  \bibinfo{author}{Doll{\'a}r, P.}, \& \bibinfo{author}{Girshick, R.}
  (\bibinfo{year}{2017}).
\newblock \bibinfo{title}{Mask {R-CNN}}.
\newblock In {\it \bibinfo{booktitle}{Proceedings of the IEEE international
  conference on computer vision}\/} (pp. \bibinfo{pages}{2961--2969}).
\bibitem[{He et~al.(2016)He, Zhang, Ren \& Sun}]{he2016deep}
\bibinfo{author}{He, K.}, \bibinfo{author}{Zhang, X.}, \bibinfo{author}{Ren,
  S.}, \& \bibinfo{author}{Sun, J.} (\bibinfo{year}{2016}).
\newblock \bibinfo{title}{Deep residual learning for image recognition}.
\newblock In {\it \bibinfo{booktitle}{Proceedings of the IEEE conference on
  computer vision and pattern recognition}\/} (pp. \bibinfo{pages}{770--778}).
\bibitem[{Henriques et~al.(2014)Henriques, Caseiro, Martins \&
  Batista}]{henriques2014high}
\bibinfo{author}{Henriques, J.~F.}, \bibinfo{author}{Caseiro, R.},
  \bibinfo{author}{Martins, P.}, \& \bibinfo{author}{Batista, J.}
  (\bibinfo{year}{2014}).
\newblock \bibinfo{title}{High-speed tracking with kernelized correlation
  filters}.
\newblock {\it \bibinfo{journal}{IEEE transactions on pattern analysis and
  machine intelligence}\/},  {\it \bibinfo{volume}{37}\/},
  \bibinfo{pages}{583--596}.
\bibitem[{Hoai \& Van~Phuong(2017)}]{hoai2017anomaly}
\bibinfo{author}{Hoai, D.~K.}, \& \bibinfo{author}{Van~Phuong, N.}
  (\bibinfo{year}{2017}).
\newblock \bibinfo{title}{Anomaly color detection on {UAV} images for search
  and rescue works}.
\newblock In {\it \bibinfo{booktitle}{2017 9th International Conference on
  Knowledge and Systems Engineering (KSE)}\/} (pp. \bibinfo{pages}{287--291}).
\newblock \bibinfo{organization}{IEEE}.
\bibitem[{Hong et~al.(2021)Hong, Li, Yang, Zhu, Zhao \& Lu}]{hong2021sspnet}
\bibinfo{author}{Hong, M.}, \bibinfo{author}{Li, S.}, \bibinfo{author}{Yang,
  Y.}, \bibinfo{author}{Zhu, F.}, \bibinfo{author}{Zhao, Q.}, \&
  \bibinfo{author}{Lu, L.} (\bibinfo{year}{2021}).
\newblock \bibinfo{title}{{SSPNet}: Scale selection pyramid network for tiny
  person detection from uav images}.
\newblock {\it \bibinfo{journal}{IEEE Geoscience and Remote Sensing
  Letters}\/},  {\it \bibinfo{volume}{19}\/}, \bibinfo{pages}{1--5}.
\bibitem[{Horyna et~al.(2023)Horyna, Baca, Walter, Albani, Hert, Ferrante \&
  Saska}]{horyna2023decentralized}
\bibinfo{author}{Horyna, J.}, \bibinfo{author}{Baca, T.},
  \bibinfo{author}{Walter, V.}, \bibinfo{author}{Albani, D.},
  \bibinfo{author}{Hert, D.}, \bibinfo{author}{Ferrante, E.}, \&
  \bibinfo{author}{Saska, M.} (\bibinfo{year}{2023}).
\newblock \bibinfo{title}{Decentralized swarms of unmanned aerial vehicles for
  search and rescue operations without explicit communication}.
\newblock {\it \bibinfo{journal}{Autonomous Robots}\/},  {\it
  \bibinfo{volume}{47}\/}, \bibinfo{pages}{77--93}.
\bibitem[{Hu \& Dong(2023)}]{hu2023fspn}
\bibinfo{author}{Hu, J.}, \& \bibinfo{author}{Dong, F.} (\bibinfo{year}{2023}).
\newblock \bibinfo{title}{{FSPN}: Feature selection pyramid network for tiny
  person detection}.
\newblock In {\it \bibinfo{booktitle}{2023 International Conference on Image
  Processing, Computer Vision and Machine Learning (ICICML)}\/} (pp.
  \bibinfo{pages}{595--599}).
\newblock \bibinfo{organization}{IEEE}.
\bibitem[{Hu(1962)}]{HuMoments1962}
\bibinfo{author}{Hu, M.-K.} (\bibinfo{year}{1962}).
\newblock \bibinfo{title}{Visual pattern recognition by moment invariants}.
\newblock {\it \bibinfo{journal}{{IRE} Transactions on Information Theory}\/},
  {\it \bibinfo{volume}{8}\/}, \bibinfo{pages}{179--187}.
\bibitem[{Huang \& Chong(2023)}]{huang2023reidtracker}
\bibinfo{author}{Huang, K.}, \& \bibinfo{author}{Chong, W.}
  (\bibinfo{year}{2023}).
\newblock \bibinfo{title}{{ReIDTracker} {Sea}: the technical report of
  {BoaTrack} and {SeaDronesSee}-{MOT} challenge at {MaCVi} of {WACV24}}.
\newblock {\it \bibinfo{journal}{arXiv preprint arXiv:2311.07616}\/}, .
\bibitem[{Huang et~al.(2020)Huang, Chen, Chen, Negenborn \&
  Van~Gelder}]{huang2020ship}
\bibinfo{author}{Huang, Y.}, \bibinfo{author}{Chen, L.}, \bibinfo{author}{Chen,
  P.}, \bibinfo{author}{Negenborn, R.~R.}, \& \bibinfo{author}{Van~Gelder, P.}
  (\bibinfo{year}{2020}).
\newblock \bibinfo{title}{Ship collision avoidance methods: State-of-the-art}.
\newblock {\it \bibinfo{journal}{Safety science}\/},  {\it
  \bibinfo{volume}{121}\/}, \bibinfo{pages}{451--473}.
\bibitem[{Kaiyu \& Chaojian(2009)}]{kaiyu2009vision}
\bibinfo{author}{Kaiyu, X.}, \& \bibinfo{author}{Chaojian, S.}
  (\bibinfo{year}{2009}).
\newblock \bibinfo{title}{Vision enhancement system for {SAR} based on infrared
  video}.
\newblock In {\it \bibinfo{booktitle}{2009 International Conference on Image
  Analysis and Signal Processing}\/} (pp. \bibinfo{pages}{196--199}).
\newblock \bibinfo{organization}{IEEE}.
\bibitem[{Kang et~al.(2022)Kang, Tariq, Oh \& Woo}]{kang2022survey}
\bibinfo{author}{Kang, J.}, \bibinfo{author}{Tariq, S.}, \bibinfo{author}{Oh,
  H.}, \& \bibinfo{author}{Woo, S.~S.} (\bibinfo{year}{2022}).
\newblock \bibinfo{title}{A survey of deep learning-based object detection
  methods and datasets for overhead imagery}.
\newblock {\it \bibinfo{journal}{IEEE Access}\/},  {\it
  \bibinfo{volume}{10}\/}, \bibinfo{pages}{20118--20134}.
\bibitem[{Khan~Mohammadi et~al.(2024)Khan~Mohammadi, Schneidereit,
  Mansouri~Yarahmadi \& Breu{\ss}}]{khan2024investigating}
\bibinfo{author}{Khan~Mohammadi, M.}, \bibinfo{author}{Schneidereit, T.},
  \bibinfo{author}{Mansouri~Yarahmadi, A.}, \& \bibinfo{author}{Breu{\ss}, M.}
  (\bibinfo{year}{2024}).
\newblock \bibinfo{title}{Investigating training datasets of real and synthetic
  images for outdoor swimmer localisation with {YOLO}}.
\newblock {\it \bibinfo{journal}{AI}\/},  {\it \bibinfo{volume}{5}\/},
  \bibinfo{pages}{576--593}.
\bibitem[{Kiefer et~al.(2023{\natexlab{a}})Kiefer, Kristan, Per{\v{s}},
  {\v{Z}}ust, Poiesi, Andrade, Bernardino, Dawkins, Raitoharju, Quan
  et~al.}]{kiefer20231st}
\bibinfo{author}{Kiefer, B.}, \bibinfo{author}{Kristan, M.},
  \bibinfo{author}{Per{\v{s}}, J.}, \bibinfo{author}{{\v{Z}}ust, L.},
  \bibinfo{author}{Poiesi, F.}, \bibinfo{author}{Andrade, F.},
  \bibinfo{author}{Bernardino, A.}, \bibinfo{author}{Dawkins, M.},
  \bibinfo{author}{Raitoharju, J.}, \bibinfo{author}{Quan, Y.} et~al.
  (\bibinfo{year}{2023}{\natexlab{a}}).
\newblock \bibinfo{title}{1st workshop on maritime computer vision (macvi)
  2023: Challenge results}.
\newblock In {\it \bibinfo{booktitle}{Proceedings of the IEEE/CVF Winter
  Conference on Applications of Computer Vision}\/} (pp.
  \bibinfo{pages}{265--302}).
\bibitem[{Kiefer et~al.(2022)Kiefer, Ott \& Zell}]{kiefer2022leveraging}
\bibinfo{author}{Kiefer, B.}, \bibinfo{author}{Ott, D.}, \&
  \bibinfo{author}{Zell, A.} (\bibinfo{year}{2022}).
\newblock \bibinfo{title}{Leveraging synthetic data in object detection on
  unmanned aerial vehicles}.
\newblock In {\it \bibinfo{booktitle}{2022 26th International Conference on
  Pattern Recognition (ICPR)}\/} (pp. \bibinfo{pages}{3564--3571}).
\newblock \bibinfo{organization}{IEEE}.
\bibitem[{Kiefer et~al.(2023{\natexlab{b}})Kiefer, Quan \&
  Zell}]{kiefer2023memory}
\bibinfo{author}{Kiefer, B.}, \bibinfo{author}{Quan, Y.}, \&
  \bibinfo{author}{Zell, A.} (\bibinfo{year}{2023}{\natexlab{b}}).
\newblock \bibinfo{title}{Memory maps for video object detection and tracking
  on {UAVs}}.
\newblock {\it \bibinfo{journal}{arXiv preprint arXiv:2303.03508}\/}, .
\bibitem[{Kiefer \& Zell(2023)}]{kiefer2023fast}
\bibinfo{author}{Kiefer, B.}, \& \bibinfo{author}{Zell, A.}
  (\bibinfo{year}{2023}).
\newblock \bibinfo{title}{Fast region of interest proposals on maritime
  {UAVs}}.
\newblock In {\it \bibinfo{booktitle}{2023 IEEE International Conference on
  Robotics and Automation (ICRA)}\/} (pp. \bibinfo{pages}{3317--3324}).
\newblock \bibinfo{organization}{IEEE}.
\bibitem[{Kim et~al.(2022)Kim, Jeong \& Choi}]{kim2022effects}
\bibinfo{author}{Kim, C.}, \bibinfo{author}{Jeong, J.}, \&
  \bibinfo{author}{Choi, J.} (\bibinfo{year}{2022}).
\newblock \bibinfo{title}{Effects of class imbalance and data scarcity on the
  performance of binary classification machine learning models developed based
  on toxcast/tox21 assay data}.
\newblock {\it \bibinfo{journal}{Chemical Research in Toxicology}\/}, .
\bibitem[{Kim \& Lee(2014)}]{kim2014small}
\bibinfo{author}{Kim, S.}, \& \bibinfo{author}{Lee, J.} (\bibinfo{year}{2014}).
\newblock \bibinfo{title}{Small infrared target detection by region-adaptive
  clutter rejection for sea-based infrared search and track}.
\newblock {\it \bibinfo{journal}{Sensors}\/},  {\it \bibinfo{volume}{14}\/},
  \bibinfo{pages}{13210--13242}.
\bibitem[{Koshimura et~al.(2020)Koshimura, Moya, Mas \&
  Bai}]{koshimura2020tsunami}
\bibinfo{author}{Koshimura, S.}, \bibinfo{author}{Moya, L.},
  \bibinfo{author}{Mas, E.}, \& \bibinfo{author}{Bai, Y.}
  (\bibinfo{year}{2020}).
\newblock \bibinfo{title}{Tsunami damage detection with remote sensing: A
  review}.
\newblock {\it \bibinfo{journal}{Geosciences}\/},  {\it
  \bibinfo{volume}{10}\/}, \bibinfo{pages}{177}.
\bibitem[{Kyrkou et~al.(2018)Kyrkou, Plastiras, Theocharides, Venieris \&
  Bouganis}]{kyrkou2018dronet}
\bibinfo{author}{Kyrkou, C.}, \bibinfo{author}{Plastiras, G.},
  \bibinfo{author}{Theocharides, T.}, \bibinfo{author}{Venieris, S.~I.}, \&
  \bibinfo{author}{Bouganis, C.-S.} (\bibinfo{year}{2018}).
\newblock \bibinfo{title}{{DroNet}: Efficient convolutional neural network
  detector for real-time {UAV} applications}.
\newblock In {\it \bibinfo{booktitle}{2018 Design, Automation \& Test in Europe
  Conference \& Exhibition (DATE)}\/} (pp. \bibinfo{pages}{967--972}).
\newblock \bibinfo{organization}{IEEE}.
\bibitem[{Leira et~al.(2015)Leira, Johansen \& Fossen}]{leira2015automatic}
\bibinfo{author}{Leira, F.~S.}, \bibinfo{author}{Johansen, T.~A.}, \&
  \bibinfo{author}{Fossen, T.~I.} (\bibinfo{year}{2015}).
\newblock \bibinfo{title}{Automatic detection, classification and tracking of
  objects in the ocean surface from {UAVs} using a thermal camera}.
\newblock In {\it \bibinfo{booktitle}{2015 IEEE Aerospace Conference}\/} (pp.
  \bibinfo{pages}{1--10}).
\newblock \bibinfo{organization}{IEEE}.
\bibitem[{Li et~al.(2023)Li, Zhang, Jiang \& Zhang}]{li2023survey}
\bibinfo{author}{Li, J.}, \bibinfo{author}{Zhang, G.}, \bibinfo{author}{Jiang,
  C.}, \& \bibinfo{author}{Zhang, W.} (\bibinfo{year}{2023}).
\newblock \bibinfo{title}{A survey of maritime unmanned search system: theory,
  applications and future directions}.
\newblock {\it \bibinfo{journal}{Ocean Engineering}\/},  {\it
  \bibinfo{volume}{285}\/}, \bibinfo{pages}{115359}.
\bibitem[{Li et~al.(2022)Li, Gao, Deng, Tu, Zha \& Huang}]{li2022long}
\bibinfo{author}{Li, L.}, \bibinfo{author}{Gao, X.}, \bibinfo{author}{Deng,
  J.}, \bibinfo{author}{Tu, Y.}, \bibinfo{author}{Zha, Z.-J.}, \&
  \bibinfo{author}{Huang, Q.} (\bibinfo{year}{2022}).
\newblock \bibinfo{title}{Long short-term relation transformer with global
  gating for video captioning}.
\newblock {\it \bibinfo{journal}{IEEE Transactions on Image Processing}\/},
  {\it \bibinfo{volume}{31}\/}, \bibinfo{pages}{2726--2738}.
\bibitem[{Lin et~al.(2017)Lin, Goyal, Girshick, He \&
  Doll{\'a}r}]{lin2017focal}
\bibinfo{author}{Lin, T.-Y.}, \bibinfo{author}{Goyal, P.},
  \bibinfo{author}{Girshick, R.}, \bibinfo{author}{He, K.}, \&
  \bibinfo{author}{Doll{\'a}r, P.} (\bibinfo{year}{2017}).
\newblock \bibinfo{title}{Focal loss for dense object detection}.
\newblock In {\it \bibinfo{booktitle}{Proceedings of the IEEE International
  Conference on Computer Vision}\/} (pp. \bibinfo{pages}{2980--2988}).
\bibitem[{Lin et~al.(2023)Lin, Liu, Pattillo, Yu \&
  Aloimonous}]{lin2023seadronesim}
\bibinfo{author}{Lin, X.}, \bibinfo{author}{Liu, C.},
  \bibinfo{author}{Pattillo, A.}, \bibinfo{author}{Yu, M.}, \&
  \bibinfo{author}{Aloimonous, Y.} (\bibinfo{year}{2023}).
\newblock \bibinfo{title}{Seadronesim: Simulation of aerial images for
  detection of objects above water}.
\newblock In {\it \bibinfo{booktitle}{Proceedings of the IEEE/CVF Winter
  Conference on Applications of Computer Vision}\/} (pp.
  \bibinfo{pages}{216--223}).
\bibitem[{Liu et~al.(2022)Liu, Sun, Gu \& Deng}]{liu2022sf}
\bibinfo{author}{Liu, H.}, \bibinfo{author}{Sun, F.}, \bibinfo{author}{Gu, J.},
  \& \bibinfo{author}{Deng, L.} (\bibinfo{year}{2022}).
\newblock \bibinfo{title}{Sf-yolov5: A lightweight small object detection
  algorithm based on improved feature fusion mode}.
\newblock {\it \bibinfo{journal}{Sensors}\/},  {\it \bibinfo{volume}{22}\/},
  \bibinfo{pages}{5817}.
\bibitem[{Liu et~al.(2024)Liu, Qi, Xu \& Li}]{liu2024yolov5s}
\bibinfo{author}{Liu, K.}, \bibinfo{author}{Qi, Y.}, \bibinfo{author}{Xu, G.},
  \& \bibinfo{author}{Li, J.} (\bibinfo{year}{2024}).
\newblock \bibinfo{title}{{YOLOv5s} maritime distress target detection method
  based on swin transformer}.
\newblock {\it \bibinfo{journal}{IET Image Processing}\/}, .
\bibitem[{Liu et~al.(2018)Liu, Qi, Qin, Shi \& Jia}]{liu2018path}
\bibinfo{author}{Liu, S.}, \bibinfo{author}{Qi, L.}, \bibinfo{author}{Qin, H.},
  \bibinfo{author}{Shi, J.}, \& \bibinfo{author}{Jia, J.}
  (\bibinfo{year}{2018}).
\newblock \bibinfo{title}{Path aggregation network for instance segmentation}.
\newblock In {\it \bibinfo{booktitle}{Proceedings of the IEEE conference on
  computer vision and pattern recognition}\/} (pp.
  \bibinfo{pages}{8759--8768}).
\bibitem[{Liu et~al.(2016)Liu, Anguelov, Erhan, Szegedy, Reed, Fu \&
  Berg}]{liu2016ssd}
\bibinfo{author}{Liu, W.}, \bibinfo{author}{Anguelov, D.},
  \bibinfo{author}{Erhan, D.}, \bibinfo{author}{Szegedy, C.},
  \bibinfo{author}{Reed, S.}, \bibinfo{author}{Fu, C.-Y.}, \&
  \bibinfo{author}{Berg, A.~C.} (\bibinfo{year}{2016}).
\newblock \bibinfo{title}{{SSD}: Single shot multibox detector}.
\newblock In {\it \bibinfo{booktitle}{Computer Vision--ECCV 2016: 14th European
  Conference, Amsterdam, The Netherlands, October 11--14, 2016, Proceedings,
  Part I 14}\/} (pp. \bibinfo{pages}{21--37}).
\newblock \bibinfo{organization}{Springer}.
\bibitem[{Liu et~al.(2021)Liu, Lin, Cao, Hu, Wei, Zhang, Lin \&
  Guo}]{liu2021swin}
\bibinfo{author}{Liu, Z.}, \bibinfo{author}{Lin, Y.}, \bibinfo{author}{Cao,
  Y.}, \bibinfo{author}{Hu, H.}, \bibinfo{author}{Wei, Y.},
  \bibinfo{author}{Zhang, Z.}, \bibinfo{author}{Lin, S.}, \&
  \bibinfo{author}{Guo, B.} (\bibinfo{year}{2021}).
\newblock \bibinfo{title}{Swin transformer: Hierarchical vision transformer
  using shifted windows}.
\newblock In {\it \bibinfo{booktitle}{Proceedings of the IEEE/CVF International
  Conference on Computer Vision}\/} (pp. \bibinfo{pages}{10012--10022}).
\bibitem[{Lomonaco et~al.(2018)Lomonaco, Trotta, Ziosi, Avila \&
  D{\'\i}az-Rodr{\'\i}guez}]{lomonaco2018intelligent}
\bibinfo{author}{Lomonaco, V.}, \bibinfo{author}{Trotta, A.},
  \bibinfo{author}{Ziosi, M.}, \bibinfo{author}{Avila, J. D. D.~Y.}, \&
  \bibinfo{author}{D{\'\i}az-Rodr{\'\i}guez, N.} (\bibinfo{year}{2018}).
\newblock \bibinfo{title}{Intelligent drone swarm for search and rescue
  operations at sea}.
\newblock {\it \bibinfo{journal}{arXiv preprint arXiv:1811.05291}\/}, .
\bibitem[{Lu et~al.(2023)Lu, Niu, Lan, Liu, Wang, Yu \& Hu}]{lu2023high}
\bibinfo{author}{Lu, W.}, \bibinfo{author}{Niu, C.}, \bibinfo{author}{Lan, C.},
  \bibinfo{author}{Liu, W.}, \bibinfo{author}{Wang, S.}, \bibinfo{author}{Yu,
  J.}, \& \bibinfo{author}{Hu, T.} (\bibinfo{year}{2023}).
\newblock \bibinfo{title}{High-quality object detection method for {UAV} images
  based on improved {DINO} and masked image modeling}.
\newblock {\it \bibinfo{journal}{Remote Sensing}\/},  {\it
  \bibinfo{volume}{15}\/}, \bibinfo{pages}{4740}.
\bibitem[{Luiten et~al.(2021)Luiten, Osep, Dendorfer, Torr, Geiger,
  Leal-Taix{\'e} \& Leibe}]{luiten2021hota}
\bibinfo{author}{Luiten, J.}, \bibinfo{author}{Osep, A.},
  \bibinfo{author}{Dendorfer, P.}, \bibinfo{author}{Torr, P.},
  \bibinfo{author}{Geiger, A.}, \bibinfo{author}{Leal-Taix{\'e}, L.}, \&
  \bibinfo{author}{Leibe, B.} (\bibinfo{year}{2021}).
\newblock \bibinfo{title}{Hota: A higher order metric for evaluating
  multi-object tracking}.
\newblock {\it \bibinfo{journal}{International Journal of Computer Vision}\/},
  {\it \bibinfo{volume}{129}\/}, \bibinfo{pages}{548--578}.
\bibitem[{Lygouras et~al.(2018)Lygouras, Gasteratos, Tarchanidis \&
  Mitropoulos}]{lygouras2018rolfer}
\bibinfo{author}{Lygouras, E.}, \bibinfo{author}{Gasteratos, A.},
  \bibinfo{author}{Tarchanidis, K.}, \& \bibinfo{author}{Mitropoulos, A.}
  (\bibinfo{year}{2018}).
\newblock \bibinfo{title}{{ROLFER}: A fully autonomous aerial rescue support
  system}.
\newblock {\it \bibinfo{journal}{Microprocessors and Microsystems}\/},  {\it
  \bibinfo{volume}{61}\/}, \bibinfo{pages}{32--42}.
\bibitem[{Lygouras et~al.(2019)Lygouras, Santavas, Taitzoglou, Tarchanidis,
  Mitropoulos \& Gasteratos}]{lygouras2019unsupervised}
\bibinfo{author}{Lygouras, E.}, \bibinfo{author}{Santavas, N.},
  \bibinfo{author}{Taitzoglou, A.}, \bibinfo{author}{Tarchanidis, K.},
  \bibinfo{author}{Mitropoulos, A.}, \& \bibinfo{author}{Gasteratos, A.}
  (\bibinfo{year}{2019}).
\newblock \bibinfo{title}{Unsupervised human detection with an embedded vision
  system on a fully autonomous {UAV} for search and rescue operations}.
\newblock {\it \bibinfo{journal}{Sensors}\/},  {\it \bibinfo{volume}{19}\/},
  \bibinfo{pages}{3542}.
\bibitem[{Lyu et~al.(2022)Lyu, Shao, Cheng, Yin \& Gao}]{lyu2022sea}
\bibinfo{author}{Lyu, H.}, \bibinfo{author}{Shao, Z.}, \bibinfo{author}{Cheng,
  T.}, \bibinfo{author}{Yin, Y.}, \& \bibinfo{author}{Gao, X.}
  (\bibinfo{year}{2022}).
\newblock \bibinfo{title}{Sea-surface object detection based on electro-optical
  sensors: A review}.
\newblock {\it \bibinfo{journal}{IEEE Intelligent Transportation Systems
  Magazine}\/},  {\it \bibinfo{volume}{15}\/}, \bibinfo{pages}{190--216}.
\bibitem[{Lyu et~al.(2023)Lyu, Zhao, Huang \& Huang}]{lyu2023unmanned}
\bibinfo{author}{Lyu, M.}, \bibinfo{author}{Zhao, Y.}, \bibinfo{author}{Huang,
  C.}, \& \bibinfo{author}{Huang, H.} (\bibinfo{year}{2023}).
\newblock \bibinfo{title}{Unmanned aerial vehicles for search and rescue: A
  survey}.
\newblock {\it \bibinfo{journal}{Remote Sensing}\/},  {\it
  \bibinfo{volume}{15}\/}, \bibinfo{pages}{3266}.
\bibitem[{Ma et~al.(2019)Ma, Zhang, Sun \& Chen}]{ma2019maritime}
\bibinfo{author}{Ma, M.}, \bibinfo{author}{Zhang, H.}, \bibinfo{author}{Sun,
  X.}, \& \bibinfo{author}{Chen, J.} (\bibinfo{year}{2019}).
\newblock \bibinfo{title}{Maritime targets classification based on {CNN} using
  {Gaofen}-3 {SAR} images}.
\newblock {\it \bibinfo{journal}{The Journal of Engineering}\/},  {\it
  \bibinfo{volume}{2019}\/}, \bibinfo{pages}{7843--7846}.
\bibitem[{Maire et~al.(2014)Maire, Mejias \& Hodgson}]{maire2014convolutional}
\bibinfo{author}{Maire, F.}, \bibinfo{author}{Mejias, L.}, \&
  \bibinfo{author}{Hodgson, A.} (\bibinfo{year}{2014}).
\newblock \bibinfo{title}{A convolutional neural network for automatic analysis
  of aerial imagery}.
\newblock In {\it \bibinfo{booktitle}{2014 International Conference on Digital
  Image Computing: Techniques and Applications (DICTA)}\/} (pp.
  \bibinfo{pages}{1--8}).
\newblock \bibinfo{organization}{IEEE}.
\bibitem[{Martinez-Alpiste et~al.(2021)Martinez-Alpiste, Golcarenarenji, Wang
  \& Alcaraz-Calero}]{martinez2021search}
\bibinfo{author}{Martinez-Alpiste, I.}, \bibinfo{author}{Golcarenarenji, G.},
  \bibinfo{author}{Wang, Q.}, \& \bibinfo{author}{Alcaraz-Calero, J.~M.}
  (\bibinfo{year}{2021}).
\newblock \bibinfo{title}{Search and rescue operation using {UAVs}: A case
  study}.
\newblock {\it \bibinfo{journal}{Expert Systems with Applications}\/},  {\it
  \bibinfo{volume}{178}\/}, \bibinfo{pages}{114937}.
\bibitem[{Martinez-Esteso et~al.(2025)Martinez-Esteso, Castellanos, Rosello,
  Calvo-Zaragoza \& Gallego}]{martinez2025use}
\bibinfo{author}{Martinez-Esteso, J.~P.}, \bibinfo{author}{Castellanos, F.~J.},
  \bibinfo{author}{Rosello, A.}, \bibinfo{author}{Calvo-Zaragoza, J.}, \&
  \bibinfo{author}{Gallego, A.~J.} (\bibinfo{year}{2025}).
\newblock \bibinfo{title}{On the use of synthetic data for body detection in
  maritime search and rescue operations}.
\newblock {\it \bibinfo{journal}{Engineering Applications of Artificial
  Intelligence}\/},  {\it \bibinfo{volume}{139}\/}, \bibinfo{pages}{109586}.
\bibitem[{Mendon{\c{c}}a et~al.(2016)Mendon{\c{c}}a, Marques, Marques,
  Lourenco, Pinto, Santana, Coito, Lobo \& Barata}]{mendoncca2016cooperative}
\bibinfo{author}{Mendon{\c{c}}a, R.}, \bibinfo{author}{Marques, M.~M.},
  \bibinfo{author}{Marques, F.}, \bibinfo{author}{Lourenco, A.},
  \bibinfo{author}{Pinto, E.}, \bibinfo{author}{Santana, P.},
  \bibinfo{author}{Coito, F.}, \bibinfo{author}{Lobo, V.}, \&
  \bibinfo{author}{Barata, J.} (\bibinfo{year}{2016}).
\newblock \bibinfo{title}{A cooperative multi-robot team for the surveillance
  of shipwreck survivors at sea}.
\newblock In {\it \bibinfo{booktitle}{OCEANS 2016 MTS/IEEE Monterey}\/} (pp.
  \bibinfo{pages}{1--6}).
\newblock \bibinfo{organization}{IEEE}.
\bibitem[{Mirjalili et~al.(2014)Mirjalili, Mirjalili \&
  Lewis}]{mirjalili2014grey}
\bibinfo{author}{Mirjalili, S.}, \bibinfo{author}{Mirjalili, S.~M.}, \&
  \bibinfo{author}{Lewis, A.} (\bibinfo{year}{2014}).
\newblock \bibinfo{title}{Grey wolf optimizer}.
\newblock {\it \bibinfo{journal}{Advances in engineering software}\/},  {\it
  \bibinfo{volume}{69}\/}, \bibinfo{pages}{46--61}.
\bibitem[{Mishra et~al.(2020)Mishra, Garg, Narang \& Mishra}]{mishra2020drone}
\bibinfo{author}{Mishra, B.}, \bibinfo{author}{Garg, D.},
  \bibinfo{author}{Narang, P.}, \& \bibinfo{author}{Mishra, V.}
  (\bibinfo{year}{2020}).
\newblock \bibinfo{title}{Drone-surveillance for search and rescue in natural
  disaster}.
\newblock {\it \bibinfo{journal}{Computer Communications}\/},  {\it
  \bibinfo{volume}{156}\/}, \bibinfo{pages}{1--10}.
\bibitem[{Moosbauer et~al.(2019)Moosbauer, Konig, Jakel \&
  Teutsch}]{moosbauer2019benchmark}
\bibinfo{author}{Moosbauer, S.}, \bibinfo{author}{Konig, D.},
  \bibinfo{author}{Jakel, J.}, \& \bibinfo{author}{Teutsch, M.}
  (\bibinfo{year}{2019}).
\newblock \bibinfo{title}{A benchmark for deep learning based object detection
  in maritime environments}.
\newblock In {\it \bibinfo{booktitle}{Proceedings of the IEEE/CVF Conference on
  Computer Vision and Pattern Recognition Workshops}\/} (pp.
  \bibinfo{pages}{0--0}).
\bibitem[{Murphy et~al.(2019)Murphy, Sreenan \& Brown}]{murphy2019autonomous}
\bibinfo{author}{Murphy, S.~O.}, \bibinfo{author}{Sreenan, C.}, \&
  \bibinfo{author}{Brown, K.~N.} (\bibinfo{year}{2019}).
\newblock \bibinfo{title}{Autonomous unmanned aerial vehicle for search and
  rescue using software defined radio}.
\newblock In {\it \bibinfo{booktitle}{2019 IEEE 89th Vehicular Technology
  Conference (VTC2019-Spring)}\/} (pp. \bibinfo{pages}{1--6}).
\newblock \bibinfo{organization}{IEEE}.
\bibitem[{Nair et~al.(2019)Nair, Rodrigues, Dsouza, Bellary \&
  Gonsalves}]{nair2019designing}
\bibinfo{author}{Nair, S.}, \bibinfo{author}{Rodrigues, G.},
  \bibinfo{author}{Dsouza, C.}, \bibinfo{author}{Bellary, S.}, \&
  \bibinfo{author}{Gonsalves, V.} (\bibinfo{year}{2019}).
\newblock \bibinfo{title}{Designing of beach rescue drone using {GPS} and
  zigbee technologies}.
\newblock In {\it \bibinfo{booktitle}{2019 International Conference on
  Communication and Electronics Systems (ICCES)}\/} (pp.
  \bibinfo{pages}{1154--1158}).
\newblock \bibinfo{organization}{IEEE}.
\bibitem[{Nandi \& Prasad(2024)}]{nandi2024advances}
\bibinfo{author}{Nandi, G.}, \& \bibinfo{author}{Prasad, Y.}
  (\bibinfo{year}{2024}).
\newblock \bibinfo{title}{Advances in marine animal detection techniques: A
  comprehensive review and analysis}.
\newblock {\it \bibinfo{journal}{Critical Approaches to Data Engineering
  Systems and Analysis}\/},  (pp. \bibinfo{pages}{34--49}).
\bibitem[{Nie et~al.(2020)Nie, Duan, Ding, Hu \& Wong}]{nie2020attention}
\bibinfo{author}{Nie, X.}, \bibinfo{author}{Duan, M.}, \bibinfo{author}{Ding,
  H.}, \bibinfo{author}{Hu, B.}, \& \bibinfo{author}{Wong, E.~K.}
  (\bibinfo{year}{2020}).
\newblock \bibinfo{title}{Attention mask {R}-{CNN} for ship detection and
  segmentation from remote sensing images}.
\newblock {\it \bibinfo{journal}{Ieee Access}\/},  {\it \bibinfo{volume}{8}\/},
  \bibinfo{pages}{9325--9334}.
\bibitem[{Niedzielski et~al.(2021)Niedzielski, Jurecka, Mizi{\'n}ski, Pawul \&
  Motyl}]{niedzielski2021first}
\bibinfo{author}{Niedzielski, T.}, \bibinfo{author}{Jurecka, M.},
  \bibinfo{author}{Mizi{\'n}ski, B.}, \bibinfo{author}{Pawul, W.}, \&
  \bibinfo{author}{Motyl, T.} (\bibinfo{year}{2021}).
\newblock \bibinfo{title}{First successful rescue of a lost person using the
  human detection system: {A} case study from {Beskid} {Niski} ({SE}
  {Poland})}.
\newblock {\it \bibinfo{journal}{Remote Sensing}\/},  {\it
  \bibinfo{volume}{13}\/}, \bibinfo{pages}{4903}.
\bibitem[{Oda et~al.(2023)Oda, Shimizu, Nakamoto, Carfi \&
  Mastrogiovanni}]{oda2023falcon}
\bibinfo{author}{Oda, T.}, \bibinfo{author}{Shimizu, S.},
  \bibinfo{author}{Nakamoto, R.}, \bibinfo{author}{Carfi, A.}, \&
  \bibinfo{author}{Mastrogiovanni, F.} (\bibinfo{year}{2023}).
\newblock \bibinfo{title}{Falcon: wide angle fovea vision system for marine
  rescue drone}.
\newblock In {\it \bibinfo{booktitle}{IECON 2023-49th Annual Conference of the
  IEEE Industrial Electronics Society}\/} (pp. \bibinfo{pages}{1--6}).
\newblock \bibinfo{organization}{IEEE}.
\bibitem[{Papanicolopulu(2016)}]{papanicolopulu2016duty}
\bibinfo{author}{Papanicolopulu, I.} (\bibinfo{year}{2016}).
\newblock \bibinfo{title}{The duty to rescue at sea, in peacetime and in war: A
  general overview}.
\newblock {\it \bibinfo{journal}{International Review of the Red Cross}\/},
  {\it \bibinfo{volume}{98}\/}, \bibinfo{pages}{491--514}.
\bibitem[{Park et~al.(2023)Park, Park, Kim, Oh \& Lee}]{park2023aerial}
\bibinfo{author}{Park, J.-J.}, \bibinfo{author}{Park, K.-A.},
  \bibinfo{author}{Kim, T.-S.}, \bibinfo{author}{Oh, S.}, \&
  \bibinfo{author}{Lee, M.} (\bibinfo{year}{2023}).
\newblock \bibinfo{title}{Aerial hyperspectral remote sensing detection for
  maritime search and surveillance of floating small objects}.
\newblock {\it \bibinfo{journal}{Advances in Space Research}\/},  {\it
  \bibinfo{volume}{72}\/}, \bibinfo{pages}{2118--2136}.
\bibitem[{Pobar(2023)}]{pobar2023yolov7}
\bibinfo{author}{Pobar, M.} (\bibinfo{year}{2023}).
\newblock \bibinfo{title}{{YOLOv7} model for small object handling in maritime
  images}.
\newblock In {\it \bibinfo{booktitle}{Central European Conference on
  Information and Intelligent Systems}\/} (pp. \bibinfo{pages}{391--397}).
\newblock \bibinfo{organization}{Faculty of Organization and Informatics
  Varazdin}.
\bibitem[{Poudel et~al.(2023)Poudel, Lima \& Andrade}]{poudel2023novel}
\bibinfo{author}{Poudel, R.}, \bibinfo{author}{Lima, L.}, \&
  \bibinfo{author}{Andrade, F.} (\bibinfo{year}{2023}).
\newblock \bibinfo{title}{A novel framework to evaluate and train object
  detection models for real-time victims search and rescue at sea with
  autonomous unmanned aerial systems using high-fidelity dynamic marine
  simulation environment}.
\newblock In {\it \bibinfo{booktitle}{Proceedings of the IEEE/CVF Winter
  Conference on Applications of Computer Vision}\/} (pp.
  \bibinfo{pages}{239--247}).
\bibitem[{Qi et~al.(2022)Qi, Zhang, Wang, Mazur, Liu \& Malaviya}]{qi2022small}
\bibinfo{author}{Qi, G.}, \bibinfo{author}{Zhang, Y.}, \bibinfo{author}{Wang,
  K.}, \bibinfo{author}{Mazur, N.}, \bibinfo{author}{Liu, Y.}, \&
  \bibinfo{author}{Malaviya, D.} (\bibinfo{year}{2022}).
\newblock \bibinfo{title}{Small object detection method based on adaptive
  spatial parallel convolution and fast multi-scale fusion}.
\newblock {\it \bibinfo{journal}{Remote Sensing}\/},  {\it
  \bibinfo{volume}{14}\/}, \bibinfo{pages}{420}.
\bibitem[{Qi et~al.(2024)Qi, Lin, Deng, Chen \& Fang}]{qi2024minimizing}
\bibinfo{author}{Qi, S.}, \bibinfo{author}{Lin, B.}, \bibinfo{author}{Deng,
  Y.}, \bibinfo{author}{Chen, X.}, \& \bibinfo{author}{Fang, Y.}
  (\bibinfo{year}{2024}).
\newblock \bibinfo{title}{Minimizing maximum latency of task offloading for
  multi-uav-assisted maritime search and rescue}.
\newblock {\it \bibinfo{journal}{IEEE transactions on vehicular technology}\/},
  .
\bibitem[{Qiao et~al.(2021)Qiao, Chen \& Yuille}]{qiao2021detectors}
\bibinfo{author}{Qiao, S.}, \bibinfo{author}{Chen, L.-C.}, \&
  \bibinfo{author}{Yuille, A.} (\bibinfo{year}{2021}).
\newblock \bibinfo{title}{Detectors: Detecting objects with recursive feature
  pyramid and switchable atrous convolution}.
\newblock In {\it \bibinfo{booktitle}{Proceedings of the IEEE/CVF Conference on
  Computer Vision and Pattern Recognition}\/} (pp.
  \bibinfo{pages}{10213--10224}).
\bibitem[{Ran \& Ren(2010)}]{ran2010search}
\bibinfo{author}{Ran, X.}, \& \bibinfo{author}{Ren, L.} (\bibinfo{year}{2010}).
\newblock \bibinfo{title}{Search aid system based on machine vision and its
  visual attention model for rescue target detection}.
\newblock In {\it \bibinfo{booktitle}{2010 Second WRI Global Congress on
  Intelligent Systems}\/} (pp. \bibinfo{pages}{149--152}).
\newblock \bibinfo{organization}{IEEE} volume~\bibinfo{volume}{2}.
\bibitem[{Redmon et~al.(2016)Redmon, Divvala, Girshick \&
  Farhadi}]{redmon2016you}
\bibinfo{author}{Redmon, J.}, \bibinfo{author}{Divvala, S.},
  \bibinfo{author}{Girshick, R.}, \& \bibinfo{author}{Farhadi, A.}
  (\bibinfo{year}{2016}).
\newblock \bibinfo{title}{You only look once: Unified, real-time object
  detection}.
\newblock In {\it \bibinfo{booktitle}{Proceedings of the IEEE Conference on
  Computer Vision and Pattern Recognition}\/} (pp. \bibinfo{pages}{779--788}).
\bibitem[{Reed \& Yu(1990)}]{reed1990adaptive}
\bibinfo{author}{Reed, I.~S.}, \& \bibinfo{author}{Yu, X.}
  (\bibinfo{year}{1990}).
\newblock \bibinfo{title}{Adaptive multiple-band {CFAR} detection of an optical
  pattern with unknown spectral distribution}.
\newblock {\it \bibinfo{journal}{IEEE Transactions on Acoustics, Wpeech, and
  Signal Processing}\/},  {\it \bibinfo{volume}{38}\/},
  \bibinfo{pages}{1760--1770}.
\bibitem[{Ren et~al.(2012)Ren, Shi \& Ran}]{ren2012target}
\bibinfo{author}{Ren, L.}, \bibinfo{author}{Shi, C.}, \& \bibinfo{author}{Ran,
  X.} (\bibinfo{year}{2012}).
\newblock \bibinfo{title}{Target detection of maritime search and rescue:
  saliency accumulation method}.
\newblock In {\it \bibinfo{booktitle}{2012 9th International Conference on
  Fuzzy Systems and Knowledge Discovery}\/} (pp. \bibinfo{pages}{1972--1976}).
\newblock \bibinfo{organization}{IEEE}.
\bibitem[{Ren et~al.(2011)Ren, Shi \& Ran}]{ren2011target}
\bibinfo{author}{Ren, L.}, \bibinfo{author}{Shi, C.-J.}, \&
  \bibinfo{author}{Ran, X.} (\bibinfo{year}{2011}).
\newblock \bibinfo{title}{Target detection in maritime search and rescue using
  {SVD} and frequency domain characteristics}.
\newblock In {\it \bibinfo{booktitle}{2011 International Conference on Machine
  Learning and Cybernetics}\/} (pp. \bibinfo{pages}{556--560}).
\newblock \bibinfo{organization}{IEEE} volume~\bibinfo{volume}{2}.
\bibitem[{Ren et~al.(2015)Ren, He, Girshick \& Sun}]{ren2015faster}
\bibinfo{author}{Ren, S.}, \bibinfo{author}{He, K.}, \bibinfo{author}{Girshick,
  R.}, \& \bibinfo{author}{Sun, J.} (\bibinfo{year}{2015}).
\newblock \bibinfo{title}{Faster {R}-{CNN}: Towards real-time object detection
  with region proposal networks}.
\newblock {\it \bibinfo{journal}{Advances in Neural Information Processing
  Systems}\/},  {\it \bibinfo{volume}{28}\/}.
\bibitem[{Restas et~al.(2015)}]{restas2015drone}
\bibinfo{author}{Restas, A.} et~al. (\bibinfo{year}{2015}).
\newblock \bibinfo{title}{Drone applications for supporting disaster
  management}.
\newblock {\it \bibinfo{journal}{World Journal of Engineering and
  Technology}\/},  {\it \bibinfo{volume}{3}\/}, \bibinfo{pages}{316}.
\bibitem[{Ribeiro et~al.(2017)Ribeiro, Cruz, Matos \&
  Bernardino}]{ribeiro2017data}
\bibinfo{author}{Ribeiro, R.}, \bibinfo{author}{Cruz, G.},
  \bibinfo{author}{Matos, J.}, \& \bibinfo{author}{Bernardino, A.}
  (\bibinfo{year}{2017}).
\newblock \bibinfo{title}{A data set for airborne maritime surveillance
  environments}.
\newblock {\it \bibinfo{journal}{IEEE Transactions on Circuits and Systems for
  Video Technology}\/},  {\it \bibinfo{volume}{29}\/},
  \bibinfo{pages}{2720--2732}.
\bibitem[{Rizk et~al.(2022{\natexlab{a}})Rizk, Heller, Douguet, Baghdadi \&
  Diguet}]{rizk2022optimization}
\bibinfo{author}{Rizk, M.}, \bibinfo{author}{Heller, D.},
  \bibinfo{author}{Douguet, R.}, \bibinfo{author}{Baghdadi, A.}, \&
  \bibinfo{author}{Diguet, J.-P.} (\bibinfo{year}{2022}{\natexlab{a}}).
\newblock \bibinfo{title}{Optimization of deep-learning detection of humans in
  marine environment on edge devices}.
\newblock In {\it \bibinfo{booktitle}{2022 29th IEEE International Conference
  on Electronics, Circuits and Systems (ICECS)}\/} (pp. \bibinfo{pages}{1--4}).
\newblock \bibinfo{organization}{IEEE}.
\bibitem[{Rizk et~al.(2022{\natexlab{b}})Rizk, Slim, Baghdadi \&
  Diguet}]{rizk2022towards}
\bibinfo{author}{Rizk, M.}, \bibinfo{author}{Slim, F.},
  \bibinfo{author}{Baghdadi, A.}, \& \bibinfo{author}{Diguet, J.-P.}
  (\bibinfo{year}{2022}{\natexlab{b}}).
\newblock \bibinfo{title}{Towards real-time human detection in maritime
  environment using embedded deep learning}.
\newblock In {\it \bibinfo{booktitle}{International Conference on
  System-Integrated Intelligence}\/} (pp. \bibinfo{pages}{583--593}).
\newblock \bibinfo{organization}{Springer}.
\bibitem[{Rodin et~al.(2018)Rodin, de~Lima, de~Alcantara~Andrade, Haddad,
  Johansen \& Storvold}]{rodin2018object}
\bibinfo{author}{Rodin, C.~D.}, \bibinfo{author}{de~Lima, L.~N.},
  \bibinfo{author}{de~Alcantara~Andrade, F.~A.}, \bibinfo{author}{Haddad,
  D.~B.}, \bibinfo{author}{Johansen, T.~A.}, \& \bibinfo{author}{Storvold, R.}
  (\bibinfo{year}{2018}).
\newblock \bibinfo{title}{Object classification in thermal images using
  convolutional neural networks for search and rescue missions with unmanned
  aerial systems}.
\newblock In {\it \bibinfo{booktitle}{2018 International Joint Conference on
  Neural Networks (IJCNN)}\/} (pp. \bibinfo{pages}{1--8}).
\newblock \bibinfo{organization}{IEEE}.
\bibitem[{Ruiz-Ponce et~al.(2023)Ruiz-Ponce, Ortiz-Perez, Garcia-Rodriguez \&
  Kiefer}]{ruiz2023poseidon}
\bibinfo{author}{Ruiz-Ponce, P.}, \bibinfo{author}{Ortiz-Perez, D.},
  \bibinfo{author}{Garcia-Rodriguez, J.}, \& \bibinfo{author}{Kiefer, B.}
  (\bibinfo{year}{2023}).
\newblock \bibinfo{title}{Poseidon: A data augmentation tool for small object
  detection datasets in maritime environments}.
\newblock {\it \bibinfo{journal}{Sensors}\/},  {\it \bibinfo{volume}{23}\/},
  \bibinfo{pages}{3691}.
\bibitem[{Sambolek \& Ivasic-Kos(2020)}]{sambolek2020person}
\bibinfo{author}{Sambolek, S.}, \& \bibinfo{author}{Ivasic-Kos, M.}
  (\bibinfo{year}{2020}).
\newblock \bibinfo{title}{Person detection in drone imagery}.
\newblock In {\it \bibinfo{booktitle}{2020 5th International Conference on
  Smart and Sustainable Technologies (SpliTech)}\/} (pp.
  \bibinfo{pages}{1--6}).
\newblock \bibinfo{organization}{IEEE}.
\bibitem[{Sambolek \& Ivasic-Kos(2021)}]{sambolek2021automatic}
\bibinfo{author}{Sambolek, S.}, \& \bibinfo{author}{Ivasic-Kos, M.}
  (\bibinfo{year}{2021}).
\newblock \bibinfo{title}{Automatic person detection in search and rescue
  operations using deep {CNN} detectors}.
\newblock {\it \bibinfo{journal}{Ieee Access}\/},  {\it \bibinfo{volume}{9}\/},
  \bibinfo{pages}{37905--37922}.
\bibitem[{Sandler et~al.(2018)Sandler, Howard, Zhu, Zhmoginov \&
  Chen}]{sandler2018mobilenetv2}
\bibinfo{author}{Sandler, M.}, \bibinfo{author}{Howard, A.},
  \bibinfo{author}{Zhu, M.}, \bibinfo{author}{Zhmoginov, A.}, \&
  \bibinfo{author}{Chen, L.-C.} (\bibinfo{year}{2018}).
\newblock \bibinfo{title}{Mobilenetv2: Inverted residuals and linear
  bottlenecks}.
\newblock In {\it \bibinfo{booktitle}{Proceedings of the IEEE conference on
  computer vision and pattern recognition}\/} (pp.
  \bibinfo{pages}{4510--4520}).
\bibitem[{Seger(2019)}]{seger2019coagency}
\bibinfo{author}{Seger, J.} (\bibinfo{year}{2019}).
\newblock \bibinfo{title}{Coagency of humans and artificial intelligence in sea
  rescue environments: A closer look at where artificial intelligence can help
  humans maintain and improve situational awareness in search and rescue
  operations}.
\bibitem[{Shao et~al.(2024)Shao, Guo, Wang, Bao \& Wang}]{shao2024small}
\bibinfo{author}{Shao, X.-Y.}, \bibinfo{author}{Guo, Y.},
  \bibinfo{author}{Wang, Y.-W.}, \bibinfo{author}{Bao, Z.-W.}, \&
  \bibinfo{author}{Wang, J.-Y.} (\bibinfo{year}{2024}).
\newblock \bibinfo{title}{A small object detection algorithm based on feature
  interaction and guided learning}.
\newblock {\it \bibinfo{journal}{Journal of Visual Communication and Image
  Representation}\/},  {\it \bibinfo{volume}{98}\/}, \bibinfo{pages}{104011}.
\bibitem[{Shao et~al.(2018)Shao, Wu, Wang, Du \& Li}]{shao2018seaships}
\bibinfo{author}{Shao, Z.}, \bibinfo{author}{Wu, W.}, \bibinfo{author}{Wang,
  Z.}, \bibinfo{author}{Du, W.}, \& \bibinfo{author}{Li, C.}
  (\bibinfo{year}{2018}).
\newblock \bibinfo{title}{Seaships: A large-scale precisely annotated dataset
  for ship detection}.
\newblock {\it \bibinfo{journal}{IEEE Transactions on Multimedia}\/},  {\it
  \bibinfo{volume}{20}\/}, \bibinfo{pages}{2593--2604}.
\bibitem[{Sharafaldeen et~al.(2022)Sharafaldeen, Rizk, Heller, Baghdadi \&
  Diguet}]{sharafaldeen2022marine}
\bibinfo{author}{Sharafaldeen, J.}, \bibinfo{author}{Rizk, M.},
  \bibinfo{author}{Heller, D.}, \bibinfo{author}{Baghdadi, A.}, \&
  \bibinfo{author}{Diguet, J.-P.} (\bibinfo{year}{2022}).
\newblock \bibinfo{title}{Marine object detection based on top-view scenes
  using deep learning on edge devices}.
\newblock In {\it \bibinfo{booktitle}{2022 International Conference on Smart
  Systems and Power Management (IC2SPM)}\/} (pp. \bibinfo{pages}{35--40}).
\newblock \bibinfo{organization}{IEEE}.
\bibitem[{Sharma et~al.(2022)Sharma, Saqib, Scully-Power \&
  Blumenstein}]{sharma2022sharkspotter}
\bibinfo{author}{Sharma, N.}, \bibinfo{author}{Saqib, M.},
  \bibinfo{author}{Scully-Power, P.}, \& \bibinfo{author}{Blumenstein, M.}
  (\bibinfo{year}{2022}).
\newblock \bibinfo{title}{{SharkSpotter}: Shark detection with drones for human
  safety and environmental protection}.
\newblock {\it \bibinfo{journal}{Humanity Driven AI: Productivity, Well-being,
  Sustainability and Partnership}\/},  (pp. \bibinfo{pages}{223--237}).
\bibitem[{Shi et~al.(2008)Shi, Xu, Peng \& Ren}]{shi2008architecture}
\bibinfo{author}{Shi, C.}, \bibinfo{author}{Xu, K.}, \bibinfo{author}{Peng,
  J.}, \& \bibinfo{author}{Ren, L.} (\bibinfo{year}{2008}).
\newblock \bibinfo{title}{Architecture of vision enhancement system for
  maritime search and rescue}.
\newblock In {\it \bibinfo{booktitle}{2008 8th International Conference on ITS
  Telecommunications}\/} (pp. \bibinfo{pages}{12--17}).
\newblock \bibinfo{organization}{IEEE}.
\bibitem[{Shi et~al.(2024)Shi, Li, Liu, Zhou \& Zhou}]{shi2024mtp}
\bibinfo{author}{Shi, Y.}, \bibinfo{author}{Li, S.}, \bibinfo{author}{Liu, Z.},
  \bibinfo{author}{Zhou, Z.}, \& \bibinfo{author}{Zhou, X.}
  (\bibinfo{year}{2024}).
\newblock \bibinfo{title}{{MTP}-{YOLO}: You only look once based maritime tiny
  person detector for emergency rescue}.
\newblock {\it \bibinfo{journal}{Journal of Marine Science and Engineering}\/},
   {\it \bibinfo{volume}{12}\/}, \bibinfo{pages}{669}.
\bibitem[{Simonyan \& Zisserman(2014)}]{simonyan2014very}
\bibinfo{author}{Simonyan, K.}, \& \bibinfo{author}{Zisserman, A.}
  (\bibinfo{year}{2014}).
\newblock \bibinfo{title}{Very deep convolutional networks for large-scale
  image recognition}.
\newblock {\it \bibinfo{journal}{arXiv preprint arXiv:1409.1556}\/}, .
\bibitem[{Su et~al.(2023)Su, Chen, Song \& Li}]{su2023survey}
\bibinfo{author}{Su, L.}, \bibinfo{author}{Chen, Y.}, \bibinfo{author}{Song,
  H.}, \& \bibinfo{author}{Li, W.} (\bibinfo{year}{2023}).
\newblock \bibinfo{title}{A survey of maritime vision datasets}.
\newblock {\it \bibinfo{journal}{Multimedia Tools and Applications}\/},  {\it
  \bibinfo{volume}{82}\/}, \bibinfo{pages}{28873--28893}.
\bibitem[{Su et~al.(2024)Su, Wan, Zhang, Guo, Wu, Liu, Cong, Jia \&
  Wei}]{su2024integrated}
\bibinfo{author}{Su, Z.}, \bibinfo{author}{Wan, G.}, \bibinfo{author}{Zhang,
  W.}, \bibinfo{author}{Guo, N.}, \bibinfo{author}{Wu, Y.},
  \bibinfo{author}{Liu, J.}, \bibinfo{author}{Cong, D.}, \bibinfo{author}{Jia,
  Y.}, \& \bibinfo{author}{Wei, Z.} (\bibinfo{year}{2024}).
\newblock \bibinfo{title}{An integrated detection and multi-object tracking
  pipeline for satellite video analysis of maritime and aerial objects}.
\newblock {\it \bibinfo{journal}{Remote Sensing}\/},  {\it
  \bibinfo{volume}{16}\/}, \bibinfo{pages}{724}.
\bibitem[{Sumimoto et~al.(1994)Sumimoto, Kuramoto, Okada, Miyauchi, Imade,
  Yamamoto \& Kunishi}]{sumimoto1994machine}
\bibinfo{author}{Sumimoto, T.}, \bibinfo{author}{Kuramoto, K.},
  \bibinfo{author}{Okada, S.}, \bibinfo{author}{Miyauchi, H.},
  \bibinfo{author}{Imade, M.}, \bibinfo{author}{Yamamoto, H.}, \&
  \bibinfo{author}{Kunishi, T.} (\bibinfo{year}{1994}).
\newblock \bibinfo{title}{Machine vision for detection of the rescue target in
  the marine casualty}.
\newblock In {\it \bibinfo{booktitle}{Proceedings of IECON'94-20th Annual
  Conference of IEEE Industrial Electronics}\/} (pp.
  \bibinfo{pages}{723--726}).
\newblock \bibinfo{organization}{IEEE} volume~\bibinfo{volume}{2}.
\bibitem[{Szegedy et~al.(2016)Szegedy, Vanhoucke, Ioffe, Shlens \&
  Wojna}]{szegedy2016rethinking}
\bibinfo{author}{Szegedy, C.}, \bibinfo{author}{Vanhoucke, V.},
  \bibinfo{author}{Ioffe, S.}, \bibinfo{author}{Shlens, J.}, \&
  \bibinfo{author}{Wojna, Z.} (\bibinfo{year}{2016}).
\newblock \bibinfo{title}{Rethinking the inception architecture for computer
  vision}.
\newblock In {\it \bibinfo{booktitle}{Proceedings of the IEEE conference on
  computer vision and pattern recognition}\/} (pp.
  \bibinfo{pages}{2818--2826}).
\bibitem[{Tan et~al.(2020)Tan, Pang \& Le}]{tan2020efficientdet}
\bibinfo{author}{Tan, M.}, \bibinfo{author}{Pang, R.}, \& \bibinfo{author}{Le,
  Q.~V.} (\bibinfo{year}{2020}).
\newblock \bibinfo{title}{Efficientdet: Scalable and efficient object
  detection}.
\newblock In {\it \bibinfo{booktitle}{Proceedings of the IEEE/CVF Conference on
  Computer Vision and Pattern Recognition}\/} (pp.
  \bibinfo{pages}{10781--10790}).
\bibitem[{Tang et~al.(2023)Tang, Yang \& Tian}]{tang2023long}
\bibinfo{author}{Tang, F.}, \bibinfo{author}{Yang, F.}, \&
  \bibinfo{author}{Tian, X.} (\bibinfo{year}{2023}).
\newblock \bibinfo{title}{Long-distance person detection based on {YOLOv7}}.
\newblock {\it \bibinfo{journal}{Electronics}\/},  {\it
  \bibinfo{volume}{12}\/}, \bibinfo{pages}{1502}.
\bibitem[{Tran et~al.(2024)Tran, Shipard, Mulyono, Wiliem \&
  Fookes}]{tran2024safesea}
\bibinfo{author}{Tran, M.}, \bibinfo{author}{Shipard, J.},
  \bibinfo{author}{Mulyono, H.}, \bibinfo{author}{Wiliem, A.}, \&
  \bibinfo{author}{Fookes, C.} (\bibinfo{year}{2024}).
\newblock \bibinfo{title}{{SafeSea}: Synthetic data generation for adverse \&
  low probability maritime conditions}.
\newblock In {\it \bibinfo{booktitle}{Proceedings of the IEEE/CVF Winter
  Conference on Applications of Computer Vision}\/} (pp.
  \bibinfo{pages}{821--829}).
\bibitem[{Turi{\'c} et~al.(2010)Turi{\'c}, Dujmi{\'c} \&
  Papi{\'c}}]{turic2010two}
\bibinfo{author}{Turi{\'c}, H.}, \bibinfo{author}{Dujmi{\'c}, H.}, \&
  \bibinfo{author}{Papi{\'c}, V.} (\bibinfo{year}{2010}).
\newblock \bibinfo{title}{Two-stage segmentation of aerial images for search
  and rescue}.
\newblock {\it \bibinfo{journal}{Information Technology and Control}\/},  {\it
  \bibinfo{volume}{39}\/}.
\bibitem[{Tu{\'s}nio \& Wr{\'o}blewski(2021)}]{tusnio2021efficiency}
\bibinfo{author}{Tu{\'s}nio, N.}, \& \bibinfo{author}{Wr{\'o}blewski, W.}
  (\bibinfo{year}{2021}).
\newblock \bibinfo{title}{The efficiency of drones usage for safety and rescue
  operations in an open area: A case from {Poland}}.
\newblock {\it \bibinfo{journal}{Sustainability}\/},  {\it
  \bibinfo{volume}{14}\/}, \bibinfo{pages}{327}.
\bibitem[{Usilin et~al.(2020)Usilin, Arlazarov, Rokhlin, Rudyka, Matveev \&
  Zatsarinnyy}]{usilin2020training}
\bibinfo{author}{Usilin, S.~A.}, \bibinfo{author}{Arlazarov, V.~V.},
  \bibinfo{author}{Rokhlin, N.~S.}, \bibinfo{author}{Rudyka, S.~A.},
  \bibinfo{author}{Matveev, S.~A.}, \& \bibinfo{author}{Zatsarinnyy, A.~A.}
  (\bibinfo{year}{2020}).
\newblock \bibinfo{title}{Training {Viola}-{Jones} detectors for {3D} objects
  based on fully synthetic data for use in rescue missions with {UAV}}.
\newblock {\it \bibinfo{journal}{Bulletin of the South Ural State University,
  Series: Mathematical Modelling, Programming and Computer Software}\/},  {\it
  \bibinfo{volume}{13}\/}, \bibinfo{pages}{94--106}.
  \DOIprefix\doi{10.14529/mmp200408}.
\bibitem[{Varga et~al.(2022)Varga, Kiefer, Messmer \&
  Zell}]{varga2022seadronessee}
\bibinfo{author}{Varga, L.~A.}, \bibinfo{author}{Kiefer, B.},
  \bibinfo{author}{Messmer, M.}, \& \bibinfo{author}{Zell, A.}
  (\bibinfo{year}{2022}).
\newblock \bibinfo{title}{Seadronessee: A maritime benchmark for detecting
  humans in open water}.
\newblock In {\it \bibinfo{booktitle}{Proceedings of the IEEE/CVF Winter
  Conference on Applications of Computer Vision}\/} (pp.
  \bibinfo{pages}{2260--2270}).
\bibitem[{Verfuss et~al.(2019)Verfuss, Aniceto, Harris, Gillespie, Fielding,
  Jim{\'e}nez, Johnston, Sinclair, Sivertsen, Solb{\o}
  et~al.}]{verfuss2019review}
\bibinfo{author}{Verfuss, U.~K.}, \bibinfo{author}{Aniceto, A.~S.},
  \bibinfo{author}{Harris, D.~V.}, \bibinfo{author}{Gillespie, D.},
  \bibinfo{author}{Fielding, S.}, \bibinfo{author}{Jim{\'e}nez, G.},
  \bibinfo{author}{Johnston, P.}, \bibinfo{author}{Sinclair, R.~R.},
  \bibinfo{author}{Sivertsen, A.}, \bibinfo{author}{Solb{\o}, S.~A.} et~al.
  (\bibinfo{year}{2019}).
\newblock \bibinfo{title}{A review of unmanned vehicles for the detection and
  monitoring of marine fauna}.
\newblock {\it \bibinfo{journal}{Marine Pollution Bulletin}\/},  {\it
  \bibinfo{volume}{140}\/}, \bibinfo{pages}{17--29}.
\bibitem[{Viola \& Jones(2004)}]{viola2004robust}
\bibinfo{author}{Viola, P.}, \& \bibinfo{author}{Jones, M.~J.}
  (\bibinfo{year}{2004}).
\newblock \bibinfo{title}{Robust real-time face detection}.
\newblock {\it \bibinfo{journal}{International journal of computer vision}\/},
  {\it \bibinfo{volume}{57}\/}, \bibinfo{pages}{137--154}.
\bibitem[{Waharte \& Trigoni(2010)}]{waharte2010supporting}
\bibinfo{author}{Waharte, S.}, \& \bibinfo{author}{Trigoni, N.}
  (\bibinfo{year}{2010}).
\newblock \bibinfo{title}{Supporting search and rescue operations with {UAVs}}.
\newblock In {\it \bibinfo{booktitle}{2010 International Conference on Emerging
  Security Technologies}\/} (pp. \bibinfo{pages}{142--147}).
\newblock \bibinfo{organization}{IEEE}.
\bibitem[{Wang et~al.(2022{\natexlab{a}})Wang, Wang \& Er}]{wang2022review}
\bibinfo{author}{Wang, N.}, \bibinfo{author}{Wang, Y.}, \& \bibinfo{author}{Er,
  M.~J.} (\bibinfo{year}{2022}{\natexlab{a}}).
\newblock \bibinfo{title}{Review on deep learning techniques for marine object
  recognition: Architectures and algorithms}.
\newblock {\it \bibinfo{journal}{Control Engineering Practice}\/},  {\it
  \bibinfo{volume}{118}\/}, \bibinfo{pages}{104458}.
\bibitem[{Wang et~al.(2023{\natexlab{a}})Wang, Pan, He \& Gao}]{wang2023sea}
\bibinfo{author}{Wang, X.}, \bibinfo{author}{Pan, Z.}, \bibinfo{author}{He,
  N.}, \& \bibinfo{author}{Gao, T.} (\bibinfo{year}{2023}{\natexlab{a}}).
\newblock \bibinfo{title}{Sea-{YOLOv5s}: A {UAV} image-based model for
  detecting objects in {SeaDronesSee} dataset}.
\newblock {\it \bibinfo{journal}{Journal of Intelligent \& Fuzzy Systems}\/},
  (pp. \bibinfo{pages}{1--12}).
\bibitem[{Wang et~al.(2022{\natexlab{b}})Wang, Bashir, Khan, Ullah, Wang, Song,
  Guo \& Niu}]{wang2022remote}
\bibinfo{author}{Wang, Y.}, \bibinfo{author}{Bashir, S. M.~A.},
  \bibinfo{author}{Khan, M.}, \bibinfo{author}{Ullah, Q.},
  \bibinfo{author}{Wang, R.}, \bibinfo{author}{Song, Y.}, \bibinfo{author}{Guo,
  Z.}, \& \bibinfo{author}{Niu, Y.} (\bibinfo{year}{2022}{\natexlab{b}}).
\newblock \bibinfo{title}{Remote sensing image super-resolution and object
  detection: Benchmark and state of the art}.
\newblock {\it \bibinfo{journal}{Expert Systems with Applications}\/},  {\it
  \bibinfo{volume}{197}\/}, \bibinfo{pages}{116793}.
\bibitem[{Wang et~al.(2023{\natexlab{b}})Wang, Liu, Liu \&
  Sun}]{wang2023cooperative}
\bibinfo{author}{Wang, Y.}, \bibinfo{author}{Liu, W.}, \bibinfo{author}{Liu,
  J.}, \& \bibinfo{author}{Sun, C.} (\bibinfo{year}{2023}{\natexlab{b}}).
\newblock \bibinfo{title}{Cooperative {USV}--{UAV} marine search and rescue
  with visual navigation and reinforcement learning-based control}.
\newblock {\it \bibinfo{journal}{ISA transactions}\/},  {\it
  \bibinfo{volume}{137}\/}, \bibinfo{pages}{222--235}.
\bibitem[{Westall et~al.(2008)Westall, Ford, O'Shea \& Hrabar}]{Westall2008}
\bibinfo{author}{Westall, P.}, \bibinfo{author}{Ford, J.~J.},
  \bibinfo{author}{O'Shea, P.}, \& \bibinfo{author}{Hrabar, S.}
  (\bibinfo{year}{2008}).
\newblock \bibinfo{title}{Evaluation of maritime vision techniques for aerial
  search of humans in maritime environments}.
\newblock In {\it \bibinfo{booktitle}{2008 Digital Image Computing: Techniques
  and Applications}\/} (pp. \bibinfo{pages}{176--183}).
\bibitem[{Westall et~al.(2009)Westall, O'Shea, Ford \& Hrabar}]{Westall2009}
\bibinfo{author}{Westall, P.}, \bibinfo{author}{O'Shea, P.},
  \bibinfo{author}{Ford, J.~J.}, \& \bibinfo{author}{Hrabar, S.}
  (\bibinfo{year}{2009}).
\newblock \bibinfo{title}{Improved maritime target tracker using colour
  fusion}.
\newblock In {\it \bibinfo{booktitle}{2009 International Conference on High
  Performance Computing and Simulation}\/} (pp. \bibinfo{pages}{230--236}).
\bibitem[{Xiao et~al.(2020)Xiao, Tian, Yu, Zhang, Liu, Du \&
  Lan}]{xiao2020review}
\bibinfo{author}{Xiao, Y.}, \bibinfo{author}{Tian, Z.}, \bibinfo{author}{Yu,
  J.}, \bibinfo{author}{Zhang, Y.}, \bibinfo{author}{Liu, S.},
  \bibinfo{author}{Du, S.}, \& \bibinfo{author}{Lan, X.}
  (\bibinfo{year}{2020}).
\newblock \bibinfo{title}{A review of object detection based on deep learning}.
\newblock {\it \bibinfo{journal}{Multimedia Tools and Applications}\/},  {\it
  \bibinfo{volume}{79}\/}, \bibinfo{pages}{23729--23791}.
\bibitem[{Yamamoto et~al.(1999)Yamamoto, Yamada, Kiriya \&
  Matsukura}]{yamamoto1999optical}
\bibinfo{author}{Yamamoto, K.}, \bibinfo{author}{Yamada, K.},
  \bibinfo{author}{Kiriya, N.}, \& \bibinfo{author}{Matsukura, H.}
  (\bibinfo{year}{1999}).
\newblock \bibinfo{title}{Optical sensing and image processing to detect a life
  raft}.
\newblock In {\it \bibinfo{booktitle}{IEEE 1999 International Geoscience and
  Remote Sensing Symposium. IGARSS'99 (Cat. No. 99CH36293)}\/} (pp.
  \bibinfo{pages}{467--469}).
\newblock \bibinfo{organization}{IEEE} volume~\bibinfo{volume}{1}.
\bibitem[{Yang et~al.(2024)Yang, Huang, Jiang, Kuo, Mei, Huang \&
  Hwang}]{yang2024sea}
\bibinfo{author}{Yang, C.-Y.}, \bibinfo{author}{Huang, H.-W.},
  \bibinfo{author}{Jiang, Z.}, \bibinfo{author}{Kuo, H.-C.},
  \bibinfo{author}{Mei, J.}, \bibinfo{author}{Huang, C.-I.}, \&
  \bibinfo{author}{Hwang, J.-N.} (\bibinfo{year}{2024}).
\newblock \bibinfo{title}{Sea you later: Metadata-guided long-term
  re-identification for {uav}-based multi-object tracking}.
\newblock In {\it \bibinfo{booktitle}{Proceedings of the IEEE/CVF Winter
  Conference on Applications of Computer Vision}\/} (pp.
  \bibinfo{pages}{805--812}).
\bibitem[{Yang(2010)}]{yang2010new}
\bibinfo{author}{Yang, X.-S.} (\bibinfo{year}{2010}).
\newblock \bibinfo{title}{A new metaheuristic bat-inspired algorithm}.
\newblock In {\it \bibinfo{booktitle}{Nature inspired cooperative strategies
  for optimization (NICSO 2010)}\/} (pp. \bibinfo{pages}{65--74}).
\newblock \bibinfo{publisher}{Springer}.
\bibitem[{Yang et~al.(2023)Yang, Yin, Jing \& Shao}]{yang2023high}
\bibinfo{author}{Yang, Z.}, \bibinfo{author}{Yin, Y.}, \bibinfo{author}{Jing,
  Q.}, \& \bibinfo{author}{Shao, Z.} (\bibinfo{year}{2023}).
\newblock \bibinfo{title}{A high-precision detection model of small objects in
  maritime {UAV} perspective based on improved {YOLOv5}}.
\newblock {\it \bibinfo{journal}{Journal of Marine Science and Engineering}\/},
   {\it \bibinfo{volume}{11}\/}, \bibinfo{pages}{1680}.
\bibitem[{Yu et~al.(2022)Yu, Chen, Wu, Hassan, Li, Yan, Shi, Ye \&
  Han}]{yu2022object}
\bibinfo{author}{Yu, X.}, \bibinfo{author}{Chen, P.}, \bibinfo{author}{Wu, D.},
  \bibinfo{author}{Hassan, N.}, \bibinfo{author}{Li, G.}, \bibinfo{author}{Yan,
  J.}, \bibinfo{author}{Shi, H.}, \bibinfo{author}{Ye, Q.}, \&
  \bibinfo{author}{Han, Z.} (\bibinfo{year}{2022}).
\newblock \bibinfo{title}{Object localization under single coarse point
  supervision}.
\newblock In {\it \bibinfo{booktitle}{Proceedings of the IEEE/CVF Conference on
  Computer Vision and Pattern Recognition}\/} (pp.
  \bibinfo{pages}{4868--4877}).
\bibitem[{Yu et~al.(2020)Yu, Gong, Jiang, Ye \& Han}]{yu2020scale}
\bibinfo{author}{Yu, X.}, \bibinfo{author}{Gong, Y.}, \bibinfo{author}{Jiang,
  N.}, \bibinfo{author}{Ye, Q.}, \& \bibinfo{author}{Han, Z.}
  (\bibinfo{year}{2020}).
\newblock \bibinfo{title}{Scale match for tiny person detection}.
\newblock In {\it \bibinfo{booktitle}{Proceedings of the IEEE/CVF Winter
  Conference on Applications of Computer Vision}\/} (pp.
  \bibinfo{pages}{1257--1265}).
\bibitem[{Yun et~al.(2019)Yun, Nguyen, Nguyen, Kim, Eldin, Huyen, Lu \&
  Chow}]{yun2019small}
\bibinfo{author}{Yun, K.}, \bibinfo{author}{Nguyen, L.},
  \bibinfo{author}{Nguyen, T.}, \bibinfo{author}{Kim, D.},
  \bibinfo{author}{Eldin, S.}, \bibinfo{author}{Huyen, A.},
  \bibinfo{author}{Lu, T.}, \& \bibinfo{author}{Chow, E.}
  (\bibinfo{year}{2019}).
\newblock \bibinfo{title}{Small target detection for search and rescue
  operations using distributed deep learning and synthetic data generation}.
\newblock In {\it \bibinfo{booktitle}{Pattern Recognition and Tracking XXX}\/}
  (pp. \bibinfo{pages}{38--43}).
\newblock \bibinfo{organization}{SPIE} volume \bibinfo{volume}{10995}.
\bibitem[{Zeng et~al.(2023)Zeng, Zhang, He \& Zhang}]{zeng2023yolov7}
\bibinfo{author}{Zeng, Y.}, \bibinfo{author}{Zhang, T.}, \bibinfo{author}{He,
  W.}, \& \bibinfo{author}{Zhang, Z.} (\bibinfo{year}{2023}).
\newblock \bibinfo{title}{{Yolov7}-{UAV}: An unmanned aerial vehicle image
  object detection algorithm based on improved yolov7}.
\newblock {\it \bibinfo{journal}{Electronics}\/},  {\it
  \bibinfo{volume}{12}\/}, \bibinfo{pages}{3141}.
\bibitem[{Zhang et~al.(2022{\natexlab{a}})Zhang, Li, Liu, Zhang, Su, Zhu, Ni \&
  Shum}]{zhang2022dino}
\bibinfo{author}{Zhang, H.}, \bibinfo{author}{Li, F.}, \bibinfo{author}{Liu,
  S.}, \bibinfo{author}{Zhang, L.}, \bibinfo{author}{Su, H.},
  \bibinfo{author}{Zhu, J.}, \bibinfo{author}{Ni, L.~M.}, \&
  \bibinfo{author}{Shum, H.-Y.} (\bibinfo{year}{2022}{\natexlab{a}}).
\newblock \bibinfo{title}{Dino: Detr with improved denoising anchor boxes for
  end-to-end object detection}.
\newblock {\it \bibinfo{journal}{arXiv preprint arXiv:2203.03605}\/}, .
\bibitem[{Zhang et~al.(2021{\natexlab{a}})Zhang, Wang, Dayoub \&
  Sunderhauf}]{zhang2021varifocalnet}
\bibinfo{author}{Zhang, H.}, \bibinfo{author}{Wang, Y.},
  \bibinfo{author}{Dayoub, F.}, \& \bibinfo{author}{Sunderhauf, N.}
  (\bibinfo{year}{2021}{\natexlab{a}}).
\newblock \bibinfo{title}{{VarifocaNnet}: An iou-aware dense object detector}.
\newblock In {\it \bibinfo{booktitle}{Proceedings of the IEEE/CVF conference on
  computer vision and pattern recognition}\/} (pp.
  \bibinfo{pages}{8514--8523}).
\bibitem[{Zhang et~al.(2021{\natexlab{b}})Zhang, Zhang, Nguyen, Lee \&
  Chan}]{zhang2021chinese}
\bibinfo{author}{Zhang, H.}, \bibinfo{author}{Zhang, Q.},
  \bibinfo{author}{Nguyen, P.~A.}, \bibinfo{author}{Lee, V.~C.}, \&
  \bibinfo{author}{Chan, A.} (\bibinfo{year}{2021}{\natexlab{b}}).
\newblock \bibinfo{title}{Chinese white dolphin detection in the wild}.
\newblock In {\it \bibinfo{booktitle}{Proceedings of the 3rd ACM International
  Conference on Multimedia in Asia}\/} (pp. \bibinfo{pages}{1--5}).
\bibitem[{Zhang et~al.(2023{\natexlab{a}})Zhang, Zhang, Shi, Wang, Xu \&
  Chen}]{zhang2023sg}
\bibinfo{author}{Zhang, L.}, \bibinfo{author}{Zhang, N.}, \bibinfo{author}{Shi,
  R.}, \bibinfo{author}{Wang, G.}, \bibinfo{author}{Xu, Y.}, \&
  \bibinfo{author}{Chen, Z.} (\bibinfo{year}{2023}{\natexlab{a}}).
\newblock \bibinfo{title}{Sg-det: shuffle-ghostnet-based detector for real-time
  maritime object detection in uav images}.
\newblock {\it \bibinfo{journal}{Remote Sensing}\/},  {\it
  \bibinfo{volume}{15}\/}, \bibinfo{pages}{3365}.
\bibitem[{Zhang et~al.(2022{\natexlab{b}})Zhang, Xu, Zhang \&
  Tao}]{zhang2022vsa}
\bibinfo{author}{Zhang, Q.}, \bibinfo{author}{Xu, Y.}, \bibinfo{author}{Zhang,
  J.}, \& \bibinfo{author}{Tao, D.} (\bibinfo{year}{2022}{\natexlab{b}}).
\newblock \bibinfo{title}{Vsa: Learning varied-size window attention in vision
  transformers}.
\newblock In {\it \bibinfo{booktitle}{European conference on computer
  vision}\/} (pp. \bibinfo{pages}{466--483}).
\newblock \bibinfo{organization}{Springer}.
\bibitem[{Zhang et~al.(2021{\natexlab{c}})Zhang, Li, Ji, Zhao, Li \&
  Pan}]{zhang2021survey}
\bibinfo{author}{Zhang, R.}, \bibinfo{author}{Li, S.}, \bibinfo{author}{Ji,
  G.}, \bibinfo{author}{Zhao, X.}, \bibinfo{author}{Li, J.}, \&
  \bibinfo{author}{Pan, M.} (\bibinfo{year}{2021}{\natexlab{c}}).
\newblock \bibinfo{title}{Survey on deep learning-based marine object
  detection}.
\newblock {\it \bibinfo{journal}{Journal of Advanced Transportation}\/},  {\it
  \bibinfo{volume}{2021}\/}, \bibinfo{pages}{1--18}.
\bibitem[{Zhang et~al.(2023{\natexlab{b}})Zhang, Liu \& Wu}]{zhang2023efpnet}
\bibinfo{author}{Zhang, R.}, \bibinfo{author}{Liu, Q.}, \& \bibinfo{author}{Wu,
  K.} (\bibinfo{year}{2023}{\natexlab{b}}).
\newblock \bibinfo{title}{{EFPNet}: Effective fusion pyramid network for tiny
  person detection in {UAV} images}.
\newblock In {\it \bibinfo{booktitle}{CAAI International Conference on
  Artificial Intelligence}\/} (pp. \bibinfo{pages}{281--292}).
\newblock \bibinfo{organization}{Springer}.
\bibitem[{Zhang et~al.(2022{\natexlab{c}})Zhang, Feng, Zhang, Wang \&
  Mei}]{zhang2022finding}
\bibinfo{author}{Zhang, X.}, \bibinfo{author}{Feng, Y.},
  \bibinfo{author}{Zhang, S.}, \bibinfo{author}{Wang, N.}, \&
  \bibinfo{author}{Mei, S.} (\bibinfo{year}{2022}{\natexlab{c}}).
\newblock \bibinfo{title}{Finding nonrigid tiny person with densely cropped and
  local attention object detector networks in low-altitude aerial images}.
\newblock {\it \bibinfo{journal}{IEEE Journal of Selected Topics in Applied
  Earth Observations and Remote Sensing}\/},  {\it \bibinfo{volume}{15}\/},
  \bibinfo{pages}{4371--4385}.
\bibitem[{Zhang et~al.(2023{\natexlab{c}})Zhang, Feng, Zhang, Wang, Mei \&
  He}]{zhang2023semi}
\bibinfo{author}{Zhang, X.}, \bibinfo{author}{Feng, Y.},
  \bibinfo{author}{Zhang, S.}, \bibinfo{author}{Wang, N.},
  \bibinfo{author}{Mei, S.}, \& \bibinfo{author}{He, M.}
  (\bibinfo{year}{2023}{\natexlab{c}}).
\newblock \bibinfo{title}{Semi-supervised person detection in aerial images
  with instance segmentation and maximum mean discrepancy distance}.
\newblock {\it \bibinfo{journal}{Remote Sensing}\/},  {\it
  \bibinfo{volume}{15}\/}, \bibinfo{pages}{2928}.
\bibitem[{Zhang et~al.(2023{\natexlab{d}})Zhang, Tao \&
  Yin}]{zhang2023lightweight}
\bibinfo{author}{Zhang, Y.}, \bibinfo{author}{Tao, Q.}, \&
  \bibinfo{author}{Yin, Y.} (\bibinfo{year}{2023}{\natexlab{d}}).
\newblock \bibinfo{title}{A lightweight man-overboard detection and tracking
  model using aerial images for maritime search and rescue}.
\newblock {\it \bibinfo{journal}{Remote Sensing}\/},  {\it
  \bibinfo{volume}{16}\/}, \bibinfo{pages}{165}.
\bibitem[{Zhang et~al.(2023{\natexlab{e}})Zhang, Yin \&
  Shao}]{zhang2023enhanced}
\bibinfo{author}{Zhang, Y.}, \bibinfo{author}{Yin, Y.}, \&
  \bibinfo{author}{Shao, Z.} (\bibinfo{year}{2023}{\natexlab{e}}).
\newblock \bibinfo{title}{An enhanced target detection algorithm for maritime
  search and rescue based on aerial images}.
\newblock {\it \bibinfo{journal}{Remote Sensing}\/},  {\it
  \bibinfo{volume}{15}\/}, \bibinfo{pages}{4818}.
\bibitem[{Zhao et~al.(2024)Zhao, Song, Zhou, Yu, Zhang \&
  Liu}]{zhao2024heuristic}
\bibinfo{author}{Zhao, B.}, \bibinfo{author}{Song, R.}, \bibinfo{author}{Zhou,
  Y.}, \bibinfo{author}{Yu, L.}, \bibinfo{author}{Zhang, X.}, \&
  \bibinfo{author}{Liu, J.} (\bibinfo{year}{2024}).
\newblock \bibinfo{title}{Heuristic data-driven anchor generation for
  {UAV}-based maritime rescue image object detection}.
\newblock {\it \bibinfo{journal}{Heliyon}\/}, .
\bibitem[{Zhao et~al.(2023)Zhao, Zhang \& Zhao}]{zhao2023yolov7}
\bibinfo{author}{Zhao, H.}, \bibinfo{author}{Zhang, H.}, \&
  \bibinfo{author}{Zhao, Y.} (\bibinfo{year}{2023}).
\newblock \bibinfo{title}{{Yolov7}-sea: Object detection of maritime {UAV}
  images based on improved yolov7}.
\newblock In {\it \bibinfo{booktitle}{Proceedings of the IEEE/CVF Winter
  Conference on Applications of Computer Vision}\/} (pp.
  \bibinfo{pages}{233--238}).
\bibitem[{Zhou \& Li(2021)}]{zhou2021texture}
\bibinfo{author}{Zhou, X.}, \& \bibinfo{author}{Li, S.} (\bibinfo{year}{2021}).
\newblock \bibinfo{title}{A texture enhanced feature fusion method for small
  object detection}.
\newblock In {\it \bibinfo{booktitle}{2021 11th International Conference on
  Information Technology in Medicine and Education (ITME)}\/} (pp.
  \bibinfo{pages}{112--116}).
\newblock \bibinfo{organization}{IEEE}.
\bibitem[{Zhu et~al.(2023)Zhu, Ma, Wang \& Shi}]{zhu2023yolov7}
\bibinfo{author}{Zhu, Q.}, \bibinfo{author}{Ma, K.}, \bibinfo{author}{Wang,
  Z.}, \& \bibinfo{author}{Shi, P.} (\bibinfo{year}{2023}).
\newblock \bibinfo{title}{{YOLOv7}-{CSAW} for maritime target detection}.
\newblock {\it \bibinfo{journal}{Frontiers in Neurorobotics}\/},  {\it
  \bibinfo{volume}{17}\/}.
\bibitem[{Zhu et~al.(2019)Zhu, Hu, Lin \& Dai}]{zhu2019deformable}
\bibinfo{author}{Zhu, X.}, \bibinfo{author}{Hu, H.}, \bibinfo{author}{Lin, S.},
  \& \bibinfo{author}{Dai, J.} (\bibinfo{year}{2019}).
\newblock \bibinfo{title}{Deformable convnets v2: More deformable, better
  results}.
\newblock In {\it \bibinfo{booktitle}{Proceedings of the IEEE/CVF conference on
  computer vision and pattern recognition}\/} (pp.
  \bibinfo{pages}{9308--9316}).
\bibitem[{Zhu et~al.(2020)Zhu, Su, Lu, Li, Wang \& Dai}]{zhu2020deformable}
\bibinfo{author}{Zhu, X.}, \bibinfo{author}{Su, W.}, \bibinfo{author}{Lu, L.},
  \bibinfo{author}{Li, B.}, \bibinfo{author}{Wang, X.}, \&
  \bibinfo{author}{Dai, J.} (\bibinfo{year}{2020}).
\newblock \bibinfo{title}{Deformable {DETR}: Deformable transformers for
  end-to-end object detection}.
\newblock {\it \bibinfo{journal}{arXiv preprint arXiv:2010.04159}\/}, .
\bibitem[{Zivkovic(2004)}]{zivkovic2004improved}
\bibinfo{author}{Zivkovic, Z.} (\bibinfo{year}{2004}).
\newblock \bibinfo{title}{Improved adaptive gaussian mixture model for
  background subtraction}.
\newblock In {\it \bibinfo{booktitle}{Proceedings of the 17th International
  Conference on Pattern Recognition, 2004. ICPR 2004.}\/} (pp.
  \bibinfo{pages}{28--31}).
\newblock \bibinfo{organization}{IEEE} volume~\bibinfo{volume}{2}.

\end{thebibliography}
